%% file: main.tex
\documentclass{bmvc2k}

\usepackage{enumitem}
\usepackage{amssymb}
\usepackage{tabularx}
\usepackage{booktabs}
\usepackage{tikz}
\usepackage{subcaption}
\newif\ifreview
\reviewfalse   %

\newif\ifsuppinmain
\suppinmaintrue   %

\ifreview
  \bmvcreviewcopy{124}
\fi
\newif\ifreviewmode
\reviewmodefalse   %
\ifreviewmode
  \newcommand{\PPc}[1]{}  %
  
  \definecolor{nicolascolor}{RGB}{0,128,128}
  \newcommand{\nicolasc}[1]{}  %
  
  \definecolor{nathanielcolor}{RGB}{128,0,128}
  \newcommand{\nathanielc}[1]{}  %
  
  \newcommand{\ANc}[1]{}  %
  \definecolor{cornflowerblue}{RGB}{100, 149, 237}
  \newcommand{\ameliec}[1]{}  %
  
  \definecolor{updcolor}{RGB}{0,70,255}
  \newcommand{\upd}[1]{{\color{updcolor}{#1}}}
  \newcommand{\updc}[1]{}  %
\else
  \newcommand{\upd}[1]{#1}
  \newcommand{\updc}[1]{}
  \newcommand{\PPc}[1]{}
  
  \newcommand{\nicolasc}[1]{}
  
  \newcommand{\nathanielc}[1]{}
  
  \newcommand{\ameliec}[1]{}
  
  \newcommand{\ANc}[1]{}
  
\fi

\newcommand{\myparagraph}[1]{\smallskip\noindent\textbf{#1~}}

\title{AURA: AUdio-dRiven streaming Avatar}

\addauthor{Nathaniel Cohen}{nathaniel.cohen@kyutai.org}{1,2}
\addauthor{Nicolas Dufour}{nicolas.dufour@kyutai.org}{1}
\addauthor{Amélie Royer}{amelie@kyutai.org}{1}
\addauthor{Alasdair Newson}{anewson@isir.upmc.fr}{2}
\addauthor{Patrick Pérez}{perezpatrick@google.com}{3}
\addinstitution{
 Kyutai\\
 Paris, France
}
\addinstitution{
 ISIR, Sorbonne Université\\
 Paris, France
}
\addinstitution{
 Google DeepMind\\
 (work done while at Kyutai)
}
\runninghead{Cohen et al.}{AURA: AUdio-dRiven streaming Avatar}

\begin{document}

\maketitle

\begin{figure}[h!]
    \centering
    \includegraphics[width=\linewidth]{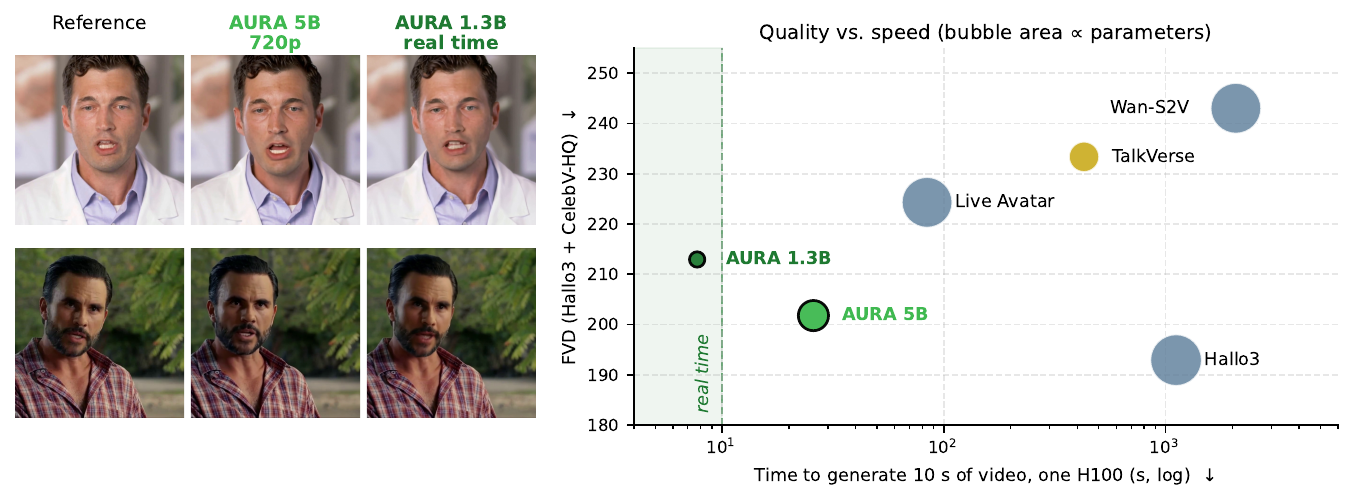}
    \caption{\upd{\textbf{AURA streams talking-head video chunk by chunk on a single GPU}: AURA 5B at $720$p and $9.7$\,fps, AURA 1.3B at $576$p in real time ($20.7$\,fps for $16$\,Hz playback).
    \textit{(Left)} Two reference identities and a frame generated by each model at $t = 4$\,s, from the same reference image and audio.
    \textit{(Right)} Quality--speed trade-off: FVD averaged over the \textsc{Hallo3 test} and \textsc{CelebV-HQ test} sets versus\ end-to-end time to generate $10$\,s of video on one H100 (log scale, lower is better). Circle areas are proportional to parameter counts, and the shaded band marks real-time generation.} Video results can be found  at \href{https://kyutai.org/aura/}{kyutai.org/aura}.
    }
    \label{fig:teaser}
\end{figure}

\begin{abstract}
Current audio-driven video diffusion models can synthesize realistic talking-head videos but generally require dozens of diffusion denoising steps over  multi-second video chunks, making them unsuitable for streaming applications such as voice assistants and virtual avatars, where low-latency and online causal generation are required properties.
\upd{AURA generates video chunk by chunk with a rolling KV cache, requiring only the audio of the current chunk \upd{to start generating}, rather than a multi-second audio look-ahead like bidirectional models. It comes in two sizes: AURA 1.3B streams $576$p video in real time on a single GPU ($20.7$\,fps for a $16$\,Hz playback rate), making it suited to interactive avatar applications, and AURA 5B streams $720$p video at $9.7$\,fps.}
Our two-stage distillation pipeline first aligns the causal student with the teacher, by training it to match the teacher's trajectories with fewer diffusion steps and a chunk-wise \upd{sliding window} causal attention mask, 
then applies Distribution Matching Distillation (DMD) with backward simulation to close the remaining quality gap. On standard talking-head benchmarks, \upd{AURA 1.3B runs $23\times$ faster than its own bidirectional teacher and reaches a quality comparable to TalkVerse at a lower resolution, and AURA 5B matches its TalkVerse teacher at $720$p while running $17\times$ faster.} To support reproducibility and further research on streaming avatar generation, we release \upd{the weights of both models} along with a dataset of $30$K synthetic videos and their precomputed teacher trajectories.

\end{abstract}

\section{Introduction}
\label{sec:intro}
\input{sections/01_introduction}

\section{Related Work}
\label{sec:related}
\input{sections/02_related_work}

\section{Method}
\label{sec:method}
\input{sections/04_method}

\section{Experiments}
\label{sec:experiments}
\input{sections/05_experiments}

\section{Conclusions}
\label{sec:conclusion}
\input{sections/06_conclusion}
\clearpage

\bibliography{egbib-2}

\ifsuppinmain
  \newpage
  \appendix
  \section{Supplementary Material}
  \input{sections/07_supplementary}
\fi

\end{document}

%% file: sections/01_introduction.tex
Recent speech-to-speech models~\cite{defossez2024moshi, roy2026personaplexvoicerolecontrol} are able to model full-duplex voice interaction 
in real-time, paving the way for voice assistants that hold natural and smooth conversations. However, these models only generate audio signal.
In contrast, human spoken communication relies on visual cues and facial expressions, via an interplay of the audio and visual modalities. For instance,
lip motion alters perceived phonemes~\cite{mcgurk1976hearing}, visual cues sharpen comprehension in noise~\cite{sumby1954visual}, and an embodied talking face enhances the sense of social presence over voice alone~\cite{cassell2000embodied}. 
To model such behaviors, audio-driven video generation models can synthesize talking-head videos from a reference image and an input speech audio, enabling applications such as virtual avatars, synthetic presenters, and non-player characters in games.
All these applications, however, share a common constraint: the avatar must be generated on-the-fly, as the audio arrives, from a stream of audio of unbounded duration.

State-of-the-art talking-head models such as TalkVerse~\cite{wang2025talkversedemocratizingminutelongaudiodriven} and Wan-S2V~\cite{gao2025wans2vaudiodrivencinematicvideo} produce high-quality audio-driven videos but are ill-suited to the streaming setting. Because they rely on bidirectional attention, they must generate entire video chunks of several seconds at once, running dozens of denoising steps over the full chunk before producing any frame, which on a single GPU takes minutes per chunk.
Recently, CausVid~\cite{yin2025slowbidirectionalfastautoregressive} and Self-Forcing~\cite{huang2025selfforcingbridgingtraintest} have explored distilling such bidirectional video diffusion teachers into few-step causal students that generate video chunk by chunk in an autoregressive manner. However, these methods target only the text-to-video and image-to-video settings. Extending this approach to the audio-driven setting, where the model must additionally track a dense time-aligned speech signal, has so far been demonstrated only on much larger models (14--18B parameters) ~\cite{huang2026liveavatarstreamingrealtime, shen2026soulxflashtalkrealtimeinfinitestreaming, low2025talkingmachinesrealtimeaudiodrivenfacetimestyle}, which remain too large for real-time inference on a single GPU (only 2 fps). A second obstacle is the scarcity of suitable training data: existing public talking-head datasets such as Hallo3 \cite{cui2025hallo3highlydynamicrealistic} and CelebV \cite{zhu2022celebvhqlargescalevideofacial} consist of clips too short to cover the chunk length of modern teachers like TalkVerse, ruling them out for distillation (see \hyperref[sec:method:dataset]{Section~\ref{sec:method:dataset}}).

\upd{We introduce AURA, a family of causal few-step speech-to-video (S2V) models that generate streaming talking-head videos from live audio. %
In particular, AURA 1.3B runs in real time, generating $576$p video at $20.7$ frames per second for a $16$\,Hz playback rate on a single GPU, which makes it directly usable for the interactive applications above, such as a talking face for a voice assistant. AURA 5B targets higher resolution: it streams $720$p video at $9.7$\,fps, a $17\times$ speedup over its TalkVerse teacher.} For training, we generate and release a new talking-head dataset comprising $30$K videos ($38.7$ hours) produced with TalkVerse, conditioned on reference images and text descriptions from the Hallo3 dataset~\cite{cui2025hallo3highlydynamicrealistic} and synthetic speech audio.
To train our model, we use a two-stage distillation pipeline, similar to CausVid: 
Stage 1 first  trains the causal student to match the  bidirectional TalkVerse teacher's Ordinary Differential Equation (ODE) trajectories in order to reduce the gap between the bidirectional teacher and the causal student architecture. 
Then, Stage 2 closes the few-step gap with Distribution Matching Distillation (DMD) and \textit{backward simulation}: by training the student on intermediate states from its own denoising rollouts, the Stage 2 loss reflects accumulating errors the student may actually produce at inference.
To further improve efficiency, we train the model to support chunk-wise causal \upd{sliding window} attention, allowing us to use a rolling KV cache with bounded memory at inference. We also use a single KV cache for all diffusion denoising timesteps, further reducing memory footprint.

With this approach, on standard talking-avatar benchmarks, AURA \upd{5B} matches \upd{or improves on} its bidirectional teacher on visual quality (FVD \cite{unterthiner2019accurategenerativemodelsvideo} and JEDi \cite{luo2024fvdenhancedevaluationmetrics} metrics) and audio-visual alignment, while running $17\times$ faster with only four denoising steps per chunk\upd{, and AURA 1.3B reaches a comparable quality at a lower resolution and in real time ($23\times$ faster than its own teacher). Both models are faster than all baselines we compare against in the same single-GPU setting.} \hyperref[fig:teaser]{Figure~\ref{fig:teaser}} shows an excerpt of qualitative samples generated by \upd{both models}, as well as \upd{their} speed--quality tradeoff against prior work.

In summary, we introduce AURA, which makes the following contributions:
\begin{itemize}[itemsep=0.2ex, leftmargin=1em]
    \item \upd{A} two-stage distillation method (ODE-trajectory initialization followed by Distribution Matching Distillation with backward simulation) paired with a streaming inference scheme based on a rolling KV cache of past chunks\upd{, shared by both models};
    
    \item \upd{\textbf{AURA 5B}, a} causal few-step S2V model that \upd{matches or improves on} its bidirectional teacher on FVD and JEDi metrics, with a $17\times$ speedup. Our inference pipeline streams $720$p video at $9.7$\,fps on a single GPU, the highest throughput among audio-driven baselines we compare against;

    \item \upd{\textbf{AURA 1.3B}, a smaller model trained with the same pipeline, %
    which streams $576$p video in \emph{real time} on a single GPU ($20.7$\,fps generated for a $16$\,Hz playback rate), a $23\times$ speedup over its own bidirectional teacher;}

    \item A public dataset of $30$K \upd{talking-head videos generated by TalkVerse from synthetic speech, released both as decoded videos and as the latent ODE trajectories of the teacher}, to support future research on streaming S2V distillation\upd{, together with the weights of our $1.3$B teacher and its own $30$K ODE trajectories};

    \item \upd{A live demo coupling AURA 1.3B with Unmute, a cascaded voice assistant built on Kyutai's streaming speech-to-text and text-to-speech models~\cite{dsm} and a text LLM, in which users hold a natural spoken conversation with a synthetic avatar in real time on a single H100 GPU, available at \href{https://kyutai.org/aura/}{kyutai.org/aura}.}
    
\end{itemize}

%% file: sections/02_related_work.tex
\paragraph{Diffusion and Flow Matching.}
\label{sec:related:diffusion}

Image Generative models have evolved through several paradigms:  
Generative Adversarial Networks (GANs)~\cite{goodfellow2014generativeadversarialnetworks,karras2019stylebasedgeneratorarchitecturegenerative,brock2019largescalegantraining} are efficient as they generate images in a single forward pass but suffer from known limitations around mode coverage and training stability. More recently,  diffusion models~\cite{ho2020denoisingdiffusionprobabilisticmodels,song2021scorebasedgenerativemodelingstochastic,dhariwal2021diffusionmodelsbeatgans,song2022denoisingdiffusionimplicitmodels,rombach2022highresolutionimagesynthesislatent,karras2022elucidatingdesignspacediffusionbased} and flow matching~\cite{lipman2023flowmatchinggenerativemodeling,liu2022flowstraightfastlearning} have become the dominant paradigm for image generation. Indeed, both frameworks are closely related and model data through an iterative multi-step refinement process, yielding  stable training with high sample diversity but at the cost of slow sampling. Most modern large-scale video generators, including our teacher TalkVerse \cite{wang2025talkversedemocratizingminutelongaudiodriven}, are trained under the flow matching framework, and our work also builds on this formulation. 

\myparagraph{Video Diffusion Models.}
\label{sec:related:videodiffusion}
Recent work has further scaled diffusion for   video generation. Early efforts such as Imagen Video~\cite{ho2022imagenvideohighdefinition}, Make-A-Video~\cite{singer2022makeavideotexttovideogenerationtextvideo}, Align Your Latents~\cite{blattmann2023alignlatentshighresolutionvideo}, and Stable Video Diffusion~\cite{blattmann2023stablevideodiffusionscaling} extended image diffusion to short clips, typically by inflating 2D architectures with temporal layers. More recent open-source and industrial models, Sora~\cite{videoworldsimulators2024}, Movie Gen~\cite{polyak2025moviegencastmedia}, Open-Sora~\cite{zheng2024opensorademocratizingefficientvideo}, CogVideoX~\cite{yang2025cogvideoxtexttovideodiffusionmodels}, HunyuanVideo~\cite{kong2025hunyuanvideosystematicframeworklarge}, and Wan~\cite{wan2025wanopenadvancedlargescale}--- adopt diffusion transformer (DiT) architectures and operate in a compressed latent space, scaling to $720$p--$1080$p resolutions and minute-long clips. These models are typically non-causal: they generate the entire video sequence jointly over $50$ or more denoising steps, and are thus not well adapted to the  latency requirements of streaming  applications. Current SotA talking-head models are based on such architectures. For instance,  Hallo3~\cite{cui2025hallo3highlydynamicrealistic}  adapts CogVideoX~\cite{yang2025cogvideoxtexttovideodiffusionmodels} for portrait animation with a DiT-adapted identity reference network,  
while closed-source systems such as OmniHuman~\cite{lin2025omnihuman1rethinkingscalinguponestage} extend DiT-based audio-driven generation to full-body animation. 
\upd{In this work, we distill two such non-causal teachers built on Wan, TalkVerse~\cite{wang2025talkversedemocratizingminutelongaudiodriven} (from Wan2.2-5B) for AURA 5B and a $1.3$B teacher that we train from Wan2.1-1.3B for AURA 1.3B, into causal, few-step streaming students.}

\myparagraph{Distillation of Diffusion Models.}
\label{sec:related:distillation}
The slow inference of diffusion models has motivated extensive work on
\emph{distillation}, training a faster few-step student to reproduce a slow
teacher's output distribution. One particularly successful line is  Distribution Matching Distillation
(DMD)~\cite{yin2024onestepdiffusiondistributionmatching}, which trains a one-step student
to minimize the reverse KL to the teacher via a learned critic; DMD2~\cite{yin2024improveddistributionmatchingdistillation}
extends it to a few-step generator and introduces \emph{backward simulation},
computing the loss on the student's own rollouts rather than forward-diffused data.
CausVid~\cite{yin2025slowbidirectionalfastautoregressive} applies DMD to distill a
non-causal video teacher into a causal few-step student for streaming
text-to-video. A challenge specific to causal video distillation is the train/test
gap: the student is trained on teacher-provided past frames but must condition on
its own previously generated frames at inference. Self-Forcing~\cite{huang2025selfforcingbridgingtraintest}
and its variants~\cite{cui2025selfforcingminutescalehighqualityvideo,zhu2026causalforcingautoregressivediffusion,liu2025rollingforcingautoregressivelong,yang2025longliverealtimeinteractivelong}
address this temporal gap by rolling the student out autoregressively through a
training-time KV cache with stop-gradient. We instead build on DMD2's backward
simulation~\cite{yin2024improveddistributionmatchingdistillation}, which exposes
the student to its own denoising trajectory, and adapt it to streaming video
(Sec.~\ref{sec:method:stage2}) so that the student also conditions on its own past
generations, without maintaining a training-time cache.

\myparagraph{Streaming Talking Heads.}
\label{sec:related:streaming}
A handful of recent works tackle audio-driven talking-head generation in a streaming and \upd{few-step} setting. Talking Machines~\citep{low2025talkingmachinesrealtimeaudiodrivenfacetimestyle} is, to our knowledge, the first to address this regime, distilling a $18$B model with DMD and sparse causal attention; weights are not released. Live Avatar~\citep{huang2026liveavatarstreamingrealtime} distills the $14$B Wan-S2V model \cite{gao2025wans2vaudiodrivencinematicvideo} into a $14$B few-step student that runs at $16$\,fps and $480$p on a $5$-GPU setup. SoulX-FlashTalk~\citep{shen2026soulxflashtalkrealtimeinfinitestreaming} adopts a similar few-step distillation strategy from a $14$B teacher into a $14$B student and requires $4$ GPUs to reach real-time throughput. In contrast, \upd{AURA streams on a single GPU: AURA 5B} distills a $5$B teacher into a same-size causal $4$-step student that delivers $720$p video at $9.7$\,fps\upd{, and AURA 1.3B, trained with the same pipeline from our own $1.3$B teacher, reaches real time at $576$p ($20.7$\,fps for a $16$\,Hz playback rate)}.
Closest to our work is Avatar Forcing~\citep{cui2026avatarforcingonestepstreamingtalking} whose pre-print was concurrently published: It uses a similar DMD-based distillation process but requires a look-ahead window of future audio to generate each frame by design, leading to a small yet incompressible streaming delay.

%% file: sections/04_method.tex
In this section, we present AURA, our two-stage pipeline that distills a non-causal audio-driven video diffusion teacher (TalkVerse) into a $4$-step causal student. %
We first recall Distribution Matching Distillation (\hyperref[sec:method:prelim:causvid]{Sec.~\ref{sec:method:prelim:causvid}}) and the TalkVerse teacher (\hyperref[sec:method:prelim:TalkVerse]{Sec.~\ref{sec:method:prelim:TalkVerse}}) to introduce necessary notations and context. \hyperref[sec:method:model]{Section~\ref{sec:method:model}}  then introduces the causal student and the per-chunk mixed-timestep design that enables a compact KV cache. \hyperref[sec:method:stage1]{Section~\ref{sec:method:stage1}} and  \hyperref[sec:method:stage2]{Section ~\ref{sec:method:stage2}} cover the two training stages, and  \hyperref[sec:method:inference]{Section~\ref{sec:method:inference}} finally describes streaming inference.

\subsection{Preliminaries}
\label{sec:method:prelim}

\label{sec:method:prelim:causvid}
\paragraph{Distribution Matching Distillation (DMD).} DMD~\citep{yin2024onestepdiffusiondistributionmatching} trains a few-step student generator $G_\theta$ to match a multi-step teacher diffusion model at the distribution level, rather than along individual sampling trajectories. Concretely, DMD minimizes, in expectation over the diffusion timestep $t \sim \mathcal{U}(0, 1)$, the reverse KL divergence 
between the student distribution $p_\text{gen}$ and the data distribution $p_\text{data}$, both diffused to noise level $t$. We adopt $t \in [0,1]$ throughout this paper, with $t=0$ corresponding to clean latents and $t=1$ to pure noise. While the densities $p_\text{gen}$ and $p_\text{data}$ are intractable, the gradient of the KL with respect to $\theta$ admits a closed-form expression as a difference of two score functions evaluated on student samples:
\begin{align}
\label{eq:dmd}
\nabla_\theta \mathcal{L}_\text{DMD} 
&\triangleq \mathbb{E}_t\bigl[\nabla_\theta \mathrm{KL}\bigl(p_{\text{gen},t} \,\|\, p_{\text{data},t}\bigr)\bigr] \nonumber \\
&\approx -\mathbb{E}_t \int \bigl(s_\text{data}(\Psi(G_\theta(z), t), t) - s_{\text{gen},\eta}(\Psi(G_\theta(z), t), t)\bigr) \frac{dG_\theta(z)}{d\theta}\, dz,
\end{align}
where $z$ denotes the noise input to the student, $\Psi(\cdot, t)$ is the forward diffusion operator that produces a sample at noise level $t$, $s_\text{data}$ is the frozen teacher score, and $s_{\text{gen},\eta}$ is an online score network (called \textit{critic} in this paper) with parameters $\eta$ that estimates $\nabla \log p_{\text{gen},t}$ on student samples. In practice, $s_{\text{gen},\eta}$ is trained jointly with $G_\theta$ via the standard denoising score-matching loss on student samples, and the two are updated in an alternating fashion.

\myparagraph{Backward simulation.}
Directly applying DMD to multi-step students yields a training-inference mismatch: at training time, the student is fed forward-diffused real samples $x_t = \Psi(x_0, t)$, whereas at inference time, intermediate inputs come from previous denoising steps of the student itself. To close this gap, DMD2 \citep{yin2024improveddistributionmatchingdistillation} introduces \emph{backward simulation}. At each training iteration, the student is unrolled for a few denoising steps to produce a synthetic intermediate sample $\hat{x}_t$, which then replaces the forward-diffused real sample as input to the DMD objective. Because $G_\theta$ is itself a few-step generator, this unrolling is cheap. The DMD gradient and the score-matching update for $s_{\text{gen},\eta}$ are then computed at $\hat{x}_t$ rather than at $\Psi(x_0, t)$, aligning the input distribution seen during training with the one produced at inference. \upd{DMD2 also adds a GAN loss on real data. We adopt backward simulation, together with the two time-scale update rule of DMD2, in the second stage of our pipeline (\hyperref[sec:method:stage2]{Sec.~\ref{sec:method:stage2}}), but not the GAN loss: it requires real videos and makes training less stable. The DMD objective alone already brings our student close to its teacher (Sec.~\ref{sec:exp:ablations}).}

\myparagraph{TalkVerse Architecture.}
\label{sec:method:prelim:TalkVerse}
We use TalkVerse~\cite{wang2025talkversedemocratizingminutelongaudiodriven} as our teacher model, a $5$B-parameter audio-driven video diffusion model built on the Wan2.2-5B~\cite{wan2025wanopenadvancedlargescale} DiT backbone.  %
Its video VAE uses compression factors $(t,h,w)=(4,16,16)$, enabling efficient $720$p generation. The model is conditioned on a reference identity image, a speech audio track encoded with Wav2Vec2~\cite{baevski2020wav2vec20frameworkselfsupervised} and injected via cross-attention to temporally aligned visual tokens, and a text prompt encoded with T5-XXL~\cite{raffel2023exploringlimitstransferlearning}. Long videos are generated with a sliding-window mechanism that conditions each window on a set of \emph{motion frames} compressed by a FramePack module~\cite{zhang2025framecontextpackingdrift}, and the reference image is assigned a positional offset placed ahead of the denoised window. Inference uses $50$ denoising steps with a flow-matching solver and classifier-free guidance. We provide a more detailed description in the supplementary material  \ref{sec:supmat:prelim:TalkVerse}. \upd{TalkVerse is directly the teacher of AURA 5B; for AURA 1.3B, we train a $1.3$B teacher with the TalkVerse pipeline (Sec.~\ref{sec:exp:setup} and supplementary material~\ref{sec:supmat:aura13b}).}

\begin{figure}[!tbh]
    \centering
    \begin{subfigure}[t]{0.39\linewidth}
        \centering
        \includegraphics[height=3.3cm]{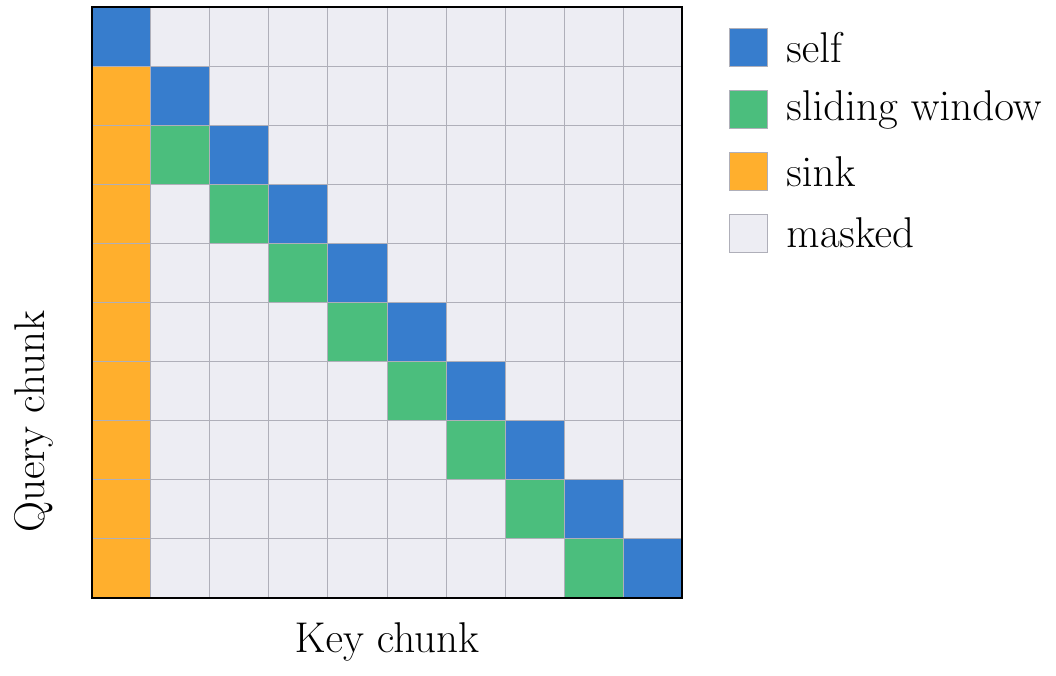}
        \caption{\footnotesize \textbf{Sliding window causal attention mask used during training}. Each  chunk of 4 latent frames can attend to itself, the preceding chunk and the initial chunk acting as a sink. The attention is bidirectional among the 4 frames inside each chunk.}
        \label{fig:attmask_mask}
    \end{subfigure}
    \hfill 
    \begin{subfigure}[t]{0.58\linewidth}
        \centering
        \includegraphics[height=3.8cm]{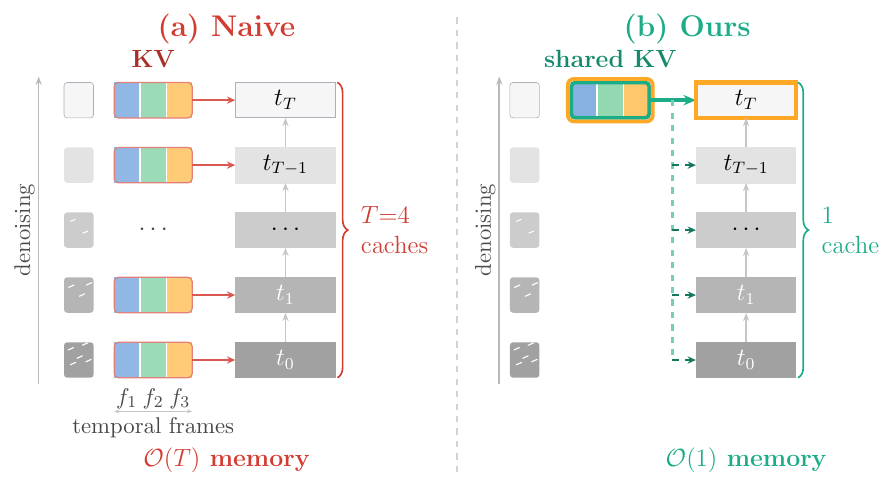}
        \caption{\footnotesize \textbf{KV caching with and without per-chunk mixed timesteps}. \textit{Left:} A naive scheme stores one KV cache per denoising step, thus memory scales linearly with the number of denoising steps $T$. \textit{Right:} With per-chunk mixed timesteps training, each past chunk can be cached at a single noise level, reducing memory cost.}
        \label{fig:attmask_kvcache}
    \end{subfigure}
    \caption{\textit{(Left)} Sliding window chunk-wise causal attention mask used during training by the causal student. \textit{(Right)} Per-chunk mixed denoising timestep training allows the use of a single shared KV cache across all denoising steps at inference.}
    \label{fig:attmask}
    \vspace{-10pt}
\end{figure}

\subsection{AURA Architecture}
\label{sec:method:model}
To turn the $50$-step bidirectional teacher (TalkVerse) into a $4$-step causal student, we introduce three architectural modifications: 
chunked causal \upd{sliding window} attention, removal of motion frames, and per-chunk mixed timesteps. In combination, these changes also enable us to keep a compact rolling KV cache as illustrated in \hyperref[fig:attmask]{Figure \ref{fig:attmask}b}.

\myparagraph{Chunked causal attention.} The student generates videos in chunks of $n_c = 4$ latent frames ($\sim 0.64$\,s at the $25$\,Hz native rate), whereas the teacher generates $30$ latent frames ($\sim 5$\,s) in a single pass. We enforce causality between chunks in the student: during training, a block-wise causal attention mask prevents frames in chunk $C_i$ from attending to frames in chunks $C_j$ for $j > i$, while attention is bidirectional within a chunk (\hyperref[fig:attmask]{Figure \ref{fig:attmask}a}). %

\myparagraph{No motion frames.}  TalkVerse normally conditions each generated video sequence on a window of past frames (motion frames) to maintain temporal coherence across windows. In our causal student, the KV cache already carries the past context required for coherence, thus we remove the motion frame mechanism entirely. This also saves us the cost of a few forward passes through the model required to process motion frames for every new chunk. %

\begin{figure}[t]
    \centering
    \includegraphics[width=0.97\linewidth]{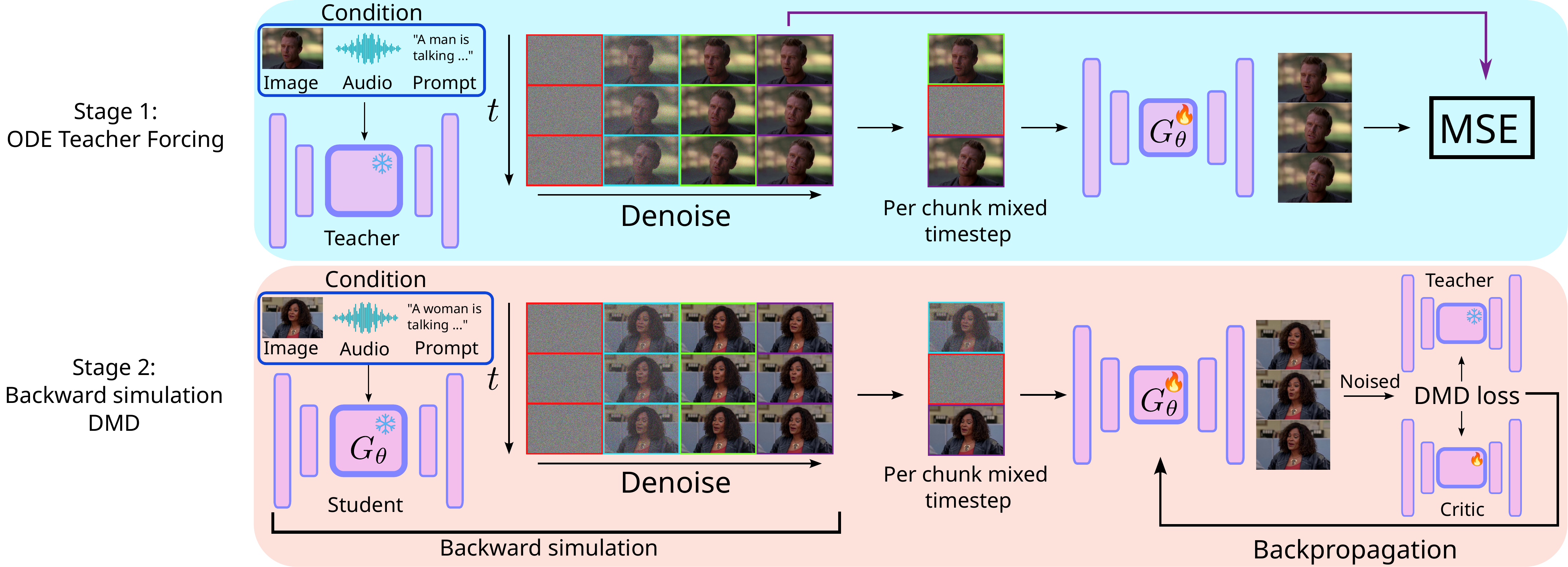}
        \caption{\textbf{Two-stage distillation pipeline}. \textit{Stage~1 (top, ODE teacher forcing)}: the frozen teacher generates a multi-step ODE trajectory from noise to clean latents, conditioned on a reference image, speech audio, and text prompt. A \emph{per-chunk mixed timestep} state is sampled from this trajectory and fed to the causal student $G_\theta$, which is trained to predict the corresponding clean latents under an MSE loss. \textit{Stage~2 (bottom, DMD with backward simulation)}: the student generates a video in inference mode without gradients (backward simulation, left), and a per-chunk mixed-timestep state is sampled from its own denoising trajectory. The student is then re-run on this state with gradients enabled, and a DMD loss is computed by comparing the score predicted by a frozen teacher and an online fake score network (right). Gradients flow back through the gradient-enabled student pass only.}

    \label{fig:pipeline}
    \vspace{-10pt}
\end{figure}

\myparagraph{Per-chunk mixed timesteps: a compact KV cache.}
At $720$p, the KV cache is already large at a single noise level due to the high spatial token count per frame. A naive implementation that would scale one (temporal) KV cache per diffusion step would increase the memory cost by the number of denoising steps per chunk. To avoid this, we train the model to handle sequences in which different chunks are at different noise levels: each chunk is assigned a single diffusion timestep $t_i$, shared by all frames within the chunk (\hyperref[fig:pipeline]{Figure~\ref{fig:pipeline}}, center). This \emph{per-chunk mixed timesteps} scheme follows CausVid~\cite{yin2025slowbidirectionalfastautoregressive}, which adopts it without explicit motivation. We argue it is a key design choice to enable a compact streaming KV cache.
At inference, the current chunk is fed at a high noise level $t_{\mathrm{curr}}$ while previously generated chunks, cached in the KV cache, are at the lowest noise level reached during their own generation. The cache therefore stores keys and values at a \emph{single} low noise level per past chunk, rather than at every diffusion step (\hyperref[fig:attmask_kvcache]{Fig.~\ref{fig:attmask_kvcache}}). In contrast, prior streaming audio-driven models that retain multiple noise levels in the cache pay a proportional memory cost~\cite{huang2026liveavatarstreamingrealtime}.

\subsection{Stage 1: ODE Teacher Forcing}
\label{sec:method:stage1}
As reported in CausVid~\cite{yin2025slowbidirectionalfastautoregressive}, directly applying DMD to a causal student starting from teacher weights is unstable as the student differs architecturally from the teacher. Therefore, we adapt their ODE-distillation initialization strategy  to our setting. %
Stage~1 trains the causal student to follow the teacher ODE under the new architecture (\hyperref[fig:pipeline]{Figure \ref{fig:pipeline}, top}), bringing the student close enough to the teacher distribution so that Stage~2 can converge.

\myparagraph{Training data.} We train Stage~1 on our synthetic dataset described in \hyperref[sec:method:dataset]{Section \ref{sec:method:dataset}}. Each video comes with latents stored at five timesteps along a $50$-step ODE trajectory from the teacher. At training time, the student is queried at the four  timesteps $\{0.25, 0.5, 0.75, 1\}$ and supervised to predict the clean latent at $t=0$.

\myparagraph{Training objective.} For each training video, we independently sample a noise level for each chunk uniformly at random from the four cached non-zero timesteps, following the per-chunk mixed timestep scheme of \hyperref[sec:method:model]{Section~\ref{sec:method:model}}. This matches the heterogeneous-noise configurations the student encounters at inference, where the current chunk is noisy while past chunks (in the KV cache) are near-clean (at lowest noise level). The student processes the resulting sequence under the block-wise causal mask and predicts a velocity field $v_\theta(z_t, t, c)$, where $z_t$ denotes the noised latent at level $t$ (\hyperref[sec:method:prelim:causvid]{Section\ref{sec:method:prelim:causvid}}) and $c$ denotes the text, audio and identity conditioning. From the predicted velocity we reconstruct the clean latent $\hat{x}_0$ and apply an MSE loss directly on $\hat{x}_0(z_t, t, c)$ rather than on the velocity. The Stage~1 loss reads:
\begin{equation}
\label{eq:stage1_loss}
    \mathcal{L}_{\text{stage1}} = \mathbb{E}_{t, z_t, c} \left[ \big\| \hat{x}_0(z_t, t, c) - x_0 \big\|_2^2 \right],
\end{equation}
where $x_0$ is the clean latent produced by the teacher. %
The objective is to match the clean latents of the teacher, not the full flow trajectory: at inference, the $4$-step student does not integrate the velocity field but alternates between $\hat{x}_0$ prediction and renoising (see  \ref{sec:supmat:sampling}).

\subsection{Stage 2: Backward Simulation DMD}
\label{sec:method:stage2}
Stage~1 yields a causal few-step student that closely follows teacher ODE trajectories, but its sample quality remains bounded by the compression from many-step to few-step inference. In Stage~2 we apply the DMD objective introduced in \hyperref[sec:method:prelim:causvid]{Section ~\ref{sec:method:prelim:causvid}} (\hyperref[eq:dmd]{Equation~\eqref{eq:dmd}}) to push the student output distribution toward the teacher distribution, with a sampling strategy adapted to our causal, chunked setting (\hyperref[fig:pipeline]{Figure \ref{fig:pipeline}, bottom}).

\myparagraph{Models.} We maintain three networks: the frozen teacher $f_{\text{teacher}}$, the trainable student $G_\theta$ (initialized from Stage~1), and a trainable critic $f_\phi$ that estimates $\nabla \log p_{\text{gen},t}$, playing the role of $s_{\text{gen},\eta}$ in \hyperref[eq:dmd]{Equation~\eqref{eq:dmd}}. Following the two time-scale update rule of DMD2~\citep{yin2024improveddistributionmatchingdistillation}, the critic is initialized from the teacher and updated $U = 5$ times between each student update.

\myparagraph{DMD objective.} Applying \hyperref[eq:dmd]{Equation~\eqref{eq:dmd}} to our setup, the student gradient becomes
\begin{equation}
\label{eq:dmd_stage2}
    \nabla_\theta \mathcal{L}_{\text{DMD}} = \mathbb{E}_{t,\, \hat{x}_0} \left[ \big( f_\phi(\hat{x}_t, t) - f_{\text{teacher}}(\hat{x}_t, t) \big) \frac{\partial \hat{x}_0}{\partial \theta} \right], \quad \hat{x}_0 = G_\theta(z, c),
\end{equation}
where $z$ is the noise input, $c$ denotes the conditioning (audio, prompt, and reference image), $t \sim \mathcal{U}(0,1)$ indexes the diffusion noise level, and $\hat{x}_t = \Psi(\hat{x}_0, t)$ is the forward-diffused student sample at level $t$. The two terms inside the bracket correspond, respectively, to the estimated student score and the teacher score, and their difference approximates the gradient of the reverse KL divergence $\mathrm{KL}(p_{\text{gen},t} \,\|\, p_{\text{data},t})$ (\hyperref[sec:method:prelim:causvid]{Section \ref{sec:method:prelim:causvid}}). The critic $f_\phi$ is trained jointly with $G_\theta$ via the standard denoising score-matching loss on student samples.

\myparagraph{Backward simulation.} A central question in few-step distillation is \emph{which} samples $\hat{x}_0$ to use when computing the DMD gradient. We adopt the backward simulation strategy of DMD2~\citep{yin2024improveddistributionmatchingdistillation} (Sec.~\ref{sec:method:prelim:causvid}) and apply it to our causal, chunked student. At each iteration, we proceed in three steps: (i) we run a full multi-step rollout of $G_\theta$ without backpropagation to obtain the intermediate noisy states $\{\hat{x}_{t_k}\}$ ($t_k \in \{0.25, 0.5, 0.75, 1\}$) along its own sampling trajectory; (ii) we re-feed these states through $G_\theta$ with gradients enabled, using per-chunk mixed timesteps (Sec.~\ref{sec:method:model}) sampled uniformly across the trajectory; (iii) we evaluate the DMD gradient (Eq.~\ref{eq:dmd_stage2}) at the resulting $\hat{x}_0$ (\hyperref[fig:pipeline]{Figure \ref{fig:pipeline}, bottom}). The student is thus supervised at the intermediate states it encounters at inference, reducing the training--inference mismatch.

\subsection{Inference: Causal Generation with  rolling KV Cache}
\label{sec:method:inference}

At inference, the student generates the video autoregressively, one chunk of $4$ latent frames at a time, in $4$ diffusion steps per chunk. Within each chunk, we adopt the predict-and-renoise scheme of DMD2~\citep{yin2024improveddistributionmatchingdistillation} (see supplementary material \ref{sec:supmat:sampling}).
To extend generation to longer sequences, we use a rolling KV cache  with a first-chunk anchor that provides temporal context across chunks, and a RoPE \cite{su2024roformer} re-indexing scheme that bridges training and inference.

\myparagraph{Rolling KV cache.}
\label{sec:method:kvcache}
The sampling scheme of DMD2 describes the denoising of a single chunk. To generate longer videos, each new chunk must attend to previously generated chunks for temporal coherence. A naive KV cache that stores keys and values for all previously generated chunks would grow linearly with video length. We instead use a \emph{rolling} cache of fixed size \upd{matching our sliding window attention training}: %
\begin{enumerate}[itemsep=0.2em]
    \item the \textbf{first chunk} is kept as a fixed anchor that  subsequent chunks always attend to. This mirrors the \emph{attention sink} mechanism used to stabilize streaming language models~\citep{xiao2024efficientstreaminglanguagemodels} and also often adopted for streaming long-video generation~\citep{liu2025rollingforcingautoregressivelong};
    \item the \textbf{latest $n$ chunks} ($n = 1$ in our setting), which provide the immediate temporal context for local motion continuity. We keep a single past chunk ($n=1$): enlarging the window to $n=2$ \upd{does not improve quality overall %
    }
    while increasing cache size and lowering throughput (supplementary material, \hyperref[tab:abl_window]{Table \ref{tab:abl_window}}).
\end{enumerate}
All other chunks are evicted from the cache once they fall outside this window,
which yields constant inference memory regardless of video duration. Moreover,
thanks to the per-chunk mixed-timestep training described in
\hyperref[sec:method:model]{Section~\ref{sec:method:model}}, cached entries only
need to store keys and values at the lowest noise level reached at the last
denoising step of the chunk, rather than at every diffusion step. The same
attention pattern is reproduced at training time by the block-causal mask
(\hyperref[fig:attmask]{Figure~\ref{fig:attmask_mask}}), in a single forward pass
and without maintaining a KV cache.

\myparagraph{RoPE indexing.}
\label{sec:method:rope}
At inference, generated-frame positions grow without bound, while the reference frame is anchored at a fixed temporal RoPE position $10$ to $40$ latents ahead of the current generation front at training time (\hyperref[sec:method:prelim:TalkVerse]{Section ~\ref{sec:method:prelim:TalkVerse}} and supplementary material \ref{sec:supmat:prelim:TalkVerse}). To preserve this training-time configuration, we explore a scheme to re-index the reference every $30$ new generated latent frames\upd{, in the spirit of the re-indexed reference or sink frames of recent streaming avatar models~\cite{huang2026liveavatarstreamingrealtime, li2026joystreamerflashrealtimeinfiniteaudiodriven, cui2026avatarforcingonestepstreamingtalking}}, shifting it by $+30$ each time so that its offset relative to the generation front cycles between $+40$ and $+10$ (\hyperref[fig:rope]{Figure ~\ref{fig:rope}} in supplementary material). This keeps the relative position of the reference within the regime the model was trained on, and avoids the position drift that would otherwise accumulate over long autoregressive rollouts.  
\upd{In practice, we find that this reindexing design helps stabilizing long generation for AURA 5B. For the smaller AURA 1.3B however, it is more beneficial to directly cap the RoPE positions of the generated frames to the range seen in training in our experiments\upd{, as Anchor Forcing~\cite{yang2026anchorforcing} and StreamAvatar~\cite{sun2026streamavatar} do} (supplementary material~\ref{sec:supmat:aura13b}).}

\subsection{Synthetic Talking-Head Dataset}
\label{sec:method:dataset}
We find that existing audio-driven video datasets are ill-suited to our training setup: the TalkVerse training set is released only as YouTube URLs and is difficult to obtain reliably, while the Hallo3 dataset~\cite{cui2025hallo3highlydynamicrealistic} contains clips too short to match our teacher's single-chunk generation window of $\sim 5$\,s. We instead build a synthetic dataset by running the TalkVerse teacher on conditioning inputs we control, and release it to support Stage~1 ODE teacher forcing (\hyperref[sec:method:stage1]{Section \ref{sec:method:stage1}}) and future work on audio-driven distillation.

\myparagraph{Conditioning inputs.} We draw reference images and their associated text prompts from Hallo3 dataset~\citep{cui2025hallo3highlydynamicrealistic}, and pair each (image, prompt) to a synthetic speech track: We first obtain textual English audio transcripts of Youtube videos provided in the LiveCC dataset~\cite{livecc}. Then, we generate speech from a subset of these transcript using an off-the-shelf text-to-speech model~\cite{dsm}. The resulting clips average $4.8$\,s, matching the teacher's chunk size.

\myparagraph{Video generation.} For each (image, prompt, audio) triplet, we run the TalkVerse teacher at $720$p and $25$\,Hz with $50$ ODE steps and a guidance scale of $4.5$. Reference image quality is heterogeneous across Hallo3, which also leads to diverse quality in the resulting video clips; No quality filtering is applied. 
For each generated video, we also store the teacher's latent ODE trajectory at five timesteps: the initial noise ($t=1$), three intermediate states ($t \in \{0.75, 0.5, 0.25\}$), and the final clean latent ($t=0$). Latents follow the TalkVerse VAE compression ($4{\times}16{\times}16$ over time and space). We additionally store the decoded clean videos as MP4 files alongside the original audio tracks.

\myparagraph{Release.} The final dataset totals $30{,}000$ clips ($38.7$\,h of synthetic videos, compared to $\sim 70$\,h for Hallo3 and CelebV). We release the audio tracks, decoded videos, and the five-timestep latent ODE trajectories to foster reproducibility and future work on talking heads generation. %
\upd{TalkVerse is only available at $5$B parameters. Since its training code is open source, we reuse its training pipeline to train our own $1.3$B bidirectional teacher on a smaller Wan2.1 backbone (Sec.~\ref{sec:exp:setup}), which serves as the teacher of AURA 1.3B. We also release its weights and the $30$K ODE trajectories it generated for Stage~1 of AURA 1.3B (Sec.~\ref{sec:supmat:aura13b}).}

%% file: sections/05_experiments.tex
\subsection{Experimental Setup}
\label{sec:exp:setup}

\paragraph{Architecture and initialization.} \upd{We train two students with the same recipe, building on the open-source TalkVerse codebase for training and inference.
\emph{AURA 5B} is initialized from its teacher, TalkVerse~\cite{wang2025talkversedemocratizingminutelongaudiodriven} ($5$B parameters, built on Wan2.2 $5$B).
\emph{AURA 1.3B} is initialized from a $1.3$B bidirectional speech-to-video teacher that we train ourselves with the TalkVerse training pipeline, adapting Wan2.1 $1.3$B text-to-video~\cite{wan2025wanopenadvancedlargescale} to speech as TalkVerse does from Wan2.2 $5$B, with LoRA adapters on the backbone and audio cross-attention layers. This teacher differs from TalkVerse in two ways. It is trained on much less data: $31$\,h of real Hallo3 video ($23$K clips), whereas TalkVerse is trained for about one epoch on its $6.3$K-hour dataset. Its audio features come from the Mimi speech codec~\cite{defossez2024moshi} instead of Wav2Vec2: Mimi is a more recent codec with a causal, streaming encoder, and we keep its first, semantic level (Sec.~\ref{sec:supmat:aura13b} of the supplementary material).
Both students are trained with LoRA adapters~\cite{hu2021loralowrankadaptationlarge} of rank $128$ on the attention layers. We additionally unfreeze their timestep embedding and output head, since our models use the per-chunk mixed-timestep scheme of \hyperref[sec:method:model]{Section~\ref{sec:method:model}} rather than the single shared timestep of the teachers. Both generate video in chunks of $n_c = 4$ latent frames with $4$ diffusion denoising steps per chunk. They differ in output resolution and rate: AURA 5B generates $720$p video ($1216\times768$ or $960\times960$, depending on the aspect ratio of the reference) played back at $25$\,Hz, the native output rate of TalkVerse, and AURA 1.3B generates $576$p video ($768\times512$ or $576\times576$, likewise) played back at $16$\,Hz, the native rate of Wan2.1; we use \emph{fps} (frames per second) elsewhere to denote throughput. Unless stated otherwise, experiments use AURA 5B.}

\myparagraph{Two-stage LoRA training.} \upd{Each stage trains its own LoRA adapters. At the end of Stage~1 (ODE teacher forcing), its LoRA is merged into the base weights, and Stage~2 (DMD with backward simulation) trains a fresh LoRA of the same rank, avoiding stacked adapters. The Stage~2 critic is initialized from the teacher, and only its LoRA adapters are trained. Both models use batch size $256$ and $5$ critic updates per student update. AURA 5B runs Stage~1 for $120$K steps at learning rate \mbox{1e-5} and Stage~2 for $5$K steps at \mbox{3.5e-7}, on $64$ H100 GPUs. AURA 1.3B runs Stage~1 for $40$K steps at \mbox{1e-5}, on trajectories of its own teacher, and Stage~2 for $6.5$K steps at \mbox{2e-7}, on $32$ H100 GPUs. Other hyperparameters, including the 1.3B teacher's, are in the supplementary material (Sec.~\ref{sec:supmat:params}, \ref{sec:supmat:aura13b}).}

\label{sec:exp:baselines}
\myparagraph{Baselines.} We compare AURA against four open-source audio-driven avatar generation models. \textit{TalkVerse}~\cite{wang2025talkversedemocratizingminutelongaudiodriven} , our teacher, serves as a high-quality reference for the distillation. \textit{Wan-S2V 14B}~\cite{gao2025wans2vaudiodrivencinematicvideo} and \textit{Hallo3}~\cite{cui2025hallo3highlydynamicrealistic} are two recent large-scale non-causal S2V model. %
\textit{Live Avatar}~\cite{huang2026liveavatarstreamingrealtime} is a streaming few-step approach and the closest competitor to AURA in inference regime. 
For all baselines we use the authors' released code and checkpoints, and run inference under each method's recommended settings.

\myparagraph{Benchmarks.} We evaluate on two test sets we generate.
\textsc{Hallo3 test} consists of $100$ first frames and prompts randomly sampled
from the Hallo3 test set~\cite{cui2025hallo3highlydynamicrealistic}, each re-paired
with a real English speech track from an internal dataset to assess speaker
generalization; all clips are $10$\,s. \textsc{CelebV-HQ test} consists of $150$
first frames from CelebV-HQ~\cite{zhu2022celebvhqlargescalevideofacial}, each paired
with $8$--$20$\,s of TTS-synthesized speech (same pipeline as Sec.~\ref{sec:method:dataset})
and the generic prompt \emph{``a person is talking''}; we synthesize rather than
reuse audio because CelebV-HQ clips are too short ($\sim$$5$\,s) to test longer
sequences.

\myparagraph{Metrics.} We evaluate along five axes. \textit{Visual quality:}
FID~\citep{heusel2018ganstrainedtimescaleupdate},
FVD~\citep{unterthiner2019accurategenerativemodelsvideo} on $16$-frame windows,
\upd{computed} at three temporal offsets, @$0$ (initialization), @$101$ and @$109$
(windows centered on and starting at a mid-video chunk boundary, capturing
inter-chunk discontinuities), and JEDi~\citep{luo2024fvdenhancedevaluationmetrics},
an MMD-based distance on V-JEPA features more reliable at our
limited sample counts, at the same offsets. \upd{The main tables report the mean over the three offsets, and the per-offset values are in the supplementary material (Sec.~\ref{sec:supmat:perofs}). (The $16$-frame windows span $1$\,s at $16$\,Hz and $0.64$\,s at $25$\,Hz, so FVD and JEDi are only approximately comparable across output rates.)} \textit{Temporal consistency:} the warping error (WE), the
mean L1 residual after RAFT~\citep{teed2020raftrecurrentallpairsfield} optical-flow
alignment of consecutive frames, capturing flicker and chunk-boundary artifacts.
\textit{Audio--visual alignment:} Sync-C and Sync-D~\citep{Chung16a} (SyncNet).
\textit{Identity preservation:} CSIM, the mean cosine similarity between
ArcFace~\citep{Deng_2022} embeddings of the reference and generated frames.
\textit{Inference timing:} on a single H$100$ GPU, generating one clip from a fixed
$10$\,s English audio and a $960\times960$ reference at each model's native
resolution and frame rate; we report throughput (fps) for the \emph{DiT} and
\emph{VAE} decoder separately and the end-to-end \emph{Total} (excluding loading
and warm-up). Total is not a simple composition of the two: for Live Avatar,
offloading the DiT or VAE between calls to fit in GPU memory adds overhead.

\subsection{Main Results}
\label{sec:exp:main}
\begin{table*}[t]
\centering
\scriptsize
\setlength{\tabcolsep}{3pt}
\renewcommand{\arraystretch}{1.05}
\caption{\textbf{Main results on \textsc{Hallo3 test}}. FVD and JEDi are \upd{averaged over the three temporal offsets ($@0$, $@101$, $@109$); per-offset breakdowns are in the supplementary material (Tab.~\ref{tab:hallo3_full}).} \emph{WE}: warping error. \emph{DiT / VAE}: throughput of the diffusion transformer and VAE decoder alone, in frames per second on a single H$100$ GPU. \emph{Total} (fps): end-to-end throughput, including model offload and overhead. \textbf{Bold}: best; \underline{underlined}: second best. \upd{AURA 1.3B outputs $480$p at $16$\,Hz (like Wan-S2V and Live Avatar), the other AURA and TalkVerse rows $25$\,Hz.}%
}
\label{tab:hallo3_main}
\resizebox{\textwidth}{!}{%
\begin{tabular}{l c c c c c c c c c c c c}
\toprule
Method & Params & Res & Steps & DiT / VAE (fps)$\uparrow$ & Total (fps)$\uparrow$ & FID$\downarrow$ & \upd{FVD}$\downarrow$ & \upd{JEDi}$\downarrow$ & CSIM$\uparrow$ & Sync-C$\uparrow$ & Sync-D$\downarrow$ & WE$\downarrow$ \\
\midrule
Hallo3        & 14.5B & 480p & 50 & $0.24 / 3.2$    & $0.22$  & \upd{\textbf{42.8}}    & \upd{183.4} & \upd{\textbf{0.180}}    & 0.74          & \underline{5.06} & 9.9          & 0.0036 \\
Wan-S2V       & 14B   & 720p & 40 & $0.08 / 6.7$    & $0.08$  & \upd{57.7}             & \upd{196.4} & \upd{\underline{0.257}} & 0.71    & \upd{\textbf{5.42}}   & \textbf{9.0} & \underline{0.0032} \\
TalkVerse     & 5B    & 720p & 50 & $0.69 / 7.4$    & $0.58$  & \upd{58.5}             & \upd{210.8} & \upd{0.398}                               & 0.74          & 3.82            & 10.5         & \textbf{0.0019} \\
Live Avatar   & 14B   & 480p & 4  & $8.4 / 4.3$     & $1.9$   & \upd{\underline{52.0}} & \upd{\underline{181.1}} & \upd{0.442}                               & \upd{0.77} & 4.90         & \underline{9.3} & 0.0039 \\
\midrule
\textbf{AURA \upd{5B} (ours)}    & \textbf{5B} & \textbf{720p} & \textbf{4} & \upd{$\underline{19.7} / \underline{22.7}$} & \upd{$\underline{9.7}$} & \upd{56.9} & \upd{195.5} & \upd{0.313} & \upd{\underline{0.79}} & \upd{4.30} & \upd{9.87} & \upd{\underline{0.0032}} \\
\upd{\textbf{AURA 1.3B (ours)}} & \upd{\textbf{1.3B}} & \upd{\textbf{480p}} & \upd{\textbf{4}} & \upd{$\mathbf{28.8 / 96.9}$} & \upd{$\mathbf{20.7}$} & \upd{56.8} & \upd{\textbf{177.1}} & \upd{0.403} & \upd{\textbf{0.84}} & \upd{4.07} & \upd{9.93} & \upd{0.0036} \\
\bottomrule
\end{tabular}

}
\end{table*}

\begin{table*}[t]
\centering
\scriptsize
\setlength{\tabcolsep}{3pt}
\renewcommand{\arraystretch}{1.05}
\caption{\textbf{Main results on \textsc{CelebV-HQ test}}. FVD and JEDi are averaged over the three temporal offsets ($@0$, $@101$, $@109$); per-offset breakdowns are in supplementary material. Columns and naming conventions follow the same as in \hyperref[tab:hallo3_main]{Table ~\ref{tab:hallo3_main}}.%
}
\label{tab:celebv_main}
\resizebox{\textwidth}{!}{%
\begin{tabular}{l c c c c c c c c c c c c}
\toprule
Method & Params & Res & Steps & DiT / VAE (fps)$\uparrow$ & Total (fps)$\uparrow$ & FID$\downarrow$ & FVD$\downarrow$ & JEDi$\downarrow$ & CSIM$\uparrow$ & Sync-C$\uparrow$ & Sync-D$\downarrow$ & WE$\downarrow$ \\
\midrule
Hallo3         & 14.5B & 480p & 50 & $0.24 / 3.2$ & $0.22$ & \upd{\textbf{39.1}} & \upd{\textbf{202.5}} & \upd{\textbf{0.243}} & \upd{0.71} & \upd{\underline{6.23}} & 9.4  & \textbf{0.0038} \\
Wan-S2V        & 14B   & 720p & 40 & $0.08 / 6.7$ & $0.08$ & \upd{\underline{50.4}} & \upd{289.6} & \upd{0.577} & 0.62 & \upd{\textbf{6.24}} & \textbf{8.9}  & 0.0061 \\
TalkVerse      & 5B    & 720p & 50 & $0.69 / 7.4$ & $0.58$ & \upd{53.2} & \upd{255.9} & \upd{0.582} & 0.59 & \upd{4.78} & 10.5 & 0.0059 \\
Live Avatar    & 14B   & 480p & 4  & $8.4 / 4.3$  & $1.9$  & \upd{52.4} & \upd{267.5} & \upd{1.010} & 0.67 & 5.90 & \underline{9.2}  & 0.0057 \\
\midrule
\textbf{AURA \upd{5B} (ours)}     & \textbf{5B} & \textbf{720p} & \textbf{4} & \upd{$\underline{19.7} / \underline{22.7}$} & \upd{$\underline{9.7}$}  & \upd{54.0} & \upd{\underline{208.1}} & \upd{\underline{0.320}} & \upd{\underline{0.72}} & \upd{4.94} & \upd{10.0} & \upd{\underline{0.0043}} \\
\upd{\textbf{AURA 1.3B (ours)}} & \upd{\textbf{1.3B}} & \upd{\textbf{480p}} & \upd{\textbf{4}} & \upd{$\mathbf{28.8 / 96.9}$} & \upd{$\mathbf{20.7}$} & \upd{51.5} & \upd{248.8} & \upd{0.574} & \upd{\textbf{0.77}} & \upd{4.76} & \upd{10.1} & \upd{0.0046} \\
\bottomrule
\end{tabular}
}
\end{table*}

\myparagraph{Quality and identity.} In \hyperref[tab:hallo3_main]{Table~\ref{tab:hallo3_main}} and~\ref{tab:celebv_main} we report our main quantitative comparison against all baselines, alongside model size, output resolution, and number of diffusion steps. Despite using only $4$ denoising steps instead of $50$, AURA matches or improves on its TalkVerse teacher across most automatic metrics, and is comparable to the much larger Wan-S2V ($14$B, multi-step) on FVD, JEDi, and audio--visual alignment. On identity preservation (CSIM), AURA outperforms all baselines on both test sets (\upd{$0.79$ for AURA 5B and $0.84$ for AURA 1.3B} on \textsc{Hallo3 test}, \upd{$0.72$ and $0.77$} on \textsc{CelebV-HQ test}), which we attribute to \upd{our RoPE indexing schemes (\hyperref[sec:method:rope]{Sec.~\ref{sec:method:rope}})} and to keeping the initial generated chunks in the rolling cache, both of which limit identity drift over long generations.

\myparagraph{Speed.} \upd{Both AURA models are faster than all baselines, including Live Avatar, which is also a few-step model.} Throughputs are reported in raw frames per second. AURA\upd{ 5B}, TalkVerse and Hallo3 generate at $25$\,Hz, Live Avatar\upd{,} Wan-S2V \upd{and AURA 1.3B} at $16$\,Hz. Measured in generated seconds of video per second of compute, AURA's lead over the latter two is larger
still. \upd{AURA 5B runs the end-to-end pipeline at $9.7$\,fps on a single H$100$, a $\mathbf{\sim 17\times}$ speedup over its TalkVerse teacher ($0.58$\,fps); its diffusion transformer and VAE decoder alone reach $19.7$ and $22.7$\,fps (on the DiT, a $\mathbf{\sim 30\times}$ speedup over the teacher's $0.69$\,fps), and the end-to-end throughput is bounded by their sequential composition. AURA 1.3B runs at $20.7$\,fps end-to-end ($28.8$\,fps for the DiT, $96.9$\,fps for the VAE decoder), faster than its $16$\,Hz playback rate: it streams in real time on a single H$100$, a $\mathbf{\sim 23\times}$ speedup over its bidirectional teacher ($0.90$\,fps, measured on a single $4.8$\,s window since this teacher generates fixed-length clips). These timings use the square reference of our protocol ($576\times576$ output); at $768\times512$, AURA 1.3B runs at $17.3$\,fps end-to-end, still faster than real time.}
\hyperref[fig:teaser]{Figure~\ref{fig:teaser}} (right) summarizes the quality--speed trade-off: AURA matches the best baselines on quality while being one to two orders of magnitude faster. \hyperref[fig:bubble_plot]{Figure~\ref{fig:bubble_plot}} in the supplementary material adds complementary visualizations.

\myparagraph{Long-form streaming stability.} A central claim of AURA is that its rolling KV cache sustains generation over an unbounded stream at constant memory. \upd{To measure how quality and identity evolve over time, we run both models ($4$ steps) on the $100$ \textsc{Hallo3 test} references driven by $60$\,s of speech and compute metrics on consecutive $10$\,s segments (\hyperref[fig:long]{Figure \ref{fig:long}} and \hyperref[tab:long_eval_supp]{Table \ref{tab:long_eval_supp}} in the supplementary material). For AURA 1.3B, FVD and lip-sync stay flat over the six segments (FVD $179$--$186$, Sync-C $4.73$--$4.81$, no segment significantly different from the first), and so does identity for most videos (median per-video CSIM change $-0.02$ over the minute; a few videos drift and pull the mean from $0.83$ to $0.78$). AURA 5B degrades slightly after the first segment (FVD from $182$ to $195$--$206$, then stable) and its identity drifts slightly (CSIM from $0.76$ to $0.69$).}

\begin{figure}[t]
  \centering
  \includegraphics[width=0.5\columnwidth]{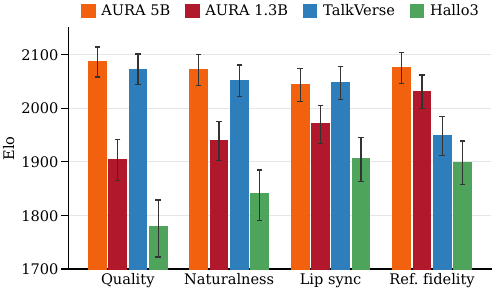}
  \caption{\textbf{Human preference study.} Per-axis Bayesian Elo ($\uparrow$) from
  \upd{$400$ pairwise comparisons by $27$ raters between AURA 5B, AURA 1.3B, TalkVerse and Hallo3} on the \textsc{Hallo3 test} clips;
  error bars are $95\%$ CIs.}
  \label{fig:userstudy}
  \vspace{-10pt}
\end{figure}

\myparagraph{Perceptual quality.} Distribution-based metrics (FID, FVD, JEDi) capture global feature statistics but do not penalize certain local artifacts: Hallo3 obtains the best FID, FVD, and JEDi on both test sets, yet its generations exhibit exaggerated mouth motion and recurrent hand artifacts that these metrics fail to flag (qualitative analysis in \hyperref[fig:qual_hallo3]{Fig.~\ref{fig:qual_hallo3}}, supplementary material). To assess perceived quality directly, we run an anonymous human evaluation on the Mabyduck platform. We collected \upd{$400$ pairwise comparisons from $27$ raters on the $100$ \textsc{Hallo3 test} clips, covering all pairs among AURA 5B, AURA 1.3B, Hallo3 and our TalkVerse teacher,} along four axes: overall visual quality, lip-sync, reference fidelity, and naturalness. Results are  aggregated into per-axis Bayesian Elo scores (\hyperref[fig:userstudy]{Fig.~\ref{fig:userstudy}}). \upd{Both AURA models are} preferred over Hallo3 on \emph{all four} axes, confirming that
Hallo3's distribution-metric lead does not translate into perceived quality. Against the non-causal $50$-step teacher, AURA \upd{5B} ranks \emph{higher} on reference fidelity, consistent with our CSIM results, \upd{and is on par on visual quality, lip-sync and naturalness,} despite being causal and $\sim\!17\times$ faster. \upd{AURA 1.3B also ranks above TalkVerse on reference fidelity, but below AURA 5B and TalkVerse on visual quality, which reflects its lower resolution, as well as on lip-sync and naturalness.}

\subsection{Ablations}
\label{sec:exp:ablations}

We ablate the main design choices of our method. Unless stated otherwise, all ablations use the 5B student trained on the full pipeline and are evaluated on the \textsc{Hallo3 test} set with \upd{the same $10$\,s clips as the main results (Sec.~\ref{sec:exp:main}); the $4$-step default row of each ablation is the AURA 5B row of Tab.~\ref{tab:hallo3_main}. We add two diagnostics, defined in Sec.~\ref{sec:supmat:diag} of the supplementary material: \emph{Sharp.}, the variance of the Laplacian of the aligned face (frame-level sharpness proxies are dominated by background and noise), and \emph{$\Delta$Col.}, a within-clip colour drift: how much further the mean colour of the face moves away from that of the reference face (distance in CIELAB) between the first ($0.5$--$2.5$\,s) and last ($7.5$--$9.5$\,s) two seconds.}

\myparagraph{Number of diffusion steps.}
\hyperref[tab:abl_steps]{Table~\ref{tab:abl_steps}} reports quality metrics when generating each chunk in $\{1, 2, 3, 4\}$ diffusion steps. \upd{Visually, videos become blurrier as the number of steps decreases (Sec.~\ref{sec:exp:qualitative}), which distribution metrics barely capture: FVD and JEDi are best at $4$ steps and degrade only slightly at $3$ and $2$ steps, FID is even marginally better at $3$ and $2$ steps, and all drop clearly at $1$ step. Face sharpness makes the blur explicit: it decreases monotonically from $50.4$ at $4$ steps to $37.5$, $36.5$ and $27.0$ at $3$, $2$ and $1$ steps. Identity (CSIM) and lip-sync remain essentially stable down to $2$ steps; the lower warping error with fewer steps reflects the same blur, which leaves fewer details to flag as inconsistent across frames, rather than a real improvement.}
We adopt $4$ steps in all other experiments as the best trade-off between per-frame quality and inference speed.

\begin{table}[t]
\centering
\small
\caption{\textbf{Effect of the number of diffusion steps} per chunk on quality and speed. DiT throughput is reported in frames per second on a single H$100$. FVD and JEDi are averaged over three temporal offsets ($@0$, $@101$, $@109$). Other experiments use $4$ steps. Best in \textbf{bold}.}
\setlength{\tabcolsep}{4pt}
\resizebox{\columnwidth}{!}{%
\begin{tabular}{lccccccccc}
\toprule
\#Steps & FID $\downarrow$ & FVD $\downarrow$ & JEDi $\downarrow$ & CSIM $\uparrow$ & Sync-C $\uparrow$ & Sync-D $\downarrow$ & WE $\downarrow$ & \upd{Sharp. $\uparrow$} & DiT (fps) $\uparrow$ \\
\midrule
4 & \upd{56.9} & \upd{\textbf{195.5}} & \upd{\textbf{0.31}} & \upd{\textbf{0.79}} & \upd{4.30} & \upd{9.87} & \upd{0.0032} & \upd{\textbf{50.4}} & 19.7 \\
3 & \upd{\textbf{55.4}} & \upd{198.8} & \upd{0.33} & 0.77 & \upd{\textbf{4.39}} & \upd{9.77} & \upd{0.0031} & \upd{37.5} & 26.0 \\
2 & \upd{55.5} & \upd{198.2} & \upd{0.37} & \upd{\textbf{0.79}} & \upd{\textbf{4.39}} & \upd{9.81} & \upd{0.0031} & \upd{36.5} & 38.5 \\
1 & \upd{58.6} & \upd{229.2} & \upd{0.66} & 0.65 & \upd{4.26} & \upd{\textbf{9.71}} & \upd{\textbf{0.0030}} & \upd{27.0} & \textbf{75.8} \\
\bottomrule
\end{tabular}}

\label{tab:abl_steps}
\end{table}

\myparagraph{Stage 1 only vs.\ Stage 1 + Stage 2.} Table~\ref{tab:abl_stages} compares a student trained only with ODE teacher forcing (Stage~1) against the full two-stage pipeline (Stage~1 + Stage~2). Adding the DMD stage substantially improves all perceptual quality metrics (FID, FVD, JEDi) and identity preservation (CSIM), confirming that Stage~1 alone is insufficient to close the gap to the teacher in the few-step regime. Lip-sync metrics also improve. As in the previous ablation, the temporal consistency metric (Warp) reaches lower value with Stage~1 only, but this reflects the same metric artifact on blurrier outputs rather than a real improvement\upd{: face sharpness is almost three times lower with Stage~1 only ($18.3$ vs.\ $50.4$).}

\begin{table}[ht]
\centering
\small
\caption{\textbf{Effect of Stage~2 on generation quality}. Adding the DMD distillation stage substantially improves perceptual quality (FID, FVD, JEDi), identity preservation (CSIM), and lip-sync. Warp reaches lower values with Stage~1 only, but this is the same metric artifact on blurrier outputs as observed in Tab.~\ref{tab:abl_steps} (see Sec.~\ref{sec:exp:qualitative}). Best value per column in \textbf{bold}.}
\setlength{\tabcolsep}{4pt}
\begin{tabular}{lcccccccc}
\toprule
Variant & FID $\downarrow$ & FVD $\downarrow$ & JEDi $\downarrow$ & CSIM $\uparrow$ & Sync-C $\uparrow$ & Sync-D $\downarrow$ & WE $\downarrow$ & \upd{Sharp. $\uparrow$} \\
\midrule
Stage 1 only & \upd{68.3} & \upd{264.8} & \upd{0.58} & 0.55 & \upd{3.55} & \upd{10.07} & \upd{\textbf{0.0023}} & \upd{18.3} \\
Stage 1 + Stage 2 & \upd{\textbf{56.9}} & \upd{\textbf{195.5}} & \upd{\textbf{0.31}} & \upd{\textbf{0.79}} & \upd{\textbf{4.30}} & \upd{\textbf{9.87}}  & \upd{0.0032} & \upd{\textbf{50.4}} \\
\bottomrule
\end{tabular}
\label{tab:abl_stages}
\vspace{-10pt}
\end{table}

\begin{figure*}[t]
    \centering
    \makebox[0.18\linewidth][c]{}\hfill
    \makebox[0.78\linewidth][s]{%
        \makebox[0.255\linewidth][c]{\small $t \approx 1$\,s}\hfill
        \makebox[0.255\linewidth][c]{\small $t \approx 3$\,s}\hfill
        \makebox[0.255\linewidth][c]{\small $t \approx 5$\,s}%
    }
    \\[2pt]
    \makebox[0.18\linewidth][c]{}\hfill\makebox[0.78\linewidth][c]{\small\upd{\textbf{AURA 5B}}}
    \\[1pt]
    \begin{minipage}[c]{0.18\linewidth}
        \centering
        \includegraphics[width=\linewidth]{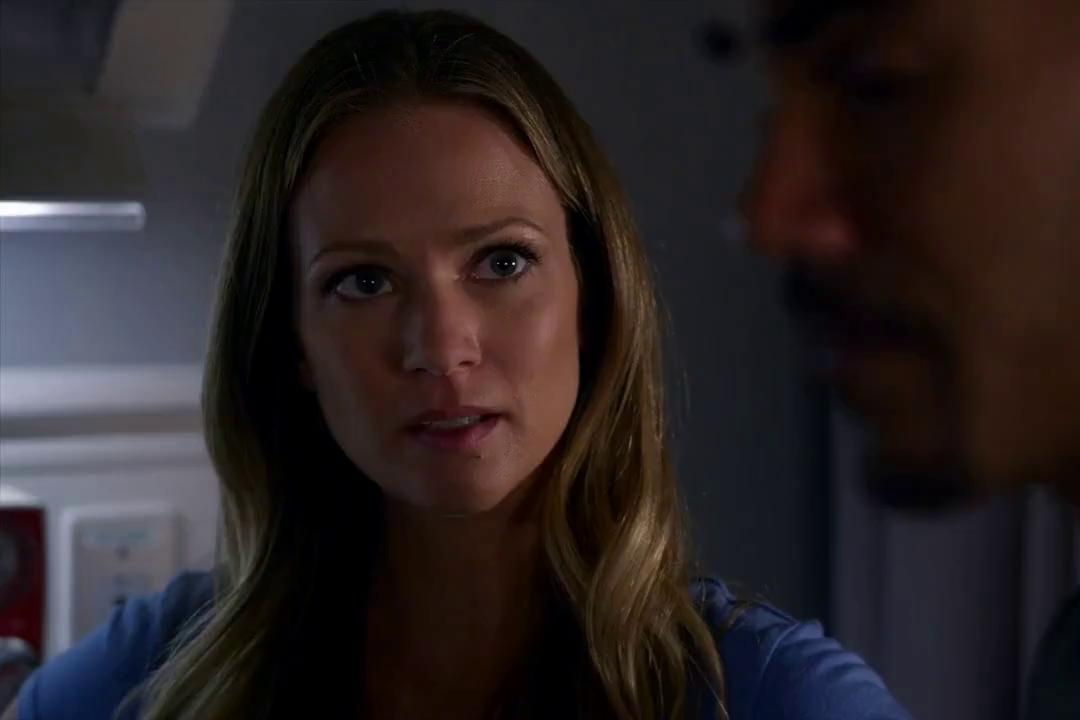}
    \end{minipage}\hfill
    \begin{minipage}[c]{0.78\linewidth}
        \centering
        \includegraphics[width=0.32\linewidth]{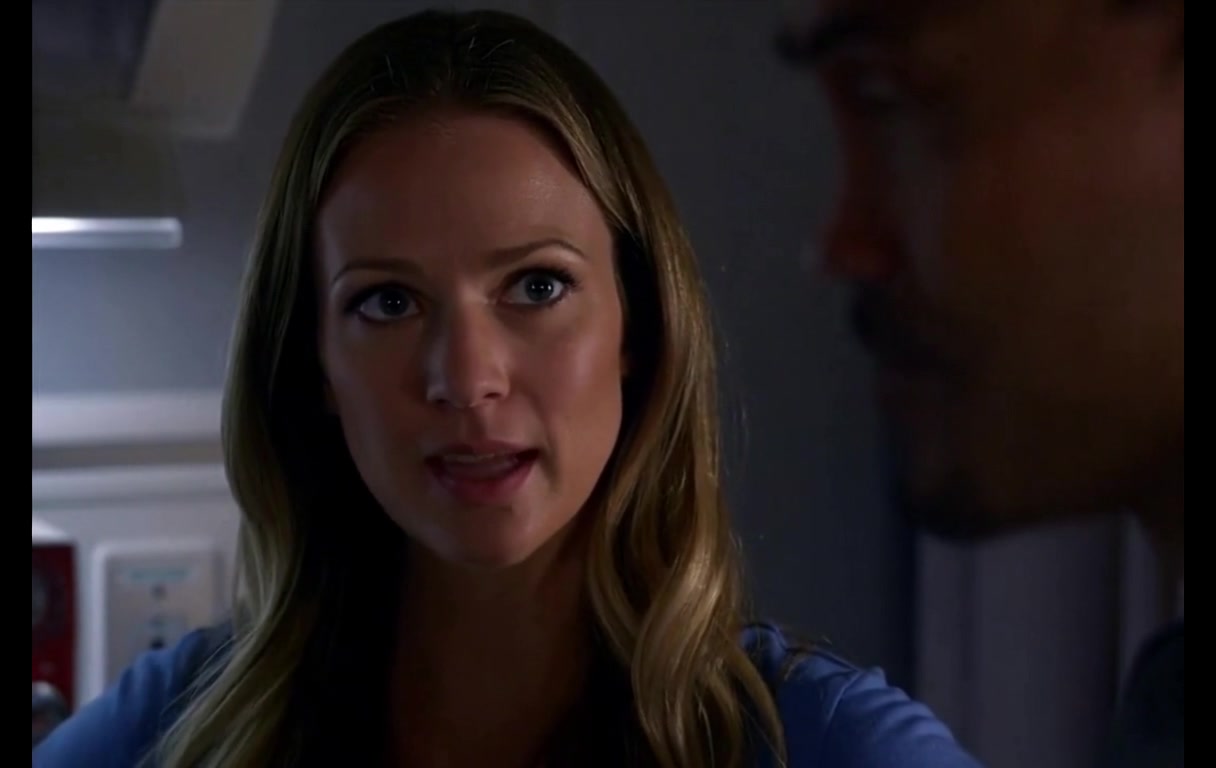}\hfill
        \includegraphics[width=0.32\linewidth]{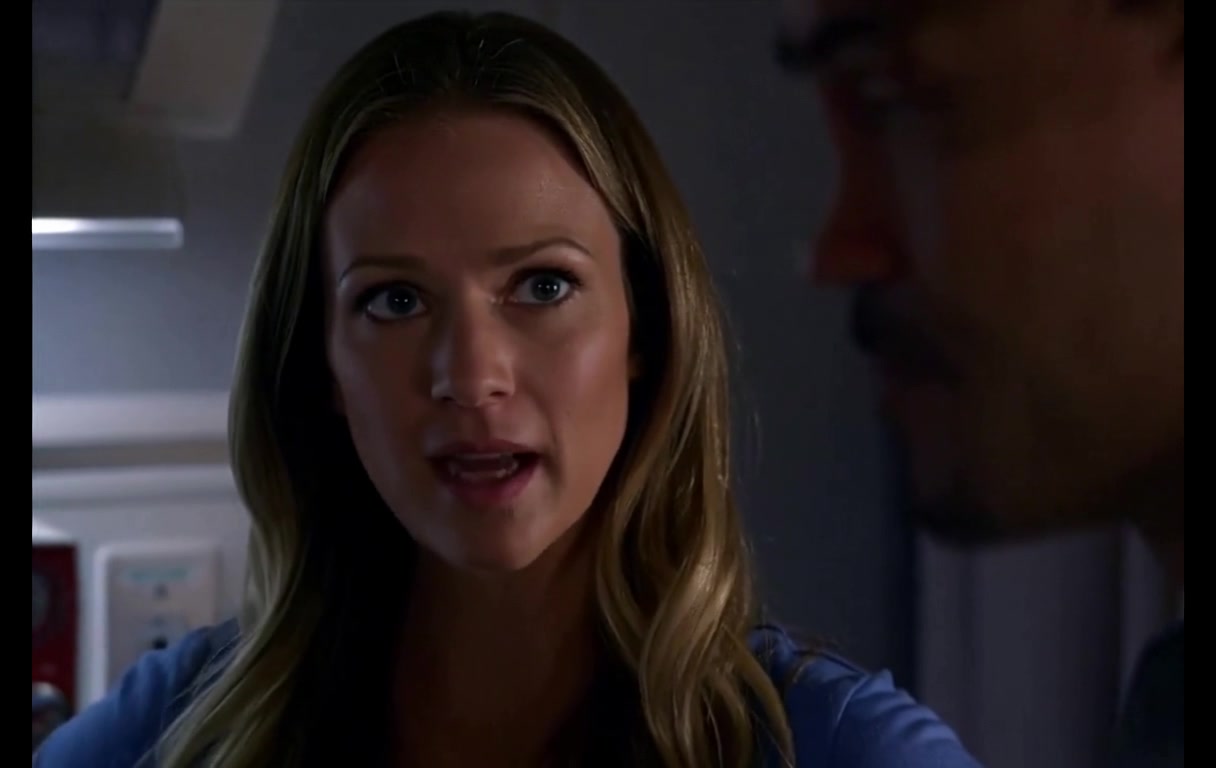}\hfill
        \includegraphics[width=0.32\linewidth]{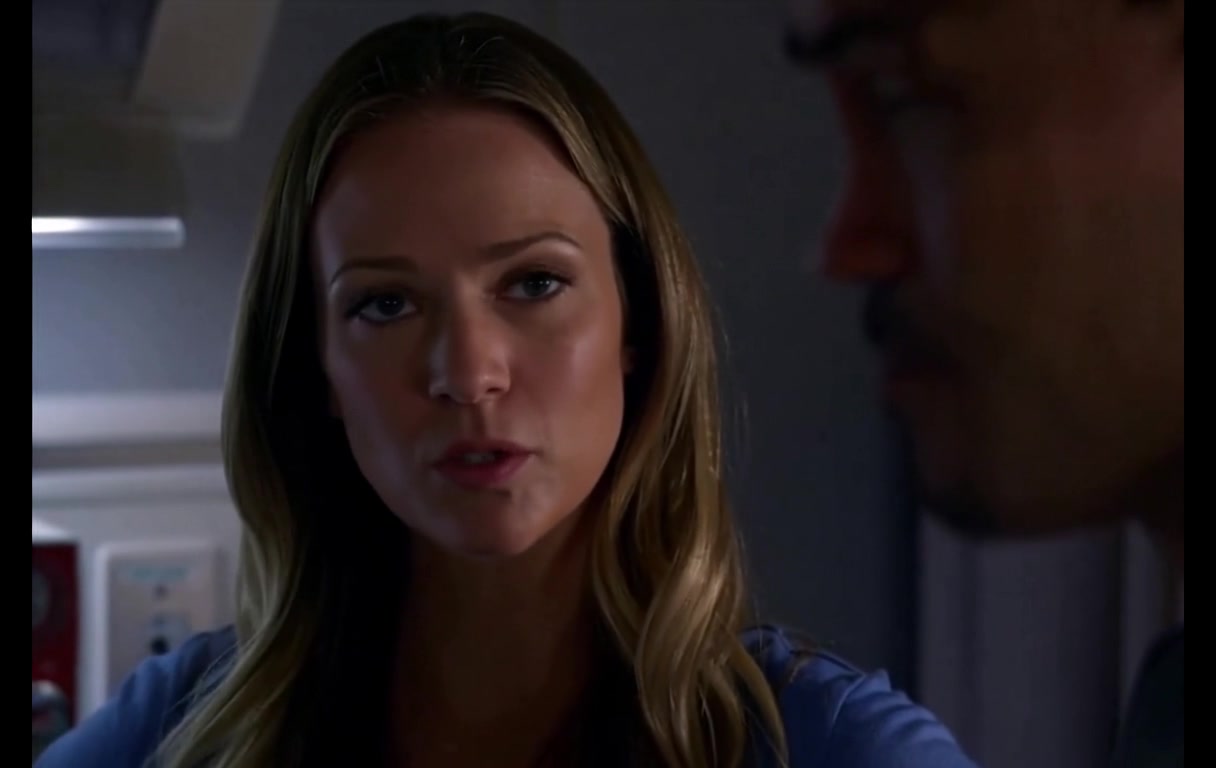}
    \end{minipage}
    \\[4pt]
\makebox[0.18\linewidth][c]{}\hfill\makebox[0.78\linewidth][c]{\small\upd{\textbf{AURA 1.3B}}}
\\[1pt]
\begin{minipage}[c]{0.18\linewidth}
    \centering
    \includegraphics[width=\linewidth]{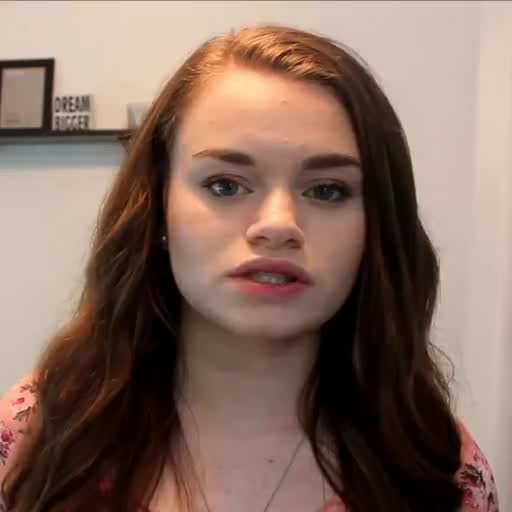}
\end{minipage}\hfill
\begin{minipage}[c]{0.78\linewidth}
    \centering
    \includegraphics[width=0.32\linewidth]{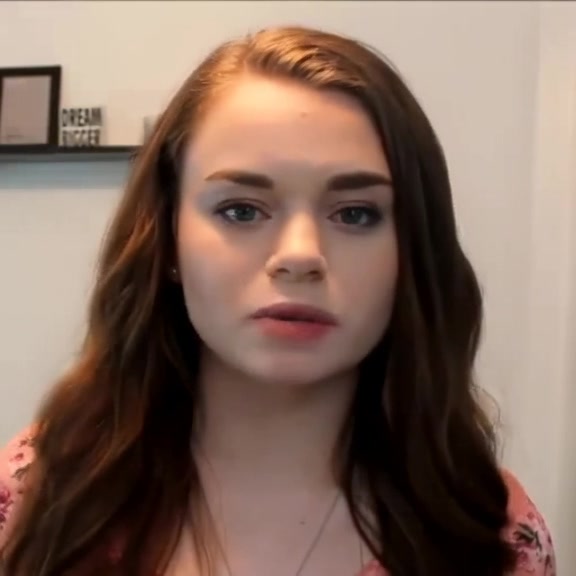}\hfill
    \includegraphics[width=0.32\linewidth]{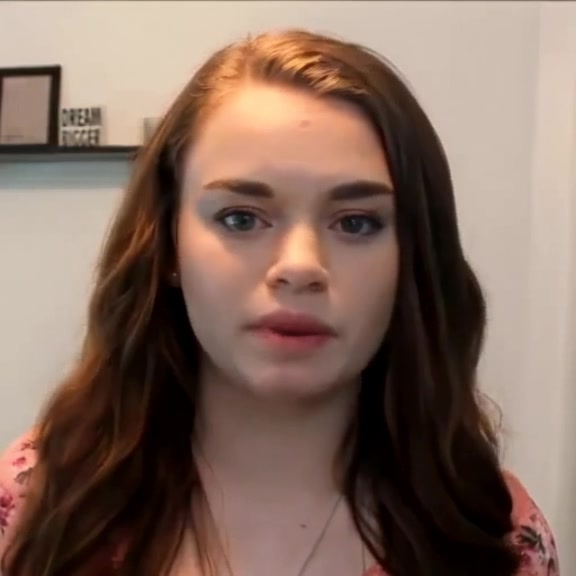}\hfill
    \includegraphics[width=0.32\linewidth]{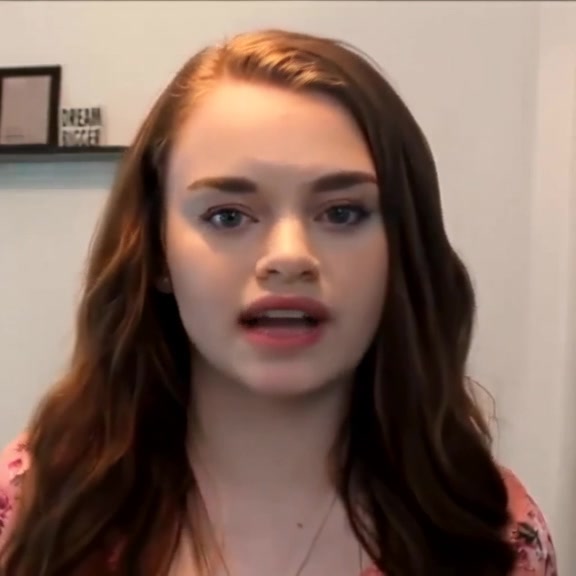}
\end{minipage}
\\[2pt]
\makebox[0.18\linewidth][c]{\small Reference}\hfill
\makebox[0.78\linewidth][c]{\small AURA generations}
    
    \caption{\textbf{Qualitative results from AURA} on two diverse test sequences with different audios\upd{, generated by AURA 5B (top) and AURA 1.3B (bottom)}. Each row shows the reference identity image (left) and three frames sampled from the generated video at approximately $t = 1, 3, 5$\,s. The samples illustrate the diversity of inputs AURA handles, including varied lighting, head poses, and identities. \upd{Figure~\ref{fig:teaser} shows both models on the same inputs; full videos, more results and comparisons with the baselines are on the \href{https://kyutai.org/aura/}{project page}.} } 
    \label{fig:qual_aura}
    \vspace{-10pt}
\end{figure*}

\myparagraph{KV cache refresh with an additional step.} We tested three variants where an additional diffusion step is added at inference to refresh the KV cache with a less noisy state for the next chunk. Stage~1 is fine-tuned to additionally handle a low timestep $t=0$ in its ODE trajectories, and Stage~2 is retrained from scratch with this extra low timestep included in the backward simulation: $t=0$ in \textit{5-step ($t=0$ refresh)} and $t=50$ in the two \textit{5-step ($t=50$)} variants. At inference, this extra step updates the cache for the next chunk; its output is discarded in the first two variants and used as the final chunk prediction in the third. Table~\ref{tab:abl_refresh} reports the results. None of the variants consistently outperforms the simpler $4$-step default: FID and lip-sync slightly improve in some variants, but FVD, JEDi, and identity preservation are \upd{degraded, markedly for the two $t{=}50$ variants (FVD $\approx 210$ vs.\ $195.5$, CSIM $0.74$ vs.\ $0.79$). These two variants also drift within the clip: the colour distance between the face and the reference grows by $3.1$ over $10$\,s ($\Delta$Col.; $0.4$ for the default), a perceptible shift, the face getting brighter over time.}
 Given the added inference cost, we keep the $4$-step default.
\begin{table}[!htbp]
\centering
\small
\caption{\textbf{KV cache refresh strategy}. Effect of an additional inference step to refresh it. Best value per column in \textbf{bold}.}
\setlength{\tabcolsep}{4pt}
\resizebox{\columnwidth}{!}{%
\begin{tabular}{lcccccccc}
\toprule
Variant & FID $\downarrow$ & FVD $\downarrow$ & JEDi $\downarrow$ & CSIM $\uparrow$ & Sync-C $\uparrow$ & WE $\downarrow$ & \upd{$\Delta$Col. $\downarrow$} & DiT (fps) $\uparrow$ \\
\midrule
4 steps (default) & \upd{56.9} & \upd{\textbf{195.5}} & \upd{\textbf{0.31}} & \upd{\textbf{0.79}} & \upd{4.30} & \upd{\textbf{0.0032}} & \upd{\textbf{0.4}} & \textbf{19.7} \\
5-step ($t{=}0$ refresh) & \upd{57.7} & \upd{196.7} & \upd{0.33} & \upd{0.78} & \upd{4.26} & \upd{\textbf{0.0032}} & \upd{\textbf{0.4}} & 15.6 \\
5-step ($t{=}50$ refresh) & \upd{\textbf{54.5}} & \upd{210.0} & \upd{0.38} & \upd{0.74} & \upd{4.29} & \upd{0.0038} & \upd{3.1} & 15.6 \\
5-step ($t{=}50$, used) & \upd{\textbf{54.5}} & \upd{209.8} & \upd{0.37} & \upd{0.74} & \upd{\textbf{4.31}} & \upd{0.0038} & \upd{3.1} & 15.6 \\
\bottomrule
\end{tabular}}

\label{tab:abl_refresh}
\vspace{-10pt}
\end{table}

\myparagraph{Rolling-window size.} \upd{Keeping two past chunks in the rolling cache instead of one ($n=2$, Tab.~\ref{tab:abl_window} and Sec.~\ref{sec:supmat:window} in the supplementary material) slightly improves FVD ($193.2$ vs.\ $195.5$) but degrades JEDi, identity and lip-sync, while increasing cache size and lowering throughput; we keep $n=1$.}

\subsection{Qualitative Results}
\label{sec:exp:qualitative}

Figures \ref{fig:teaser}  and \ref{fig:qual_aura} show representative AURA generations\upd{, from both models,} on different test sequences spanning varied lighting conditions, head poses, and identities. The generated frames remain faithful to the reference identity throughout the sequence, and exhibit natural motion patterns -- including spontaneous eye blinks, subtle head movements, and small facial expression variations -- that go beyond strict lip motion.

Figure~\ref{fig:qual_aura_vs_tv} in supplementary material compares AURA\upd{ 5B} to its TalkVerse teacher on the same input. Despite using only $4$ denoising steps instead of $50$, AURA matches the teacher's visual quality, and in some cases improves over it: in the example shown, the TalkVerse output produces a static gaze with no blinking over the full $10$\,s clip, whereas AURA generates natural eye blinks and more varied global motion. \upd{Full videos for all sequences, including AURA 1.3B samples, are on the project page at \href{https://kyutai.org/aura/}{https://kyutai.org/aura/}.}

%% file: sections/06_conclusion.tex
We present AURA, a \upd{family of} causal few-step speech-to-video \upd{models} for streaming
talking-head generation. Distilling a bidirectional teacher into a same-size
$4$-step causal student through a two-stage pipeline and a rolling KV cache, AURA \upd{5B}
matches \upd{or improves on} its TalkVerse teacher on FVD and JEDi at a $17\times$ speedup, delivering
$720$p video at $9.7$\,fps on a single GPU. \upd{AURA 1.3B, distilled with the same pipeline from a $1.3$B teacher, streams $576$p video in real time on a single GPU ($20.7$\,fps for a $16$\,Hz playback rate).} We also release a $30{,}000$-clip
synthetic dataset with teacher ODE trajectories\upd{, as well as our $1.3$B teacher and its ODE trajectories,} to support future work on
audio-driven distillation.

%% file: sections/07_supplementary.tex
\subsection{TalkVerse}
\label{sec:supmat:prelim:TalkVerse}

We use TalkVerse~\cite{wang2025talkversedemocratizingminutelongaudiodriven} as \upd{the teacher of AURA 5B}, a 5B-parameter audio-driven video generation model built on the Wan2.2-5B~\cite{wan2025wanopenadvancedlargescale} Diffusion Transformer (DiT) backbone, originally pretrained for text-and-image-to-video generation. TalkVerse extends the Wan-S2V~\cite{gao2025wans2vaudiodrivencinematicvideo} design for speech-conditioned video synthesis to the Wan2.2 backbone, and trades the 14B-parameter scale of Wan-S2V for a smaller 5B model, trained on a high-resolution open-source dataset.

\myparagraph{Latent space.} TalkVerse uses the Wan2.2 video VAE, which downsamples by $(t,h,w)=(4,16,16)$, a higher compression ratio than the $(4,8,8)$ commonly used in prior video diffusion models (e.g.\ Wan2.1). This yields a $4\times$ reduction in latent token count at the same resolution, enabling efficient high-resolution (720p) generation.

\myparagraph{Conditioning.} The model is conditioned on three inputs: (i) a reference identity image, (ii) a speech audio track, and (iii) a text prompt describing the video content. Audio is encoded with Wav2Vec2~\cite{baevski2020wav2vec20frameworkselfsupervised} and injected into a subset of DiT blocks via cross-attention from visual tokens to audio tokens, with each latent frame attending to a temporally aligned audio segment. The text prompt is encoded with T5-XXL~\cite{raffel2023exploringlimitstransferlearning} and injected through cross-attention. TalkVerse is trained on the large-scale TalkVerse-2.3M dataset of high-resolution audio-synchronized clips. 

\myparagraph{Long-video generation: temporal context.} To generate minute-long videos, TalkVerse uses a sliding-window mechanism: each window denoises a fixed number of frames conditioned on a window of preceding frames, referred to as \emph{motion frames}. To keep the token budget tractable, the motion frames are compressed by a FramePack module~\cite{zhang2025framecontextpackingdrift} and assigned negative positional embeddings to mark them as past context.

\myparagraph{Long-video generation: reference frame placement.} The reference identity image is assigned a positional embedding placed after the denoised video window, specifically position $40$ for a denoised window of $30$ latent frames. This forward offset makes the model ``chase'' the reference rather than copy it, which the authors report avoids copy-paste artifacts and looped body movements when the same reference is reused across sliding windows.

\myparagraph{Training and inference.} TalkVerse is trained with LoRA on the DiT attention layers, while audio projection, audio cross-attention, and FramePack-related modules are fully fine-tuned. Inference uses 50 denoising steps with a flow-matching solver and classifier-free guidance.

\subsection{Sampling scheme}
\label{sec:supmat:sampling}
Within each chunk, we adopt the predict-and-renoise scheme of DMD2~\citep{yin2024improveddistributionmatchingdistillation}. Standard flow matching integration $z_{t-\Delta t} = z_t + v_\theta(z_t, t, c) \cdot \Delta t$ assumes that intermediate states lie on the learned ODE trajectory, an assumption that breaks down in the few-step regime, where each step spans a large interval of noise levels. Given a schedule of $K = 4$ denoising steps with timesteps $1 = t_0 > t_1 > \cdots > t_K = 0$, the chunk is initialized as $z_{t_0} \sim \mathcal{N}(0, I)$, and each step $k$ proceeds in two parts. The student first predicts the velocity $v_\theta(z_{t_k}, t_k, c)$ and reconstructs an estimate of the clean latent under the linear flow matching parameterization:
\begin{equation}
\label{eq:predict}
    \hat{x}_0^{(k)} = z_{t_k} - t_k \cdot v_\theta(z_{t_k}, t_k, c).
\end{equation}
This estimate is then re-noised to the next noise level $t_{k+1}$ with fresh Gaussian noise $\epsilon \sim \mathcal{N}(0, I)$:
\begin{equation}
\label{eq:renoise}
    z_{t_{k+1}} = (1 - t_{k+1}) \cdot \hat{x}_0^{(k)} + t_{k+1} \cdot \epsilon.
\end{equation}
The final latent of the chunk is $\hat{x}_0 = \hat{x}_0^{(K-1)}$, obtained at the last denoising step. The re-noising in Eq.~\ref{eq:renoise} projects each intermediate estimate back onto the manifold of noisy latents the student was trained on, rather than following a straight-line integration of the velocity field. The same scheme is used during Stage~2 training to roll out the student without backpropagation and produce the noisy states $\{\hat{x}_{t_k}\}$ used by backward simulation (Sec.~\ref{sec:method:stage2}), keeping training and inference aligned.

\subsection{Training Details}
\label{sec:supmat:params}

We implement the chunk-wise causal attention mask (sliding window plus
initial-chunk sink, Fig.~\ref{fig:attmask}) with
FlexAttention~\cite{dong2024flexattention}, which compiles custom attention-mask patterns efficiently. This is only required at training, where attention spans the full
sequence; at inference the rolling KV cache reduces attention to the cached
chunks, so no custom masking is needed. Full optimizer settings, LoRA
configuration, and per-stage schedules \upd{of AURA 5B} are reported in
Table~\ref{tab:hparams} \upd{(AURA 1.3B: Sec.~\ref{sec:supmat:aura13b})}.

\begin{table}[h]
\centering
\small
\caption{\textbf{\upd{AURA 5B} training hyperparameters}. Stage~1 trains the student via ODE teacher forcing; Stage~2 trains the student and critic jointly via DMD with backward simulation, alternating $5$ critic updates per student update. All trainings use LoRA adapters with the optimizer settings below.}
\label{tab:hparams}
\resizebox{\textwidth}{!}{ 
\begin{tabular}{lccc}
\toprule
 & Stage~1 & Stage~2 (student) & Stage~2 (critic) \\
\midrule
Training steps              & $120$K       & $\approx 833$  & $\approx 4{,}167$ \\
Global batch size           & $256$        & $256$         & $256$         \\
Learning rate               & 1e-5         & 3.5e-7          & 3.5e-7          \\
Optimizer                   & AdamW        & AdamW         & AdamW         \\
Optimizer $(\beta_1, \beta_2)$ & (0.9, 0.999)     & (0.9, 0.999)          & (0.9, 0.999)          \\
Weight decay                & 0.01        & 0.01          & 0.01          \\
LoRA rank $r$               & $128$        & $128$         & $128$         \\
LoRA $\alpha$               & $128$         & $128$         & $128$         \\
LoRA target modules         & attention    & attention     & attention     \\
Extra unfrozen params       & timestep emb., head & timestep emb., head & --- \\
Initialization              & TalkVerse    & Stage~1 merged & TalkVerse    \\
\midrule
Stage~2 total optimizer steps        & --- & \multicolumn{2}{c}{$5$K} \\
Critic-to-student update ratio       & --- & \multicolumn{2}{c}{$5{:}1$} \\
Video clip length (latent frames)    & 30 & \multicolumn{2}{c}{30} \\
Teacher ODE steps (Stage~1 data)     & 50 & --- & --- \\
Teacher guidance scale (Stage~1 data) & 4.5 & --- & --- \\
\midrule
Hardware                    & \multicolumn{3}{c}{$64 \times$ H100} \\
Precision                   & \multicolumn{3}{c}{bf16} \\
\bottomrule
\end{tabular}}
\end{table}

\subsection{\upd{AURA 1.3B}}
\label{sec:supmat:aura13b}
\upd{AURA 1.3B follows the AURA 5B recipe (chunk-wise causal mask, mixed per-chunk timesteps, rolling KV cache, two-stage distillation) on a smaller backbone, with its own bidirectional teacher. Hyperparameters are listed in Tab.~\ref{tab:hparams13b}.

\myparagraph{Backbone and audio features.} The backbone is Wan2.1 $1.3$B text-to-video~\cite{wan2025wanopenadvancedlargescale} with its $16$-channel VAE ($4{\times}8{\times}8$ compression). Videos are generated at $16$\,Hz in one of two shapes, $768{\times}512$ or $576{\times}576$, chosen from the aspect ratio of the reference image; a training clip covers $20$ latent frames ($77$ frames, $4.8$\,s). Speech is encoded with the Mimi codec~\cite{defossez2024moshi}, keeping only its first, semantic quantization level as a continuous $512$-dimensional latent. Audio enters through cross-attention layers initialized from the text cross-attention layers.

\myparagraph{Teacher.} The bidirectional teacher is trained with flow matching on $23{,}147$ real Hallo3~\cite{cui2025hallo3highlydynamicrealistic} clips, each truncated to its first $20$ latent frames ($4.8$\,s, $31$\,h in total; $500$ clips held out for validation, none of the \textsc{Hallo3 test} clips), each with its own speech track, using LoRA adapters on the backbone and full-rank audio cross-attention. The loss is a flow-matching MSE weighted by face and body masks, and the audio, image and text conditions are each dropped with probability $0.1$ for classifier-free guidance. We use the checkpoint at $60$K steps. Unlike TalkVerse, the teacher has no motion frames (no FramePack module): it is only used for distillation, which requires single $20$-latent windows, so it never has to extend a video beyond one window.

\myparagraph{Stage~1 data.} Since the teacher is trained on real Hallo3 videos, we build its ODE trajectories on $30$K Hallo3 conditions (first frame, the clip's own speech, and its caption, or the generic prompt \emph{``a person is talking''} for $61\%$ of them), disjoint from \textsc{Hallo3 test}, with $50$ ODE steps and a guidance scale of $4.5$.

\myparagraph{Stage~1 and Stage~2.} Stage~1 regresses the teacher trajectories as for AURA 5B, with the clean timestep $t=0$ included among the per-chunk timesteps, and LoRA adapters on the attention layers and on the audio layers. Stage~2 uses DMD with backward simulation on $92$K conditions (Hallo3 first frames and prompts paired with synthetic speech from the pipeline of Sec.~\ref{sec:method:dataset}, disjoint from \textsc{Hallo3 test}); the student audio adapters are frozen, and the generator is read from an exponential moving average of its weights (decay $0.99$). We keep the EMA checkpoint at $6{,}500$ optimizer steps. At inference, AURA 1.3B uses the $4$-step schedule of AURA 5B and re-indexes the RoPE positions of the generated frames so that they never exceed the range seen during training.}

\begin{table}[h]
\centering
\small
\caption{\upd{\textbf{AURA 1.3B training hyperparameters.} Same optimizer settings as Tab.~\ref{tab:hparams} (AdamW, LoRA on attention, rank and $\alpha$ $128$, bf16); $5$ critic updates per student update in Stage~2.}}
\label{tab:hparams13b}
\resizebox{\textwidth}{!}{
\begin{tabular}{lcccc}
\toprule
 & \upd{Teacher} & \upd{Stage~1} & \upd{Stage~2 (student)} & \upd{Stage~2 (critic)} \\
\midrule
\upd{Data} & \upd{$23$K real Hallo3 clips} & \upd{$30$K teacher trajectories} & \multicolumn{2}{c}{\upd{$92$K conditions}} \\
\upd{Training steps} & \upd{$60$K} & \upd{$40$K} & \upd{$\approx 1{,}083$} & \upd{$\approx 5{,}417$} \\
\upd{Global batch size} & \upd{$256$} & \upd{$256$} & \upd{$256$} & \upd{$256$} \\
\upd{Learning rate} & \upd{1e-5} & \upd{1e-5} & \upd{2e-7} & \upd{2e-7} \\
\upd{Gradient clipping} & \upd{---} & \upd{---} & \multicolumn{2}{c}{\upd{$0.5$}} \\
\upd{Initialization} & \upd{Wan2.1 T2V $1.3$B} & \upd{Teacher} & \upd{Stage~1 merged} & \upd{Teacher} \\
\upd{Clip length (latent frames)} & \upd{$20$} & \upd{$20$} & \multicolumn{2}{c}{\upd{$20$}} \\
\upd{Hardware} & \multicolumn{4}{c}{\upd{$32 \times$ H100}} \\
\bottomrule
\end{tabular}}
\end{table}

\subsection{\upd{Per-offset metrics}}
\label{sec:supmat:perofs}
Table~\ref{tab:celebv_full} reports the full per-offset results on the \textsc{CelebV-HQ test} set.

\begin{table*}[t]
\centering
\scriptsize
\setlength{\tabcolsep}{3pt}
\renewcommand{\arraystretch}{1.05}
\caption{\upd{\textbf{Per-offset breakdown on \textsc{Hallo3 test}}, complementing the averaged version in the main paper (Tab.~\ref{tab:hallo3_main}).} FVD and JEDi are reported as triplets at temporal offsets at three different (temporal) frames: $@0 / @101 / @109$. \emph{WE}: warping error. \emph{DiT / VAE}: throughput of the diffusion transformer and VAE decoder alone, in frames per second on a single H$100$ GPU. \emph{Total} (fps): end-to-end throughput, including model offload and overhead. \textbf{Bold}: best; \underline{underlined}: second best (applied per offset within each triplet). \upd{AURA 1.3B outputs $480$p at $16$\,Hz (like Wan-S2V and Live Avatar), the other AURA and TalkVerse rows $25$\,Hz.}
}
\label{tab:hallo3_full}
\resizebox{\textwidth}{!}{%
\begin{tabular}{l c c c c c c c c c c c c}
\toprule
Method & Params & Res & Steps & DiT / VAE (fps)$\uparrow$ & Total (fps)$\uparrow$ & FID$\downarrow$ & FVD @$(0 / 101 / 109$)$\downarrow$ & JEDi @($0 / 101 / 109$)$\downarrow$ & CSIM$\uparrow$ & Sync-C$\uparrow$ & Sync-D$\downarrow$ & WE$\downarrow$ \\
\midrule
Hallo3        & 14.5B & 480p & 50 & $0.24 / 3.2$    & $0.22$  & \upd{\textbf{42.8}}    & \upd{185.4 / 184.7 / 180.1}            & \upd{\underline{0.25} / \textbf{0.15} / \textbf{0.14}}    & 0.74          & \underline{5.06} & 9.9          & 0.0036 \\
Wan-S2V       & 14B   & 720p & 40 & $0.08 / 6.7$    & $0.08$  & \upd{57.7}             & \upd{186.0 / 203.1 / 200.2}                                    & \upd{\textbf{0.22} / 0.28 / 0.27} & 0.71    & \upd{\textbf{5.42}}   & \textbf{9.0} & \underline{0.0032} \\
TalkVerse     & 5B    & 720p & 50 & $0.69 / 7.4$    & $0.58$  & \upd{58.5}             & \upd{222.6 / 212.1 / 197.5}                                    & \upd{0.40 / 0.43 / 0.37}                               & 0.74          & 3.82            & 10.5         & \textbf{0.0019} \\
Live Avatar   & 14B   & 480p & 4  & $8.4 / 4.3$     & $1.9$   & \upd{\underline{52.0}} & \upd{\underline{179.8} / \underline{183.8} / \underline{179.7}}      & \upd{0.30 / 0.51 / 0.52}                               & \upd{0.77} & 4.90         & \underline{9.3} & 0.0039 \\
\midrule
\textbf{AURA \upd{5B} (ours)}    & \textbf{5B} & \textbf{720p} & \textbf{4} & \upd{$\underline{19.7} / \underline{22.7}$} & \upd{$\underline{9.7}$} & \upd{56.9} & \upd{198.1 / 188.2 / 200.2} & \upd{0.51 / \underline{0.22} / \underline{0.21}} & \upd{\underline{0.79}} & \upd{4.30} & \upd{9.87} & \upd{\underline{0.0032}} \\
\upd{\textbf{AURA 1.3B (ours)}} & \upd{\textbf{1.3B}} & \upd{\textbf{480p}} & \upd{\textbf{4}} & \upd{$\mathbf{28.8 / 96.9}$} & \upd{$\mathbf{20.7}$} & \upd{56.8} & \upd{\textbf{172.3} / \textbf{179.9} / \textbf{179.2}} & \upd{0.45 / 0.35 / 0.41} & \upd{\textbf{0.84}} & \upd{4.07} & \upd{9.93} & \upd{0.0036} \\
\bottomrule
\end{tabular}

}
\end{table*}

\begin{table*}[t]
\centering
\footnotesize
\setlength{\tabcolsep}{3pt}
\renewcommand{\arraystretch}{1.1}
\caption{\textbf{Per-offset breakdown on \textsc{CelebV-HQ test}}, complementing the averaged version in the main paper. FVD and JEDi are reported as triplets at temporal offsets $@0/@101/@109$. \emph{WE}: warping error. \emph{Sync-C/D}: Sync-C ($\uparrow$) / Sync-D ($\downarrow$). \emph{fps}: end-to-end throughput on a single H100 GPU for a $10$\,s clip; per-component DiT and VAE values appear in the main paper. \textbf{Bold}: best; \underline{underlined}: second best (per offset within each triplet).}
\resizebox{\textwidth}{!}{%
\begin{tabular}{l c c c c c c c c c c}
\toprule
Method & Params & Res & Steps & fps$\uparrow$ & FID$\downarrow$ & FVD ($@0/@101/@109$)$\downarrow$ & JEDi ($@0/@101/@109$)$\downarrow$ & CSIM$\uparrow$ & Sync-C/D & WE$\downarrow$ \\
\midrule
Hallo3      & $14.5$B & $480$p & $50$ & $0.22$  & \upd{\textbf{39.1}}    & \upd{\underline{192.2}/\textbf{208.4}/\textbf{206.9}} & \upd{\textbf{0.26}/\textbf{0.25}/\textbf{0.22}} & \upd{$0.71$} & \upd{\underline{6.23}}/$9.4$        & \textbf{0.0038} \\
Wan-S2V     & $14$B   & $720$p & $40$ & $0.08$ & \upd{\underline{50.4}}           & \upd{$294.3$/$288.7$/$285.9$}                         & \upd{$0.66$/$0.52$/$0.55$}            & $0.62$           & \upd{\textbf{6.24}}/\textbf{8.9} & $0.0061$ \\
TalkVerse   & $5$B    & $720$p & $50$ & $0.58$  & \upd{$53.2$}           & \upd{$233.2$/$268.9$/$265.5$}                         & \upd{\underline{0.33}/$0.66$/$0.75$}            & $0.59$           & \upd{$4.78$}/$10.5$              & $0.0059$ \\
Live Avatar & $14$B   & $480$p & $4$  & $1.90$  & \upd{$52.4$} & \upd{$237.2$/$281.4$/$283.9$}                         & \upd{$0.48$/$1.26$/$1.29$}                      & $0.67$           & $5.90$/\underline{9.2}     & $0.0057$ \\
\midrule
\textbf{AURA \upd{5B} (ours)}      & $\mathbf{5}$\textbf{B} & $\mathbf{720}$\textbf{p} & $\mathbf{4}$ & \upd{$\underline{9.7}$}  & \upd{$54.0$} & \upd{\textbf{182.0}/\underline{219.4}/\underline{223.0}} & \upd{\underline{0.33}/\underline{0.27}/\underline{0.36}} & \upd{\underline{0.72}} & \upd{$4.94$/$10.0$} & \upd{\underline{0.0043}} \\
\upd{\textbf{AURA 1.3B (ours)}} & \upd{$\mathbf{1.3}$\textbf{B}} & \upd{$\mathbf{480}$\textbf{p}} & \upd{$\mathbf{4}$} & \upd{$\mathbf{20.7}$} & \upd{$51.5$} & \upd{$243.1$/$251.0$/$252.2$} & \upd{$0.57$/$0.53$/$0.62$} & \upd{\textbf{0.77}} & \upd{$4.76$/$10.1$} & \upd{$0.0046$} \\
\bottomrule
\end{tabular}
}
\label{tab:celebv_full}
\end{table*}

\subsection{Bubble plot for FVD and JEDi}

\begin{figure*}[t]
    \centering
    \includegraphics[width=\textwidth]{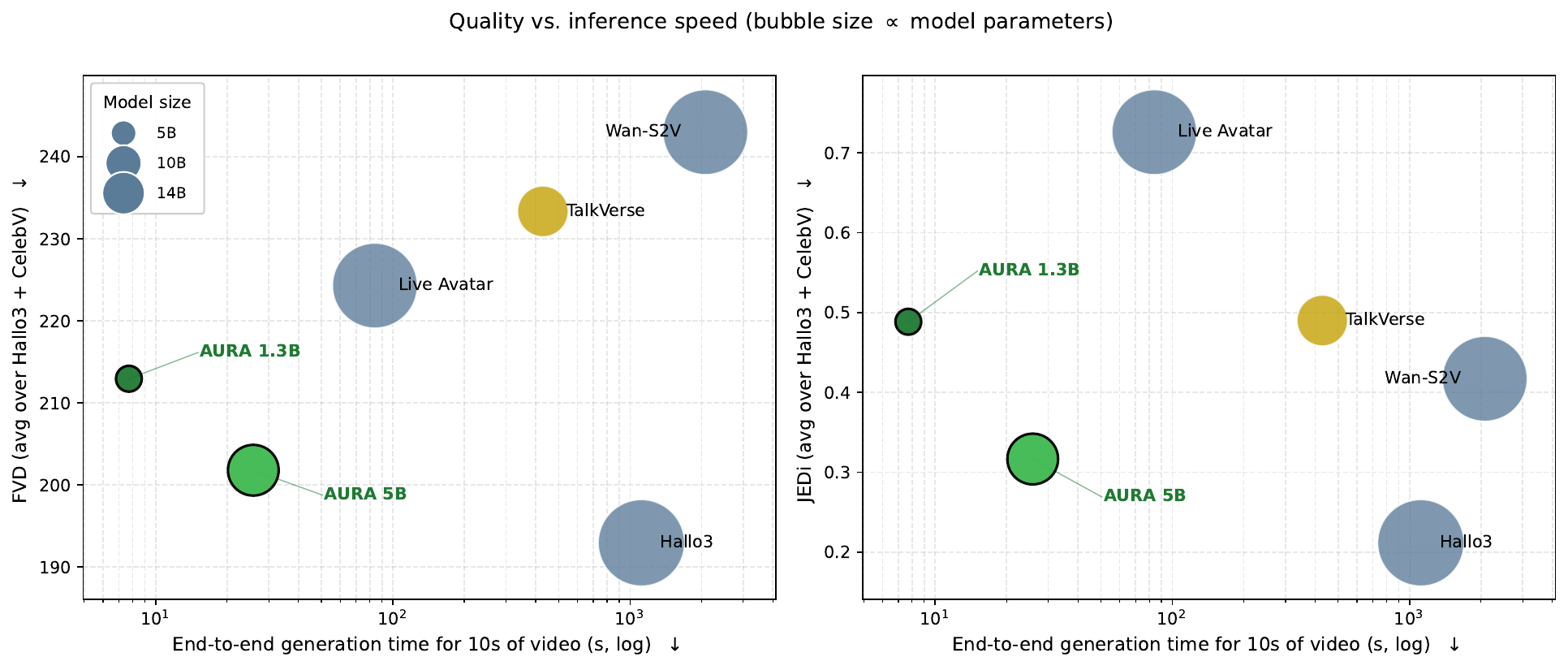}
    \caption{\textbf{Quality--speed trade-off across audio-driven avatar generation models}. The horizontal axis reports end-to-end generation time for $10$\,s of video on a single H$100$ GPU (log scale, lower is better). The vertical axis reports FVD (\textit{left}) and JEDi (\textit{right}), averaged over the Hallo3 and CelebV-HQ test sets (lower is better). Bubble area is proportional to the number of model parameters. AURA (\textit{green}) achieves quality comparable to multi-step baselines at a fraction of their generation time.}
    \label{fig:bubble_plot}
\end{figure*}

Figure~\ref{fig:bubble_plot} shows the bubble plot for both FVD and JEDi metrics.

\subsection{Full Results for long evaluations}
\begin{figure*}[t]
  \centering
  \includegraphics[width=\textwidth]{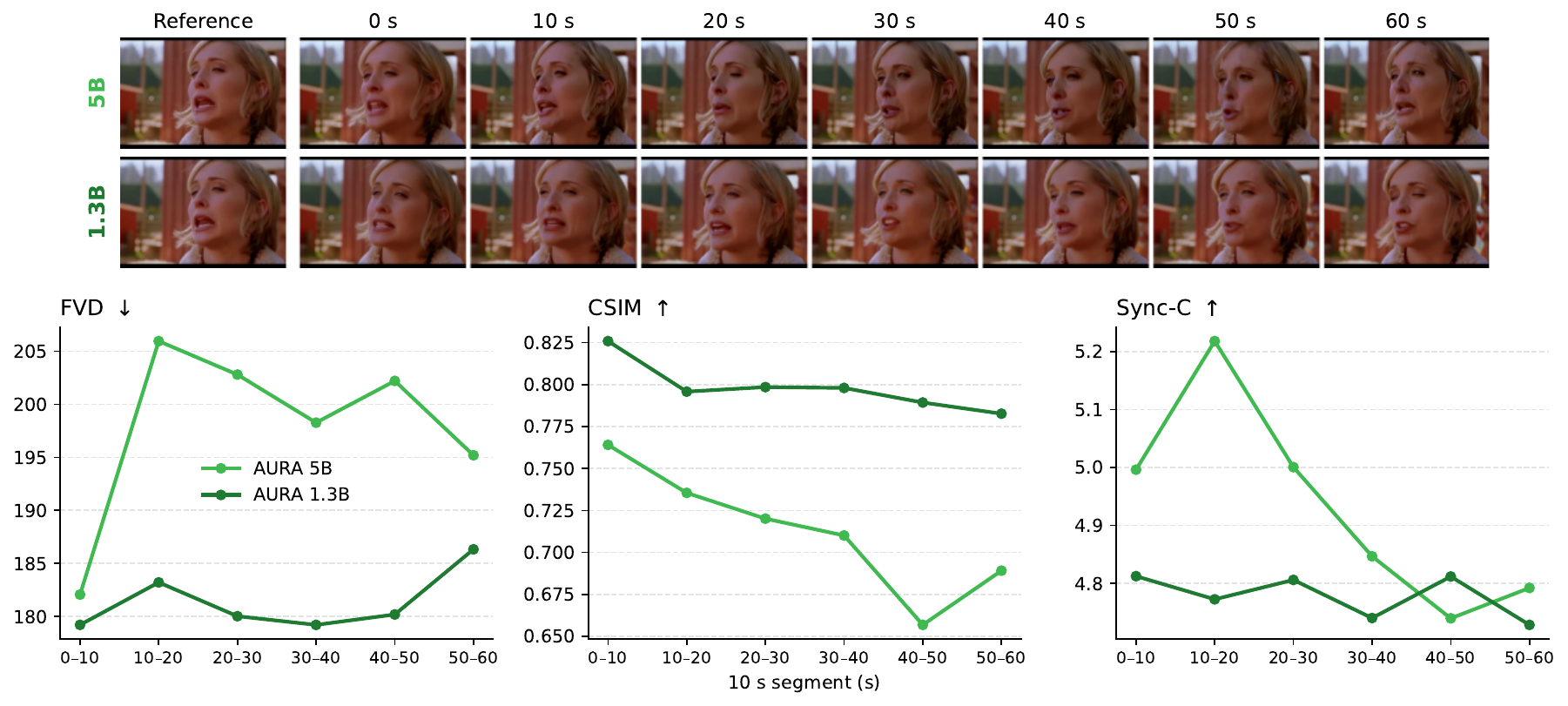}
  \caption{\upd{\textbf{One minute of streaming.} \textit{(Top)} Frames of one \textsc{Hallo3 test} clip at $t = 0, 10, \dots, 60$\,s for AURA 5B and AURA 1.3B, next to the reference image. \textit{(Bottom)} FVD, CSIM and Sync-C per $10$\,s segment over the $100$ references (Tab.~\ref{tab:long_eval_supp}).}
}
  \label{fig:long}
\end{figure*}

Table \ref{tab:long_eval_supp} provides results for long-form evaluations\upd{, and Fig.~\ref{fig:long} shows them with frames of one clip over the minute}.

\begin{table}[t]
\centering
\setlength{\tabcolsep}{4pt}
\caption{\textbf{Long-form evaluation.} \upd{AURA 5B and AURA 1.3B ($4$ steps, fixed KV cache) on the $100$
\textsc{Hallo3 test} references driven by $60$\,s of speech; metrics per $10$\,s
segment (FVD: mean of three windows per segment). Warping error in $10^{-3}$.}
}
\label{tab:long_eval_supp}
\resizebox{0.8\columnwidth}{!}{
\begin{tabular}{lccccccc}
\toprule
Metric & 0--10s & 10--20s & 20--30s & 30--40s & 40--50s & 50--60s & Mean \\
\midrule
\multicolumn{8}{l}{\textit{\upd{AURA 5B}}} \\
FVD\,$\downarrow$ & \upd{182.1} & \upd{206.0} & \upd{202.8} & \upd{198.3} & \upd{202.2} & \upd{195.2} & \upd{\textbf{197.8}} \\
CSIM\,$\uparrow$ & \upd{0.764} & \upd{0.735} & \upd{0.720} & \upd{0.710} & \upd{0.657} & \upd{0.689} & \upd{\textbf{0.713}} \\
Sync-C\,$\uparrow$ & \upd{5.00} & \upd{5.22} & \upd{5.00} & \upd{4.85} & \upd{4.74} & \upd{4.79} & \upd{\textbf{4.93}} \\
Warp\,$\downarrow$ & \upd{3.50} & \upd{3.44} & \upd{3.45} & \upd{3.43} & \upd{3.99} & \upd{3.53} & \upd{\textbf{3.56}} \\
\midrule
\multicolumn{8}{l}{\textit{\upd{AURA 1.3B}}} \\
FVD\,$\downarrow$ & \upd{179.2} & \upd{183.2} & \upd{180.0} & \upd{179.2} & \upd{180.2} & \upd{186.3} & \upd{\textbf{181.4}} \\
CSIM\,$\uparrow$ & \upd{0.826} & \upd{0.796} & \upd{0.798} & \upd{0.798} & \upd{0.789} & \upd{0.783} & \upd{\textbf{0.798}} \\
Sync-C\,$\uparrow$ & \upd{4.81} & \upd{4.77} & \upd{4.81} & \upd{4.74} & \upd{4.81} & \upd{4.73} & \upd{\textbf{4.78}} \\
Warp\,$\downarrow$ & \upd{3.70} & \upd{4.06} & \upd{4.16} & \upd{4.25} & \upd{4.18} & \upd{4.22} & \upd{\textbf{4.09}} \\
\bottomrule
\end{tabular}}
\end{table}

\subsection{RoPE re-indexing scheme \upd{(AURA 5B)}}

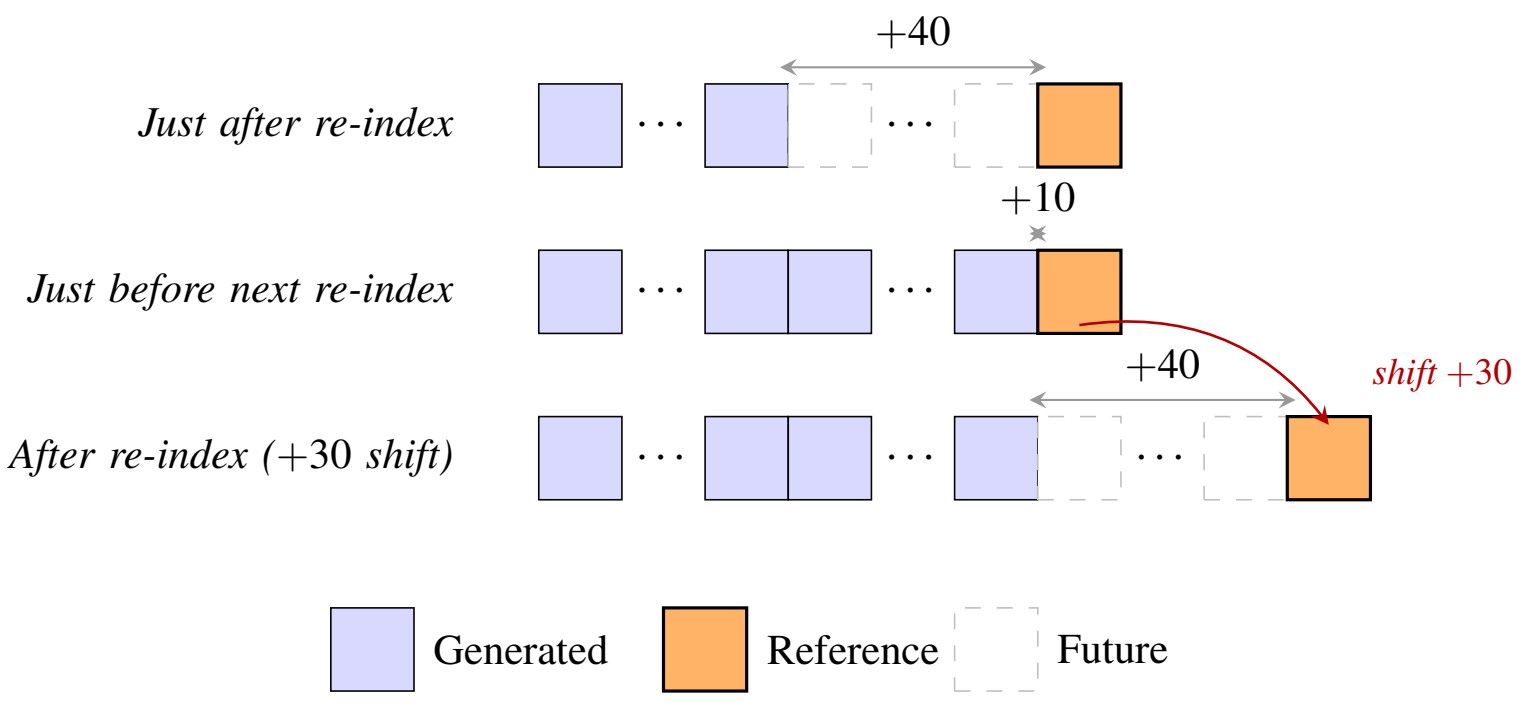
\begin{figure}[t]
    \centering
    \begin{tikzpicture}[
        x=0.55cm, y=0.55cm,
        frame/.style={draw, rectangle, minimum width=0.55cm, minimum height=0.55cm, inner sep=0pt, line width=0.3pt},
        gen/.style={frame, fill=blue!15},
        ref/.style={frame, fill=orange!60, line width=0.6pt},
        empty/.style={frame, fill=white, dashed, draw=gray!50, line width=0.3pt},
        rowlabel/.style={anchor=east, font=\footnotesize\itshape, text width=3.2cm, align=right},
        offsetlabel/.style={font=\footnotesize\itshape, anchor=south},
        dots/.style={font=\small, anchor=center},
    ]
    
    \node[rowlabel] at (-0.3, 4) {Just after re-index};
    \node[gen] at (1, 4) {};
    \node[dots] at (2, 4) {$\cdots$};
    \node[gen] at (3, 4) {};
    \node[empty] at (4, 4) {};
    \node[dots] at (5, 4) {$\cdots$};
    \node[empty] at (6, 4) {};
    \node[ref] at (7, 4) {};
    \draw[<->, >=stealth, line width=0.4pt, gray!80] (3.4, 4.7) -- (6.6, 4.7);
    \node[offsetlabel] at (5, 4.7) {$+40$};
    
    \node[rowlabel] at (-0.3, 2) {Just before next re-index};
    \node[gen] at (1, 2) {};
    \node[dots] at (2, 2) {$\cdots$};
    \node[gen] at (3, 2) {};
    \node[gen] at (4, 2) {};
    \node[dots] at (5, 2) {$\cdots$};
    \node[gen] at (6, 2) {};
    \node[ref] at (7, 2) {};
    \draw[<->, >=stealth, line width=0.4pt, gray!80] (6.4, 2.7) -- (6.6, 2.7);
    \node[offsetlabel] at (6.5, 2.7) {$+10$};
    
    \node[rowlabel] at (-0.3, 0) {After re-index ($+30$ shift)};
    \node[gen] at (1, 0) {};
    \node[dots] at (2, 0) {$\cdots$};
    \node[gen] at (3, 0) {};
    \node[gen] at (4, 0) {};
    \node[dots] at (5, 0) {$\cdots$};
    \node[gen] at (6, 0) {};
    \node[empty] at (7, 0) {};
    \node[dots] at (8, 0) {$\cdots$};
    \node[empty] at (9, 0) {};
    \node[ref] at (10, 0) {};
    \draw[<->, >=stealth, line width=0.4pt, gray!80] (6.4, 0.7) -- (9.6, 0.7);
    \node[offsetlabel] at (8, 0.7) {$+40$};
    
    \draw[->, >=stealth, line width=0.5pt, red!70!black] (7, 1.6) to[bend left=30] (10, 0.4);
    \node[font=\scriptsize\itshape, text=red!70!black, anchor=west] at (10.3, 1) {shift $+30$};
    
    \node[gen, label=right:{\footnotesize Generated}] at (-1.5, -2.3) {};
    \node[ref, label=right:{\footnotesize Reference}] at (2.5, -2.3) {};
    \node[empty, label=right:{\footnotesize Future}] at (6, -2.3) {};
    
    \end{tikzpicture}
    \caption{\textbf{RoPE re-indexing of the reference frame.} \textit{(Top)} just after a re-indexation, the reference (orange) sits $40$ latent positions ahead of the generation front. \textit{(Middle)} after $30$ generated frames, the reference is $10$ positions ahead --- the minimum offset reached during training. \textit{(Bottom)} the reference is shifted by $+30$, restoring the maximum $+40$ offset. The offset thus cycles in $[+10, +40]$, staying within the training-time regime.}
    \label{fig:rope}
\end{figure}

Figure~\ref{fig:rope} illustrates the RoPE re-indexing scheme\upd{ of the reference frame (Sec.~\ref{sec:method:rope}). It is specific to AURA 5B, whose TalkVerse teacher places the reference ahead of the generated window; AURA 1.3B instead caps the RoPE positions of the generated frames to the range seen in training (Sec.~\ref{sec:supmat:aura13b})}.

\subsection{\upd{Diagnostic metrics for the ablations}}
\label{sec:supmat:diag}
\upd{The ablations (Sec.~\ref{sec:exp:ablations}) use two diagnostics in addition to the metrics of Sec.~\ref{sec:exp:setup}, computed on the $100$ \textsc{Hallo3 test} clips ($10$\,s) and averaged over videos.}

\myparagraph{\upd{Face sharpness (Sharp.)}} \upd{Variance of the $3{\times}3$ Laplacian, a standard focus measure~\cite{pechpacheco2000diatom}, on the grey-level inner face (brows to chin, cheek to cheek). At $5$ frames per second, the largest face (RetinaFace~\cite{deng2020retinaface}, ResNet-50) is aligned on the ArcFace~\cite{Deng_2022} template at $256{\times}256$, so that its size does not depend on the output resolution; frames without a face are skipped. Higher is sharper. We do not use the whole frame: frame-level proxies (variance of the Laplacian, high-frequency FFT energy) are dominated by background texture and noise, and rank the $1$-step and Stage-1-only models as the sharpest.}

\myparagraph{\upd{Colour drift ($\Delta$Col.)}} \upd{How much further the face colour moves away from the reference over a clip. At $5$ frames per second, we compute $\Delta E$, the CIELAB distance between the mean colour of the inner face (detected and aligned as above, at $112{\times}112$) and that of the reference face. $\Delta$Col.\ is the mean $\Delta E$ over the last two seconds ($7.5$--$9.5$\,s) minus the mean over the first two ($0.5$--$2.5$\,s); lower is better. This follows the start-versus-end drift measure of FramePack~\cite{zhang2025framecontextpackingdrift} and Rolling Forcing~\cite{liu2025rollingforcingautoregressivelong}, restricted to the face and taken against the reference, hence signed. For scale, a $\Delta E$ of about $2.3$ is commonly taken as a just-noticeable difference~\cite{mahy1994evaluation}: the default model stays at $\Delta E \approx 2$ from the reference ($\Delta$Col.\ $0.4$), whereas the $t{=}50$ refresh variants drift by $3.1$, to $5.7$--$5.8$.}

\subsection{Rolling cache window size}
\label{sec:supmat:window}
We ablate the number of past chunks $n$ retained in the rolling KV cache
(Sec.~\ref{sec:method:kvcache}). Our default keeps a single past chunk ($n=1$)
alongside the first-chunk anchor; here we compare it against a larger window that
also retains the second-to-last chunk ($n=2$). \upd{The $n=2$ variant is a dedicated $4$-step student, whose Stage~2 is trained with the $n=2$ attention mask; both are evaluated on \textsc{Hallo3 test} ($10$\,s clips).}
 As reported in \hyperref[tab:abl_window]{Table~\ref{tab:abl_window}},
\upd{enlarging the window to $n=2$ slightly improves FVD ($193.2$ vs.\ $195.5$) at an equal FID, but degrades JEDi, identity, lip-sync and warping error}, while increasing cache size and lowering throughput. This supports our minimal single-chunk cache as the best quality--efficiency trade-off.

\begin{table}[ht]
\centering
\small
\setlength{\tabcolsep}{4pt}
\begin{tabular}{lccccccc}
\toprule
Variant & FID $\downarrow$ & FVD $\downarrow$ & JEDi $\downarrow$ & CSIM $\uparrow$ & Sync-C $\uparrow$ & Warp $\downarrow$ & DiT (fps) $\uparrow$ \\
\midrule
$n=1$ (default) & \upd{56.9} & \upd{195.5} & \upd{\textbf{0.31}} & \upd{\textbf{0.79}} & \upd{\textbf{4.30}} & \upd{\textbf{0.0032}} & \textbf{20} \\
$n=2$           & \upd{\textbf{56.8}} & \upd{\textbf{193.2}} & \upd{0.37} & \upd{0.78} & \upd{4.20} & \upd{0.0035} & 18 \\
\bottomrule
\end{tabular}
\caption{\textbf{Rolling cache window size.} Effect of the number $n$ of retained past
chunks on quality and speed, on \upd{\textsc{Hallo3 test} ($10$\,s clips). Enlarging the window from $n=1$ to $n=2$ slightly improves FVD but degrades JEDi, CSIM, Sync-C and WE,} while
increasing cache size and reducing throughput. FVD and JEDi are averaged over the
three temporal offsets ($@0$, $@101$, $@109$). Best per column in \textbf{bold}.}
\label{tab:abl_window}
\end{table}
\subsection{Hallo3 failure case}

\begin{figure}[t]
    \centering
    \begin{minipage}[b]{0.36\linewidth}
        \centering
        \includegraphics[width=\linewidth]{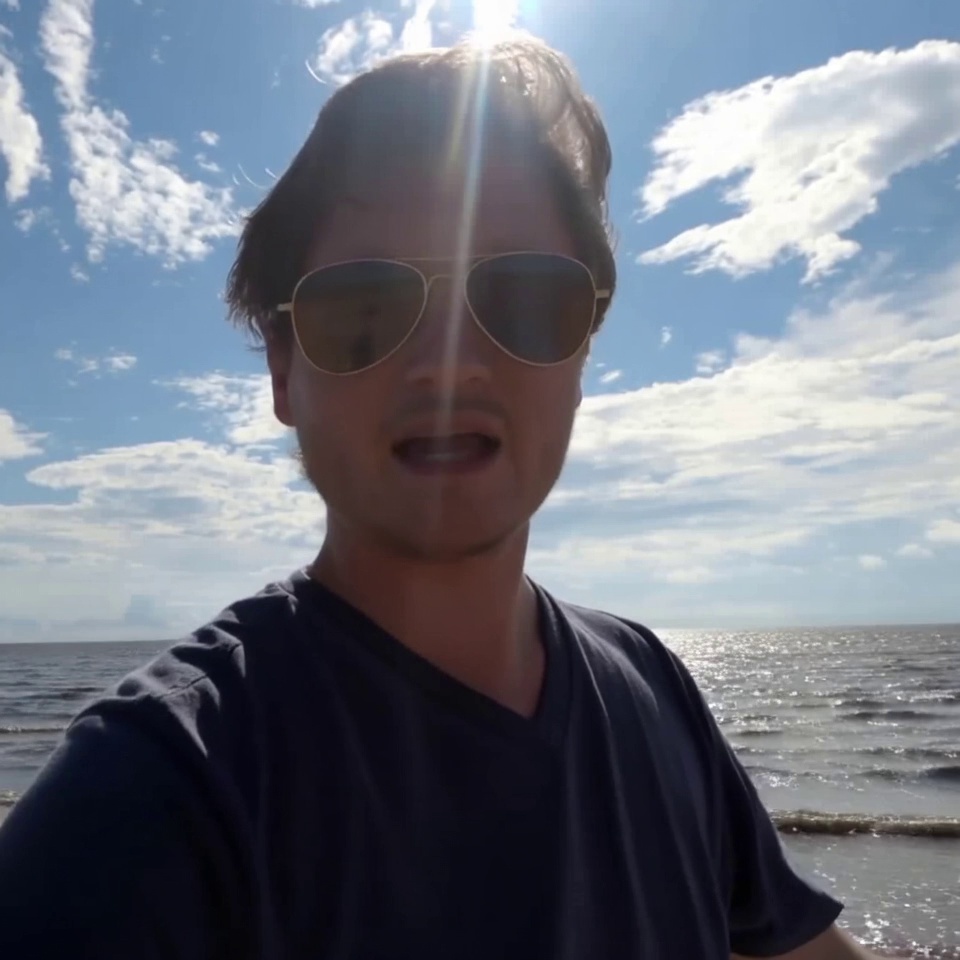}
    \end{minipage}
    \hspace{4pt}
    \begin{minipage}[b]{0.36\linewidth}
        \centering
        \includegraphics[width=\linewidth]{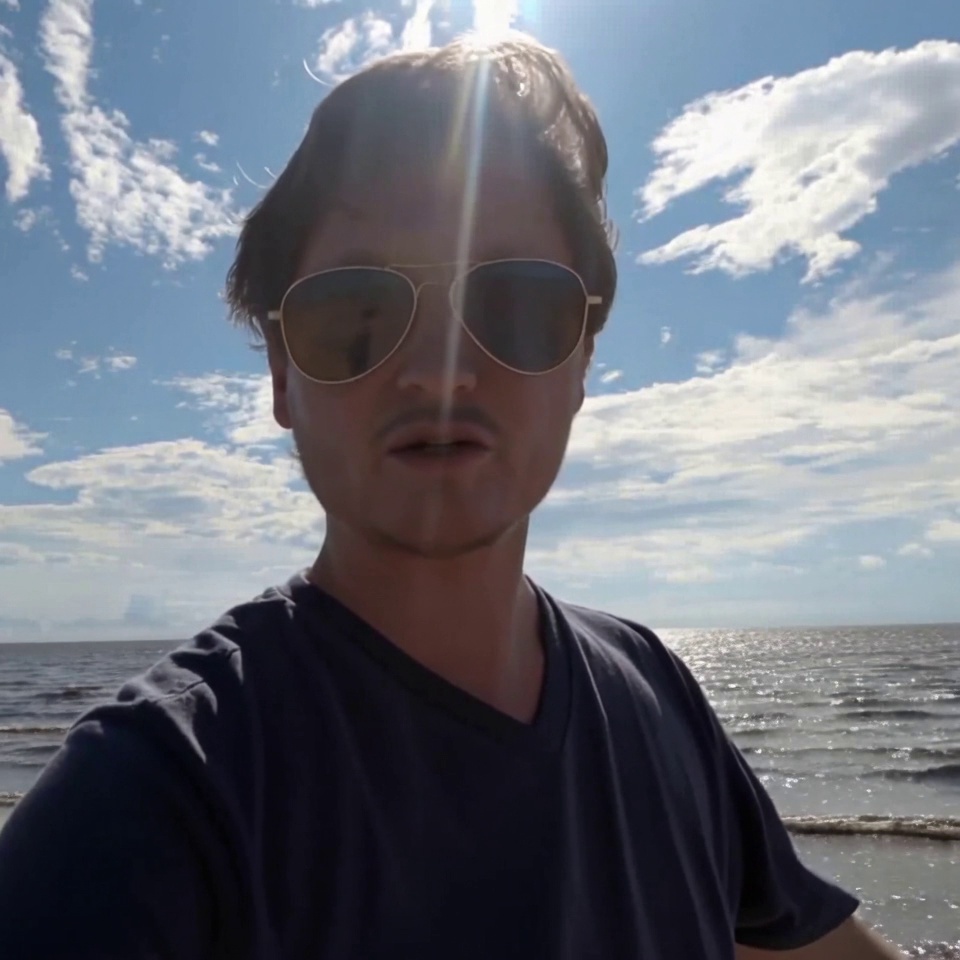}
    \end{minipage}
    \\[2pt]
    {\small AURA (ours)}
    \\[6pt]
    \begin{minipage}[b]{0.36\linewidth}
        \centering
        \includegraphics[width=\linewidth]{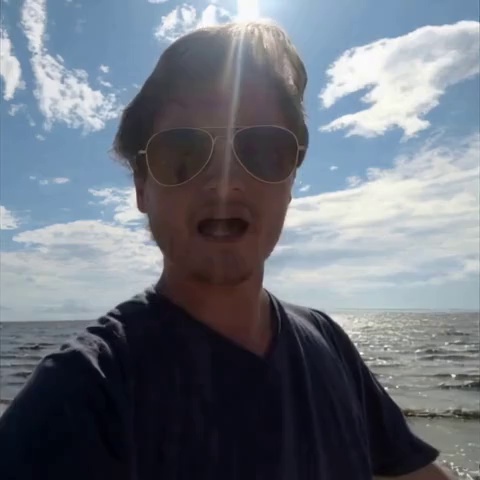}
    \end{minipage}
    \hspace{4pt}
    \begin{minipage}[b]{0.36\linewidth}
        \centering
        \includegraphics[width=\linewidth]{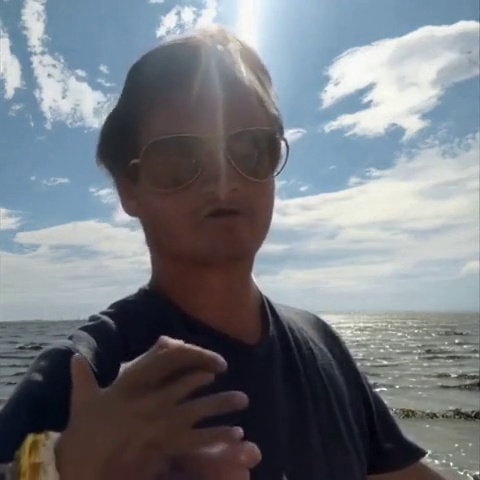}
    \end{minipage}
    \\[2pt]
    {\small Hallo3}
    \vspace{6pt}
    \caption{\textbf{Qualitative comparison on the Hallo3 test set}. Despite stronger FID/FVD scores (Table~\ref{tab:hallo3_main}), Hallo3 generations exhibit recurrent hand artifacts (\textit{bottom}) that distribution-based metrics fail to penalize. AURA (\textit{top}) produces cleaner outputs at a fraction of the generation time.}
    \label{fig:qual_hallo3}
\end{figure}

Figure~\ref{fig:qual_hallo3} shows artifacts observed in some Hallo3 outputs.

\subsection{Comparison with teacher \upd{(AURA 5B)}}

\begin{figure*}[t]
    \centering
    \makebox[0.18\linewidth][c]{}\hfill
    \makebox[0.78\linewidth][s]{%
        \makebox[0.255\linewidth][c]{\small $t \approx 1$\,s}\hfill
        \makebox[0.255\linewidth][c]{\small $t \approx 4$\,s}\hfill
        \makebox[0.255\linewidth][c]{\small $t \approx 5$\,s}%
    }
    \\[2pt]
    \begin{minipage}[c]{0.18\linewidth}
        \centering
        \includegraphics[width=\linewidth]{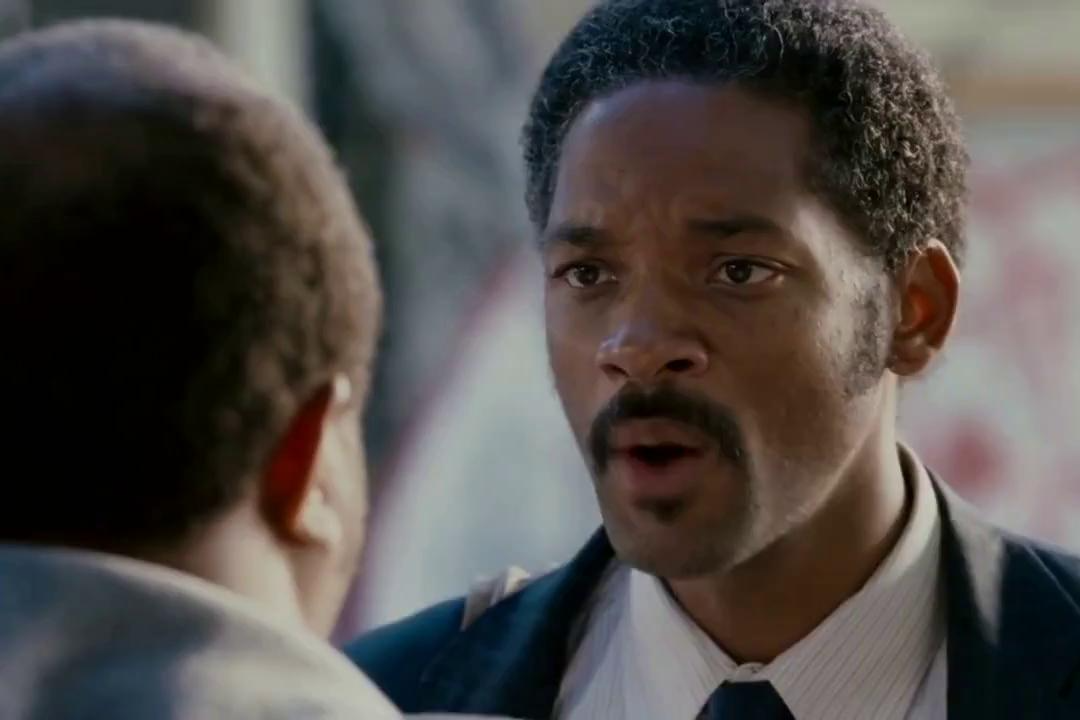}
        \\[2pt]
        {\small \emph{Reference}}
    \end{minipage}%
    \hfill
    \begin{minipage}[c]{0.78\linewidth}
        \centering
        \includegraphics[width=0.32\linewidth]{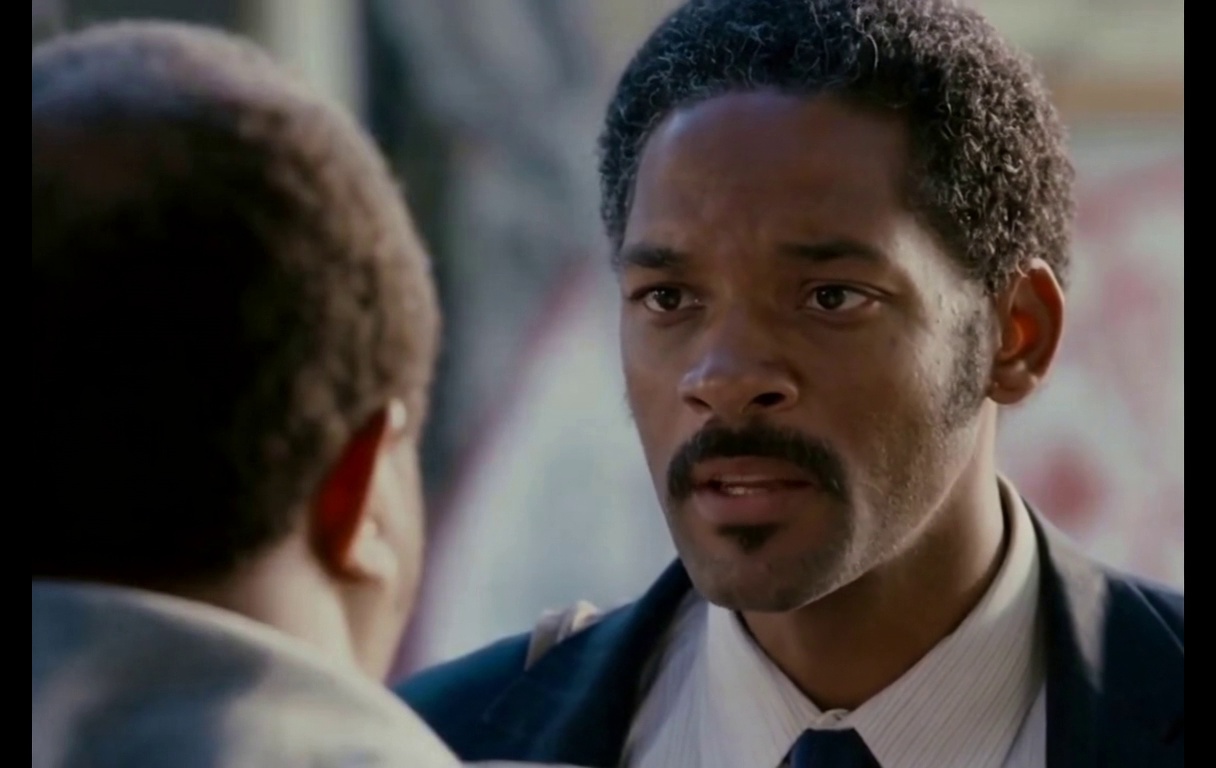}\hfill
        \includegraphics[width=0.32\linewidth]{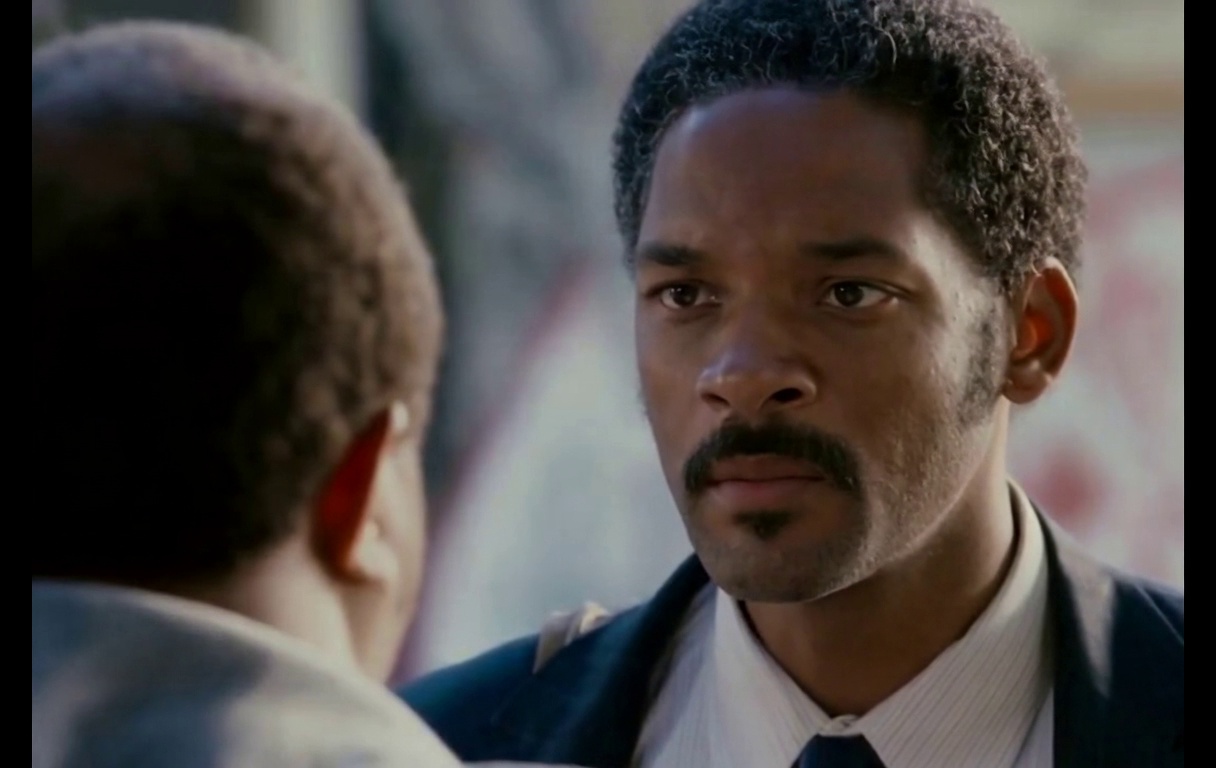}\hfill
        \includegraphics[width=0.32\linewidth]{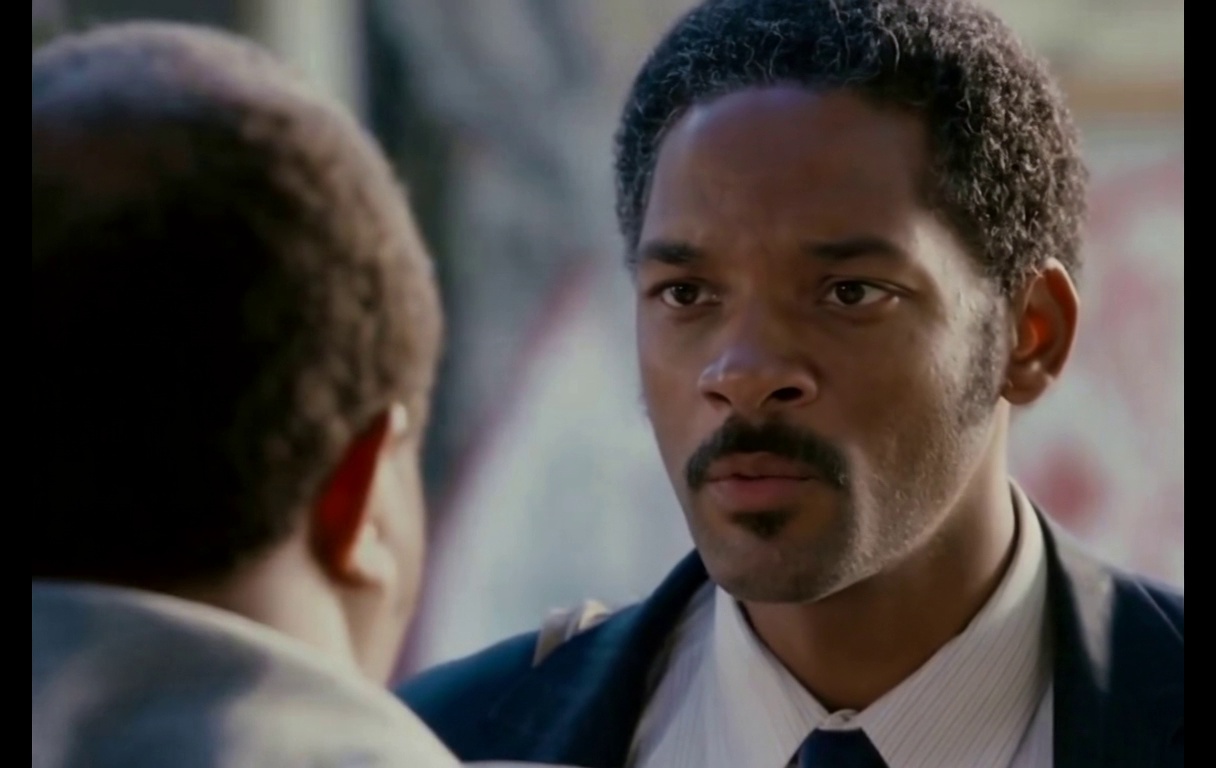}
        \\[1pt]
        {\small TalkVerse (teacher) -- $50$ steps, $0.58$\,fps}
        \\[6pt]
        \includegraphics[width=0.32\linewidth]{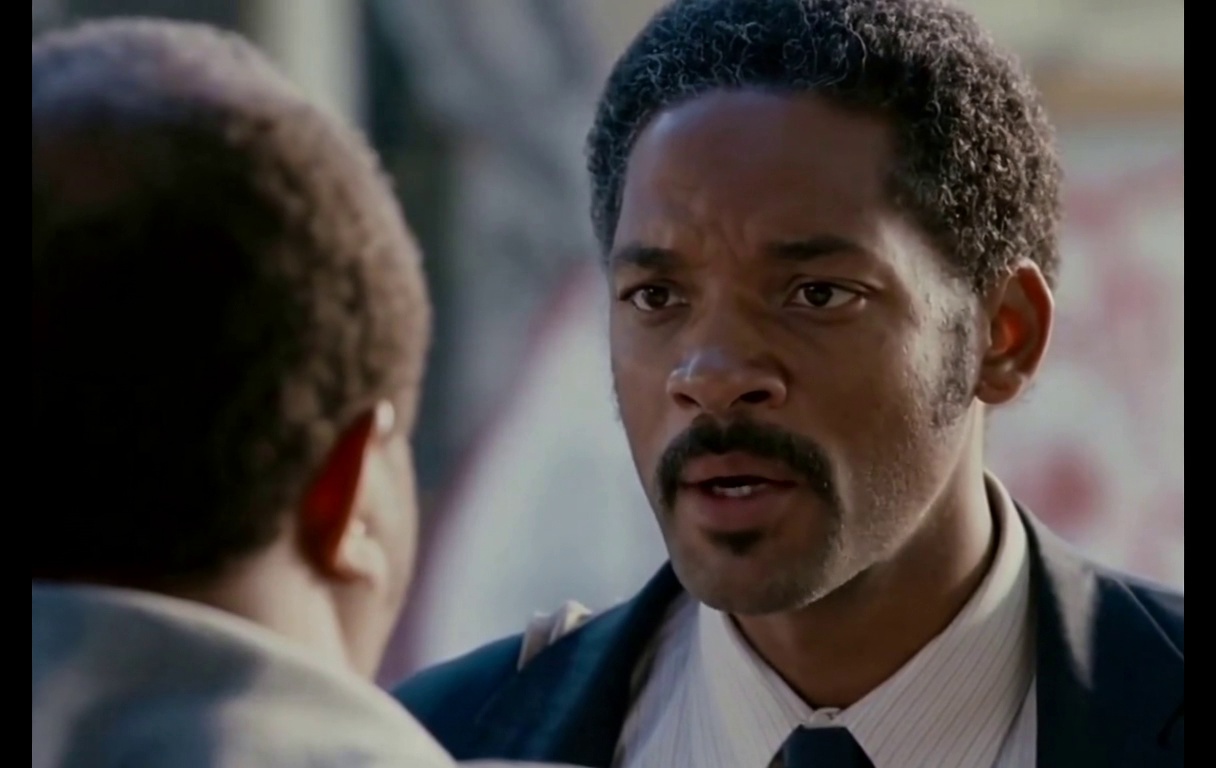}\hfill
        \includegraphics[width=0.32\linewidth]{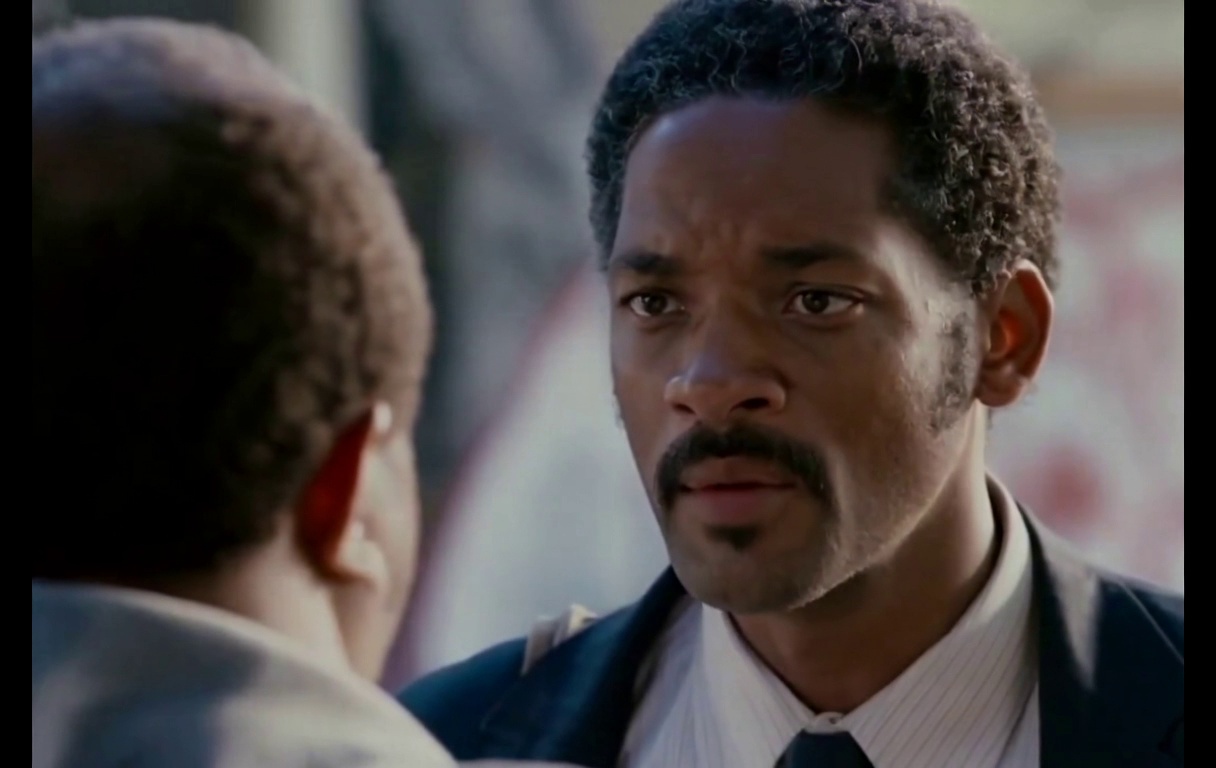}\hfill
        \includegraphics[width=0.32\linewidth]{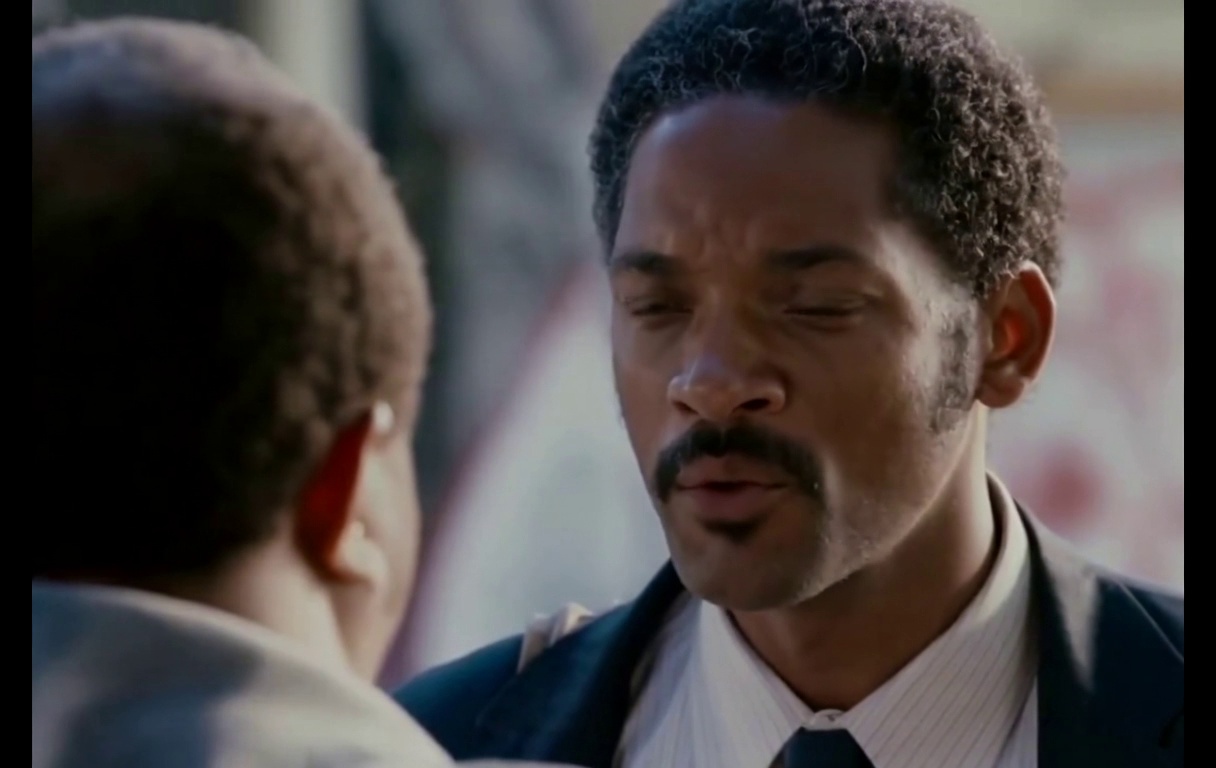}
        \\[1pt]
        {\small AURA\upd{ 5B} (ours) -- $4$ steps, $9.7$\,fps}
    \end{minipage}
    \vspace{4pt}
    \caption{\textbf{Qualitative comparison between AURA\upd{ 5B} and its TalkVerse teacher}. Given the same reference image (\textit{left}) and a $10$\,s audio input, AURA\upd{ 5B} (\textit{bottom}) generates video of comparable visual quality to the teacher (\textit{top}) while running approximately $17\times$ faster end-to-end on a single H100 GPU. \upd{Full videos are on the \href{https://kyutai.org/aura/}{project page}.}}
    \label{fig:qual_aura_vs_tv}
\end{figure*}

Figure~\ref{fig:qual_aura_vs_tv} compares AURA\upd{ 5B} and \upd{its teacher} TalkVerse \upd{on} samples generated from the same reference image and audio input.

%% file: main.bbl
\begin{thebibliography}{64}
\providecommand{\natexlab}[1]{#1}
\providecommand{\url}[1]{\texttt{#1}}
\expandafter\ifx\csname urlstyle\endcsname\relax
  \providecommand{\doi}[1]{doi: #1}\else
  \providecommand{\doi}{doi: \begingroup \urlstyle{rm}\Url}\fi

\bibitem[Baevski et~al.(2020)Baevski, Zhou, Mohamed, and
  Auli]{baevski2020wav2vec20frameworkselfsupervised}
Alexei Baevski, Henry Zhou, Abdelrahman Mohamed, and Michael Auli.
\newblock wav2vec 2.0: A framework for self-supervised learning of speech
  representations.
\newblock \emph{Advances in neural information processing systems},
  33:\penalty0 12449--12460, 2020.

\bibitem[Blattmann et~al.(2023{\natexlab{a}})Blattmann, Dockhorn, Kulal,
  Mendelevitch, Kilian, Lorenz, Levi, English, Voleti, Letts, Jampani, and
  Rombach]{blattmann2023stablevideodiffusionscaling}
Andreas Blattmann, Tim Dockhorn, Sumith Kulal, Daniel Mendelevitch, Maciej
  Kilian, Dominik Lorenz, Yam Levi, Zion English, Vikram Voleti, Adam Letts,
  Varun Jampani, and Robin Rombach.
\newblock Stable video diffusion: Scaling latent video diffusion models to
  large datasets.
\newblock \emph{Computing Research Repository (CoRR)}, abs/2311.15127,
  2023{\natexlab{a}}.
\newblock URL \url{https://doi.org/10.48550/arXiv.2311.15127}.

\bibitem[Blattmann et~al.(2023{\natexlab{b}})Blattmann, Rombach, Ling,
  Dockhorn, Kim, Fidler, and
  Kreis]{blattmann2023alignlatentshighresolutionvideo}
Andreas Blattmann, Robin Rombach, Huan Ling, Tim Dockhorn, Seung~Wook Kim,
  Sanja Fidler, and Karsten Kreis.
\newblock Align your latents: High-resolution video synthesis with latent
  diffusion models.
\newblock In \emph{Proceedings of the IEEE/CVF conference on computer vision
  and pattern recognition, CVPR'23}, pages 22563--22575, 2023{\natexlab{b}}.

\bibitem[Brock et~al.(2019)Brock, Donahue, and
  Simonyan]{brock2019largescalegantraining}
Andrew Brock, Jeff Donahue, and Karen Simonyan.
\newblock Large scale {GAN} training for high fidelity natural image synthesis.
\newblock In \emph{International Conference on Learning Representations,
  ICLR'19}, 2019.

\bibitem[Brooks et~al.(2024)Brooks, Peebles, Holmes, DePue, Guo, Jing, Schnurr,
  Taylor, Luhman, Luhman, Ng, Wang, and Ramesh]{videoworldsimulators2024}
Tim Brooks, Bill Peebles, Connor Holmes, Will DePue, Yufei Guo, Li~Jing, David
  Schnurr, Joe Taylor, Troy Luhman, Eric Luhman, Clarence Ng, Ricky Wang, and
  Aditya Ramesh.
\newblock Video generation models as world simulators, 2024.
\newblock URL
  \url{https://openai.com/research/video-generation-models-as-world-simulators}.

\bibitem[Cassell et~al.(2000)Cassell, Sullivan, Prevost, and
  Churchill]{cassell2000embodied}
Justine Cassell, Joseph Sullivan, Scott Prevost, and Elizabeth~F. Churchill,
  editors.
\newblock \emph{Embodied Conversational Agents}.
\newblock MIT Press, Cambridge, MA, 2000.
\newblock ISBN 9780262032780.

\bibitem[Chen et~al.(2025)Chen, Zeng, Lin, Li, Ma, and Shou]{livecc}
Joya Chen, Ziyun Zeng, Yiqi Lin, Wei Li, Zejun Ma, and Mike~Zheng Shou.
\newblock Livecc: Learning video llm with streaming speech transcription at
  scale.
\newblock In \emph{CVPR}, 2025.

\bibitem[Chung and Zisserman(2016)]{Chung16a}
Joon~Son Chung and Andrew Zisserman.
\newblock Out of time: automated lip sync in the wild.
\newblock In \emph{Workshop on Multi-view Lip-reading, ACCV}, 2016.

\bibitem[Cui et~al.(2025)Cui, Li, Zhan, Shang, Cheng, Ma, Mu, Zhou, Wang, and
  Zhu]{cui2025hallo3highlydynamicrealistic}
Jiahao Cui, Hui Li, Yun Zhan, Hanlin Shang, Kaihui Cheng, Yuqi Ma, Shan Mu,
  Hang Zhou, Jingdong Wang, and Siyu Zhu.
\newblock Hallo3: Highly dynamic and realistic portrait image animation with
  video diffusion transformer.
\newblock In \emph{Proceedings of the Computer Vision and Pattern Recognition
  Conference, CVPR'25}, pages 21086--21095, 2025.

\bibitem[Cui et~al.(2026{\natexlab{a}})Cui, Wu, Li, Yang, Li, Wang, Bai, Ban,
  and Hsieh]{cui2025selfforcingminutescalehighqualityvideo}
Justin Cui, Jie Wu, Ming Li, Tao Yang, Xiaojie Li, Rui Wang, Andrew Bai,
  Yuanhao Ban, and Cho-Jui Hsieh.
\newblock Self-forcing++: Towards minute-scale high-quality video generation.
\newblock In \emph{The Fourteenth International Conference on Learning
  Representations, ICLR'26}, 2026{\natexlab{a}}.

\bibitem[Cui et~al.(2026{\natexlab{b}})Cui, Hu, Zhang, Yang, Shi, and
  Liu]{cui2026avatarforcingonestepstreamingtalking}
Liyuan Cui, Wentao Hu, Wenyuan Zhang, Zesong Yang, Fan Shi, and Xiaoqiang Liu.
\newblock Avatarforcing: One-step streaming talking avatars via local-future
  sliding-window denoising, 2026{\natexlab{b}}.
\newblock URL \url{https://arxiv.org/abs/2603.14331}.

\bibitem[D{\'e}fossez et~al.(2024)D{\'e}fossez, Mazar{\'e}, Orsini, Royer,
  P{\'e}rez, J{\'e}gou, Grave, and Zeghidour]{defossez2024moshi}
Alexandre D{\'e}fossez, Laurent Mazar{\'e}, Manu Orsini, Am{\'e}lie Royer,
  Patrick P{\'e}rez, Herv{\'e} J{\'e}gou, Edouard Grave, and Neil Zeghidour.
\newblock Moshi: a speech-text foundation model for real-time dialogue, 2024.
\newblock URL \url{https://arxiv.org/abs/2410.00037}.

\bibitem[Deng et~al.(2020)Deng, Guo, Ververas, Kotsia, and
  Zafeiriou]{deng2020retinaface}
Jiankang Deng, Jia Guo, Evangelos Ververas, Irene Kotsia, and Stefanos
  Zafeiriou.
\newblock {RetinaFace}: Single-shot multi-level face localisation in the wild.
\newblock In \emph{Proceedings of the IEEE/CVF Conference on Computer Vision
  and Pattern Recognition (CVPR)}, pages 5203--5212, 2020.

\bibitem[Deng et~al.(2022)Deng, Guo, Yang, Xue, Kotsia, and
  Zafeiriou]{Deng_2022}
Jiankang Deng, Jia Guo, Jing Yang, Niannan Xue, Irene Kotsia, and Stefanos
  Zafeiriou.
\newblock {ArcFace}: Additive angular margin loss for deep face recognition.
\newblock \emph{IEEE Transactions on Pattern Analysis and Machine
  Intelligence}, 44\penalty0 (10):\penalty0 5962–5979, October 2022.
\newblock ISSN 1939-3539.
\newblock \doi{10.1109/tpami.2021.3087709}.
\newblock URL \url{http://dx.doi.org/10.1109/TPAMI.2021.3087709}.

\bibitem[Dhariwal and Nichol(2021)]{dhariwal2021diffusionmodelsbeatgans}
Prafulla Dhariwal and Alex Nichol.
\newblock {Diffusion Models Beat GANs on Image Synthesis}.
\newblock \emph{Advances in neural information processing systems},
  34:\penalty0 8780--8794, 2021.

\bibitem[Dong et~al.(2024)Dong, Feng, Guessous, Liang, and
  He]{dong2024flexattention}
Juechu Dong, Boyuan Feng, Driss Guessous, Yanbo Liang, and Horace He.
\newblock {FlexAttention}: A programming model for generating optimized
  attention kernels.
\newblock \emph{arXiv preprint arXiv:2412.05496}, 2024.

\bibitem[Gao et~al.(2025)Gao, Hu, Hu, Huang, Ji, Meng, Qi, Qiao, Shen, Song,
  Sun, Tian, Wang, Wang, Wang, Xiao, Xu, Zhang, Zhang, Zhang, Zhang, Zhou, and
  Zhuo]{gao2025wans2vaudiodrivencinematicvideo}
Xin Gao, Li~Hu, Siqi Hu, Mingyang Huang, Chaonan Ji, Dechao Meng, Jinwei Qi,
  Penchong Qiao, Zhen Shen, Yafei Song, Ke~Sun, Linrui Tian, Guangyuan Wang,
  Qi~Wang, Zhongjian Wang, Jiayu Xiao, Sheng Xu, Bang Zhang, Peng Zhang, Xindi
  Zhang, Zhe Zhang, Jingren Zhou, and Lian Zhuo.
\newblock {Wan-S2V}: Audio-driven cinematic video generation, 2025.
\newblock URL \url{https://arxiv.org/abs/2508.18621}.

\bibitem[Goodfellow et~al.(2014)Goodfellow, Pouget-Abadie, Mirza, Xu,
  Warde-Farley, Ozair, Courville, and
  Bengio]{goodfellow2014generativeadversarialnetworks}
Ian~J. Goodfellow, Jean Pouget-Abadie, Mehdi Mirza, Bing Xu, David
  Warde-Farley, Sherjil Ozair, Aaron Courville, and Yoshua Bengio.
\newblock Generative adversarial nets.
\newblock \emph{Advances in neural information processing systems}, 27, 2014.

\bibitem[Heusel et~al.(2017)Heusel, Ramsauer, Unterthiner, Nessler, and
  Hochreiter]{heusel2018ganstrainedtimescaleupdate}
Martin Heusel, Hubert Ramsauer, Thomas Unterthiner, Bernhard Nessler, and Sepp
  Hochreiter.
\newblock {GAN}s trained by a two time-scale update rule converge to a local
  nash equilibrium.
\newblock \emph{Advances in neural information processing systems}, 30, 2017.

\bibitem[Ho et~al.(2020)Ho, Jain, and
  Abbeel]{ho2020denoisingdiffusionprobabilisticmodels}
Jonathan Ho, Ajay Jain, and Pieter Abbeel.
\newblock Denoising diffusion probabilistic models.
\newblock \emph{Advances in neural information processing systems},
  33:\penalty0 6840--6851, 2020.

\bibitem[Ho et~al.(2022)Ho, Chan, Saharia, Whang, Gao, Gritsenko, Kingma,
  Poole, Norouzi, Fleet, and Salimans]{ho2022imagenvideohighdefinition}
Jonathan Ho, William Chan, Chitwan Saharia, Jay Whang, Ruiqi Gao, Alexey
  Gritsenko, Diederik~P. Kingma, Ben Poole, Mohammad Norouzi, David~J. Fleet,
  and Tim Salimans.
\newblock Imagen video: High definition video generation with diffusion models,
  2022.
\newblock URL \url{https://arxiv.org/abs/2210.02303}.

\bibitem[Hu et~al.(2022)Hu, yelong shen, Wallis, Allen-Zhu, Li, Wang, Wang, and
  Chen]{hu2021loralowrankadaptationlarge}
Edward~J Hu, yelong shen, Phillip Wallis, Zeyuan Allen-Zhu, Yuanzhi Li, Shean
  Wang, Lu~Wang, and Weizhu Chen.
\newblock Lo{RA}: Low-rank adaptation of large language models.
\newblock In \emph{International Conference on Learning Representations,
  ICLR'22}, 2022.
\newblock URL \url{https://openreview.net/forum?id=nZeVKeeFYf9}.

\bibitem[Huang et~al.(2026{\natexlab{a}})Huang, Li, He, Zhou, and
  Shechtman]{huang2025selfforcingbridgingtraintest}
Xun Huang, Zhengqi Li, Guande He, Mingyuan Zhou, and Eli Shechtman.
\newblock Self forcing: Bridging the train-test gap in autoregressive video
  diffusion.
\newblock \emph{Advances in Neural Information Processing Systems},
  38:\penalty0 167283--167308, 2026{\natexlab{a}}.

\bibitem[Huang et~al.(2026{\natexlab{b}})Huang, Guo, Wu, Zhang, Huang, Gan,
  Liu, Zhao, Chen, Liu, and Hoi]{huang2026liveavatarstreamingrealtime}
Yubo Huang, Hailong Guo, Fangtai Wu, Shifeng Zhang, Shijie Huang, Qijun Gan,
  Lin Liu, Sirui Zhao, Enhong Chen, Jiaming Liu, and Steven Hoi.
\newblock Live avatar: Streaming real-time audio-driven avatar generation with
  infinite length, 2026{\natexlab{b}}.
\newblock URL \url{https://arxiv.org/abs/2512.04677}.

\bibitem[Karras et~al.(2019)Karras, Laine, and
  Aila]{karras2019stylebasedgeneratorarchitecturegenerative}
Tero Karras, Samuli Laine, and Timo Aila.
\newblock A style-based generator architecture for generative adversarial
  networks.
\newblock In \emph{Proceedings of the IEEE/CVF conference on computer vision
  and pattern recognition, CVPR'19}, pages 4401--4410, 2019.

\bibitem[Karras et~al.(2022)Karras, Aittala, Aila, and
  Laine]{karras2022elucidatingdesignspacediffusionbased}
Tero Karras, Miika Aittala, Timo Aila, and Samuli Laine.
\newblock Elucidating the design space of diffusion-based generative models.
\newblock \emph{Advances in neural information processing systems},
  35:\penalty0 26565--26577, 2022.

\bibitem[Kong et~al.(2025)Kong, Tian, Zhang, Min, Dai, Zhou, Xiong, Li, Wu,
  Zhang, Wu, Lin, Yuan, Long, Wang, Wang, Li, Huang, Yang, Tan, Wang, Song,
  Bai, Wu, Xue, Wang, Wang, Liu, Li, Li, Wang, Yu, Deng, Li, Chen, Cui, Peng,
  Yu, He, Xu, Zhou, Xu, Tao, Lu, Liu, Zhou, Wang, Yang, Wang, Liu, Jiang, and
  Zhong]{kong2025hunyuanvideosystematicframeworklarge}
Weijie Kong, Qi~Tian, Zijian Zhang, Rox Min, Zuozhuo Dai, Jin Zhou, Jiangfeng
  Xiong, Xin Li, Bo~Wu, Jianwei Zhang, Kathrina Wu, Qin Lin, Junkun Yuan,
  Yanxin Long, Aladdin Wang, Andong Wang, Changlin Li, Duojun Huang, Fang Yang,
  Hao Tan, Hongmei Wang, Jacob Song, Jiawang Bai, Jianbing Wu, Jinbao Xue, Joey
  Wang, Kai Wang, Mengyang Liu, Pengyu Li, Shuai Li, Weiyan Wang, Wenqing Yu,
  Xinchi Deng, Yang Li, Yi~Chen, Yutao Cui, Yuanbo Peng, Zhentao Yu, Zhiyu He,
  Zhiyong Xu, Zixiang Zhou, Zunnan Xu, Yangyu Tao, Qinglin Lu, Songtao Liu, Dax
  Zhou, Hongfa Wang, Yong Yang, Di~Wang, Yuhong Liu, Jie Jiang, and Caesar
  Zhong.
\newblock Hunyuanvideo: A systematic framework for large video generative
  models, 2025.
\newblock URL \url{https://arxiv.org/abs/2412.03603}.

\bibitem[Li et~al.(2026)Li, Wang, Zhou, Feng, Luo, Zhang, Wu, and
  He]{li2026joystreamerflashrealtimeinfiniteaudiodriven}
Chaochao Li, Ruikui Wang, Liangbo Zhou, Jinheng Feng, Huaishao Luo, Huan Zhang,
  Youzheng Wu, and Xiaodong He.
\newblock Joystreamer-flash: Real-time and infinite audio-driven avatar
  generation with autoregressive diffusion, 2026.
\newblock URL \url{https://arxiv.org/abs/2512.11423}.

\bibitem[Lin et~al.(2025)Lin, Jiang, Yang, Zheng, Liang, Zhang, and
  Liu]{lin2025omnihuman1rethinkingscalinguponestage}
Gaojie Lin, Jianwen Jiang, Jiaqi Yang, Zerong Zheng, Chao Liang, Yuan Zhang,
  and Jingtuo Liu.
\newblock Omnihuman-1: Rethinking the scaling-up of one-stage conditioned human
  animation models.
\newblock In \emph{Proceedings of the IEEE/CVF International Conference on
  Computer Vision, ICCV'25}, pages 13847--13858, 2025.

\bibitem[Lipman et~al.(2023)Lipman, Chen, Ben-Hamu, Nickel, and
  Le]{lipman2023flowmatchinggenerativemodeling}
Yaron Lipman, Ricky T.~Q. Chen, Heli Ben-Hamu, Maximilian Nickel, and Matthew
  Le.
\newblock Flow matching for generative modeling.
\newblock In \emph{The Eleventh International Conference on Learning
  Representations, ICLR'23}, 2023.
\newblock URL \url{https://openreview.net/forum?id=PqvMRDCJT9t}.

\bibitem[Liu et~al.(2026)Liu, Hu, Xu, Shan, and
  Lu]{liu2025rollingforcingautoregressivelong}
Kunhao Liu, Wenbo Hu, Jiale Xu, Ying Shan, and Shijian Lu.
\newblock Rolling forcing: Autoregressive long video diffusion in real time,
  2026.

\bibitem[Liu et~al.(2023)Liu, Gong, and qiang
  liu]{liu2022flowstraightfastlearning}
Xingchao Liu, Chengyue Gong, and qiang liu.
\newblock Flow straight and fast: Learning to generate and transfer data with
  rectified flow.
\newblock In \emph{The Eleventh International Conference on Learning
  Representations}, 2023.
\newblock URL \url{https://openreview.net/forum?id=XVjTT1nw5z}.

\bibitem[Low and
  Wang(2025)]{low2025talkingmachinesrealtimeaudiodrivenfacetimestyle}
Chetwin Low and Weimin Wang.
\newblock Talkingmachines: Real-time audio-driven facetime-style video via
  autoregressive diffusion models, 2025.
\newblock URL \url{https://arxiv.org/abs/2506.03099}.

\bibitem[Luo et~al.(2025)Luo, Favero, Luo, Jolicoeur-Martineau, and
  Pal]{luo2024fvdenhancedevaluationmetrics}
Ge~Ya Luo, Gian~Mario Favero, Zhi~Hao Luo, Alexia Jolicoeur-Martineau, and
  Christopher Pal.
\newblock Beyond {FVD}: An enhanced evaluation metrics for video generation
  distribution quality.
\newblock In \emph{International Conference on Learning Representations,
  ICLR'25}, volume 2025, pages 66218--66252, 2025.

\bibitem[Mahy et~al.(1994)Mahy, Van~Eycken, and
  Oosterlinck]{mahy1994evaluation}
M.~Mahy, L.~Van~Eycken, and A.~Oosterlinck.
\newblock Evaluation of uniform color spaces developed after the adoption of
  {CIELAB} and {CIELUV}.
\newblock \emph{Color Research \& Application}, 19\penalty0 (2):\penalty0
  105--121, 1994.

\bibitem[McGurk and MacDonald(1976)]{mcgurk1976hearing}
Harry McGurk and John MacDonald.
\newblock Hearing lips and seeing voices.
\newblock \emph{Nature}, 264\penalty0 (5588):\penalty0 746--748, December 1976.
\newblock \doi{10.1038/264746a0}.

\bibitem[Pech-Pacheco et~al.(2000)Pech-Pacheco, Crist{\'o}bal,
  Chamorro-Mart{\'\i}nez, and Fern{\'a}ndez-Valdivia]{pechpacheco2000diatom}
Jos{\'e}~Luis Pech-Pacheco, Gabriel Crist{\'o}bal, Jes{\'u}s
  Chamorro-Mart{\'\i}nez, and Joaqu{\'\i}n Fern{\'a}ndez-Valdivia.
\newblock Diatom autofocusing in brightfield microscopy: a comparative study.
\newblock In \emph{Proceedings of the 15th International Conference on Pattern
  Recognition (ICPR)}, volume~3, pages 314--317, 2000.

\bibitem[Polyak et~al.(2025)Polyak, Zohar, Brown, Tjandra, Sinha, Lee, Vyas,
  Shi, Ma, Chuang, Yan, Choudhary, Wang, Sethi, Pang, Ma, Misra, Hou, Wang,
  Jagadeesh, Li, Zhang, Singh, Williamson, Le, Yu, Singh, Zhang, Vajda, Duval,
  Girdhar, Sumbaly, Rambhatla, Tsai, Azadi, Datta, Chen, Bell, Ramaswamy,
  Sheynin, Bhattacharya, Motwani, Xu, Li, Hou, Hsu, Yin, Dai, Taigman, Luo,
  Liu, Wu, Zhao, Kirstain, He, He, Pumarola, Thabet, Sanakoyeu, Mallya, Guo,
  Araya, Kerr, Wood, Liu, Peng, Vengertsev, Schonfeld, Blanchard, Juefei-Xu,
  Nord, Liang, Hoffman, Kohler, Fire, Sivakumar, Chen, Yu, Gao, Georgopoulos,
  Moritz, Sampson, Li, Parmeggiani, Fine, Fowler, Petrovic, and
  Du]{polyak2025moviegencastmedia}
Adam Polyak, Amit Zohar, Andrew Brown, Andros Tjandra, Animesh Sinha, Ann Lee,
  Apoorv Vyas, Bowen Shi, Chih-Yao Ma, Ching-Yao Chuang, David Yan, Dhruv
  Choudhary, Dingkang Wang, Geet Sethi, Guan Pang, Haoyu Ma, Ishan Misra,
  Ji~Hou, Jialiang Wang, Kiran Jagadeesh, Kunpeng Li, Luxin Zhang, Mannat
  Singh, Mary Williamson, Matt Le, Matthew Yu, Mitesh~Kumar Singh, Peizhao
  Zhang, Peter Vajda, Quentin Duval, Rohit Girdhar, Roshan Sumbaly, Sai~Saketh
  Rambhatla, Sam Tsai, Samaneh Azadi, Samyak Datta, Sanyuan Chen, Sean Bell,
  Sharadh Ramaswamy, Shelly Sheynin, Siddharth Bhattacharya, Simran Motwani,
  Tao Xu, Tianhe Li, Tingbo Hou, Wei-Ning Hsu, Xi~Yin, Xiaoliang Dai, Yaniv
  Taigman, Yaqiao Luo, Yen-Cheng Liu, Yi-Chiao Wu, Yue Zhao, Yuval Kirstain,
  Zecheng He, Zijian He, Albert Pumarola, Ali Thabet, Artsiom Sanakoyeu, Arun
  Mallya, Baishan Guo, Boris Araya, Breena Kerr, Carleigh Wood, Ce~Liu, Cen
  Peng, Dimitry Vengertsev, Edgar Schonfeld, Elliot Blanchard, Felix Juefei-Xu,
  Fraylie Nord, Jeff Liang, John Hoffman, Jonas Kohler, Kaolin Fire, Karthik
  Sivakumar, Lawrence Chen, Licheng Yu, Luya Gao, Markos Georgopoulos, Rashel
  Moritz, Sara~K. Sampson, Shikai Li, Simone Parmeggiani, Steve Fine, Tara
  Fowler, Vladan Petrovic, and Yuming Du.
\newblock Movie gen: A cast of media foundation models, 2025.
\newblock URL \url{https://arxiv.org/abs/2410.13720}.

\bibitem[Raffel et~al.(2020)Raffel, Shazeer, Roberts, Lee, Narang, Matena,
  Zhou, Li, and Liu]{raffel2023exploringlimitstransferlearning}
Colin Raffel, Noam Shazeer, Adam Roberts, Katherine Lee, Sharan Narang, Michael
  Matena, Yanqi Zhou, Wei Li, and Peter~J. Liu.
\newblock Exploring the limits of transfer learning with a unified text-to-text
  transformer.
\newblock \emph{Journal of machine learning research}, 21\penalty0
  (140):\penalty0 1--67, 2020.

\bibitem[Rombach et~al.(2022)Rombach, Blattmann, Lorenz, Esser, and
  Ommer]{rombach2022highresolutionimagesynthesislatent}
Robin Rombach, Andreas Blattmann, Dominik Lorenz, Patrick Esser, and Björn
  Ommer.
\newblock High-resolution image synthesis with latent diffusion models.
\newblock In \emph{Proceedings of the IEEE/CVF conference on computer vision
  and pattern recognition, CVPR'22}, pages 10684--10695, 2022.

\bibitem[Roy et~al.(2026)Roy, Raiman, gil Lee, Ene, Kirby, Kim, Kim, and
  Catanzaro]{roy2026personaplexvoicerolecontrol}
Rajarshi Roy, Jonathan Raiman, Sang gil Lee, Teodor-Dumitru Ene, Robert Kirby,
  Sungwon Kim, Jaehyeon Kim, and Bryan Catanzaro.
\newblock Personaplex: Voice and role control for full duplex conversational
  speech models, 2026.
\newblock URL \url{https://arxiv.org/abs/2602.06053}.

\bibitem[Shen et~al.(2026)Shen, Qiao, Yu, Zhou, Yu, Zhan, Wang, Tao, Yin, and
  Liu]{shen2026soulxflashtalkrealtimeinfinitestreaming}
Le~Shen, Qian Qiao, Tan Yu, Ke~Zhou, Tianhang Yu, Yu~Zhan, Zhenjie Wang, Ming
  Tao, Shunshun Yin, and Siyuan Liu.
\newblock {SoulX-FlashTalk}: Real-time infinite streaming of audio-driven
  avatars via self-correcting bidirectional distillation, 2026.
\newblock URL \url{https://arxiv.org/abs/2512.23379}.

\bibitem[Singer et~al.(2023)Singer, Polyak, Hayes, Yin, An, Zhang, Hu, Yang,
  Ashual, Gafni, Parikh, Gupta, and
  Taigman]{singer2022makeavideotexttovideogenerationtextvideo}
Uriel Singer, Adam Polyak, Thomas Hayes, Xi~Yin, Jie An, Songyang Zhang, Qiyuan
  Hu, Harry Yang, Oron Ashual, Oran Gafni, Devi Parikh, Sonal Gupta, and Yaniv
  Taigman.
\newblock Make-a-video: Text-to-video generation without text-video data.
\newblock In \emph{The Eleventh International Conference on Learning
  Representations, ICLR'23}, 2023.

\bibitem[Song et~al.(2021{\natexlab{a}})Song, Meng, and
  Ermon]{song2022denoisingdiffusionimplicitmodels}
Jiaming Song, Chenlin Meng, and Stefano Ermon.
\newblock Denoising diffusion implicit models.
\newblock In \emph{International Conference on Learning Representations},
  2021{\natexlab{a}}.
\newblock URL \url{https://openreview.net/forum?id=St1giarCHLP}.

\bibitem[Song et~al.(2021{\natexlab{b}})Song, Sohl-Dickstein, Kingma, Kumar,
  Ermon, and Poole]{song2021scorebasedgenerativemodelingstochastic}
Yang Song, Jascha Sohl-Dickstein, Diederik~P. Kingma, Abhishek Kumar, Stefano
  Ermon, and Ben Poole.
\newblock Score-based generative modeling through stochastic differential
  equations.
\newblock In \emph{International Conference on Learning Representations,
  ICLR'21}, 2021{\natexlab{b}}.

\bibitem[Su et~al.(2024)Su, Ahmed, Lu, Pan, Bo, and Liu]{su2024roformer}
Jianlin Su, Murtadha Ahmed, Yu~Lu, Shengfeng Pan, Wen Bo, and Yunfeng Liu.
\newblock Roformer: Enhanced transformer with rotary position embedding.
\newblock \emph{Neurocomputing}, 568:\penalty0 127063, 2024.

\bibitem[Sumby and Pollack(1954)]{sumby1954visual}
W.~H. Sumby and Irwin Pollack.
\newblock Visual contribution to speech intelligibility in noise.
\newblock \emph{The Journal of the Acoustical Society of America}, 26\penalty0
  (2):\penalty0 212--215, March 1954.
\newblock \doi{10.1121/1.1907309}.

\bibitem[Sun et~al.(2026)Sun, Peng, Ma, Chen, Zhou, Zhou, Zhang, Zhang, Zhou,
  Lu, and Liu]{sun2026streamavatar}
Zhiyao Sun, Ziqiao Peng, Yifeng Ma, Yi~Chen, Zhengguang Zhou, Zixiang Zhou,
  Guozhen Zhang, Youliang Zhang, Yuan Zhou, Qinglin Lu, and Yong-Jin Liu.
\newblock {StreamAvatar}: Streaming diffusion models for real-time interactive
  human avatars.
\newblock In \emph{Proceedings of the IEEE/CVF Conference on Computer Vision
  and Pattern Recognition (CVPR)}, 2026.

\bibitem[Teed and Deng(2020)]{teed2020raftrecurrentallpairsfield}
Zachary Teed and Jia Deng.
\newblock {RAFT}: Recurrent all-pairs field transforms for optical flow.
\newblock In \emph{European conference on computer vision, ECCV'20}, pages
  402--419. Springer, 2020.

\bibitem[Unterthiner et~al.(2019)Unterthiner, van Steenkiste, Kurach, Marinier,
  Michalski, and Gelly]{unterthiner2019accurategenerativemodelsvideo}
Thomas Unterthiner, Sjoerd van Steenkiste, Karol Kurach, Raphael Marinier,
  Marcin Michalski, and Sylvain Gelly.
\newblock Towards accurate generative models of video: A new metric \&
  challenges, 2019.
\newblock URL \url{https://arxiv.org/abs/1812.01717}.

\bibitem[Wan et~al.(2025)Wan, Wang, Ai, Wen, Mao, Xie, Chen, Yu, Zhao, Yang,
  Zeng, Wang, Zhang, Zhou, Wang, Chen, Zhu, Zhao, Yan, Huang, Feng, Zhang, Li,
  Wu, Chu, Feng, Zhang, Sun, Fang, Wang, Gui, Weng, Shen, Lin, Wang, Wang,
  Zhou, Wang, Shen, Yu, Shi, Huang, Xu, Kou, Lv, Li, Liu, Wang, Zhang, Huang,
  Li, Wu, Liu, Pan, Zheng, Hong, Shi, Feng, Jiang, Han, Wu, and
  Liu]{wan2025wanopenadvancedlargescale}
Team Wan, Ang Wang, Baole Ai, Bin Wen, Chaojie Mao, Chen-Wei Xie, Di~Chen,
  Feiwu Yu, Haiming Zhao, Jianxiao Yang, Jianyuan Zeng, Jiayu Wang, Jingfeng
  Zhang, Jingren Zhou, Jinkai Wang, Jixuan Chen, Kai Zhu, Kang Zhao, Keyu Yan,
  Lianghua Huang, Mengyang Feng, Ningyi Zhang, Pandeng Li, Pingyu Wu, Ruihang
  Chu, Ruili Feng, Shiwei Zhang, Siyang Sun, Tao Fang, Tianxing Wang, Tianyi
  Gui, Tingyu Weng, Tong Shen, Wei Lin, Wei Wang, Wei Wang, Wenmeng Zhou, Wente
  Wang, Wenting Shen, Wenyuan Yu, Xianzhong Shi, Xiaoming Huang, Xin Xu, Yan
  Kou, Yangyu Lv, Yifei Li, Yijing Liu, Yiming Wang, Yingya Zhang, Yitong
  Huang, Yong Li, You Wu, Yu~Liu, Yulin Pan, Yun Zheng, Yuntao Hong, Yupeng
  Shi, Yutong Feng, Zeyinzi Jiang, Zhen Han, Zhi-Fan Wu, and Ziyu Liu.
\newblock Wan: Open and advanced large-scale video generative models, 2025.
\newblock URL \url{https://arxiv.org/abs/2503.20314}.

\bibitem[Wang et~al.(2025)Wang, Wang, Ma, Lin, and
  Zhou]{wang2025talkversedemocratizingminutelongaudiodriven}
Zhenzhi Wang, Jian Wang, Ke~Ma, Dahua Lin, and Bing Zhou.
\newblock Talkverse: Democratizing minute-long audio-driven video generation,
  2025.
\newblock URL \url{https://arxiv.org/abs/2512.14938}.

\bibitem[Xiao et~al.(2024)Xiao, Tian, Chen, Han, and
  Lewis]{xiao2024efficientstreaminglanguagemodels}
Guangxuan Xiao, Yuandong Tian, Beidi Chen, Song Han, and Mike Lewis.
\newblock Efficient streaming language models with attention sinks.
\newblock In \emph{International Conference on Learning Representations,
  ICLR'24}, volume 2024, pages 21875--21895, 2024.

\bibitem[Yang et~al.(2025{\natexlab{a}})Yang, Huang, Chu, Xiao, Zhao, Wang, Li,
  Xie, Chen, Lu, Han, and Chen]{yang2025longliverealtimeinteractivelong}
Shuai Yang, Wei Huang, Ruihang Chu, Yicheng Xiao, Yuyang Zhao, Xianbang Wang,
  Muyang Li, Enze Xie, Yingcong Chen, Yao Lu, Song Han, and Yukang Chen.
\newblock {LongLive}: Real-time interactive long video generation,
  2025{\natexlab{a}}.
\newblock URL \url{https://arxiv.org/abs/2509.22622}.

\bibitem[Yang et~al.(2026)Yang, Zhang, Huang, Chen, Wu, He, Cai, Li, and
  Jiang]{yang2026anchorforcing}
Yang Yang, Tianyi Zhang, Wei Huang, Jinwei Chen, Boxi Wu, Xiaofei He, Deng Cai,
  Bo~Li, and Peng-Tao Jiang.
\newblock Anchor forcing: Anchor memory and tri-region {RoPE} for interactive
  streaming video diffusion, 2026.
\newblock URL \url{https://arxiv.org/abs/2603.13405}.

\bibitem[Yang et~al.(2025{\natexlab{b}})Yang, Teng, Zheng, Ding, Huang, Xu,
  Yang, Hong, Zhang, Feng, Yin, Zhang, Wang, Cheng, Xu, Gu, Dong, and
  Tang]{yang2025cogvideoxtexttovideodiffusionmodels}
Zhuoyi Yang, Jiayan Teng, Wendi Zheng, Ming Ding, Shiyu Huang, Jiazheng Xu,
  Yuanming Yang, Wenyi Hong, Xiaohan Zhang, Guanyu Feng, Da~Yin, Yuxuan Zhang,
  Weihan Wang, Yean Cheng, Bin Xu, Xiaotao Gu, Yuxiao Dong, and Jie Tang.
\newblock Cogvideox: Text-to-video diffusion models with an expert transformer.
\newblock In \emph{International Conference on Learning Representations,
  ICLR'25}, volume 2025, pages 83048--83077, 2025{\natexlab{b}}.

\bibitem[Yin et~al.(2024{\natexlab{a}})Yin, Gharbi, Park, Zhang, Shechtman,
  Durand, and Freeman]{yin2024improveddistributionmatchingdistillation}
Tianwei Yin, Michaël Gharbi, Taesung Park, Richard Zhang, Eli Shechtman, Fredo
  Durand, and William~T. Freeman.
\newblock Improved distribution matching distillation for fast image synthesis.
\newblock \emph{Advances in neural information processing systems},
  37:\penalty0 47455--47487, 2024{\natexlab{a}}.

\bibitem[Yin et~al.(2024{\natexlab{b}})Yin, Gharbi, Zhang, Shechtman, Durand,
  Freeman, and Park]{yin2024onestepdiffusiondistributionmatching}
Tianwei Yin, Michaël Gharbi, Richard Zhang, Eli Shechtman, Fredo Durand,
  William~T. Freeman, and Taesung Park.
\newblock One-step diffusion with distribution matching distillation.
\newblock In \emph{Proceedings of the IEEE/CVF conference on computer vision
  and pattern recognition, CVPR'24}, pages 6613--6623, 2024{\natexlab{b}}.

\bibitem[Yin et~al.(2025)Yin, Zhang, Zhang, Freeman, Durand, Shechtman, and
  Huang]{yin2025slowbidirectionalfastautoregressive}
Tianwei Yin, Qiang Zhang, Richard Zhang, William~T. Freeman, Fredo Durand, Eli
  Shechtman, and Xun Huang.
\newblock From slow bidirectional to fast autoregressive video diffusion
  models.
\newblock In \emph{Proceedings of the IEEE/CVF Conference on Computer Vision
  and Pattern Recognition, CVPR'25}, pages 22963--22974, 2025.

\bibitem[Zeghidour et~al.(2025)Zeghidour, Kharitonov, Orsini, Volhejn,
  de~Marmiesse, Grave, Pérez, Mazaré, and Défossez]{dsm}
Neil Zeghidour, Eugene Kharitonov, Manu Orsini, Václav Volhejn, Gabriel
  de~Marmiesse, Edouard Grave, Patrick Pérez, Laurent Mazaré, and Alexandre
  Défossez.
\newblock Streaming sequence-to-sequence learning with delayed streams
  modeling.
\newblock Technical report, 2025.
\newblock URL \url{https://arxiv.org/abs/2509.08753}.

\bibitem[Zhang et~al.(2026)Zhang, Cai, Li, Wetzstein, and
  Agrawala]{zhang2025framecontextpackingdrift}
Lvmin Zhang, Shengqu Cai, Muyang Li, Gordon Wetzstein, and Maneesh Agrawala.
\newblock Frame context packing and drift prevention in next-frame-prediction
  video diffusion models.
\newblock \emph{Advances in Neural Information Processing Systems},
  38:\penalty0 30546--30566, 2026.

\bibitem[Zheng et~al.(2024)Zheng, Peng, Yang, Shen, Li, Liu, Zhou, Li, and
  You]{zheng2024opensorademocratizingefficientvideo}
Zangwei Zheng, Xiangyu Peng, Tianji Yang, Chenhui Shen, Shenggui Li, Hongxin
  Liu, Yukun Zhou, Tianyi Li, and Yang You.
\newblock Open-sora: Democratizing efficient video production for all, 2024.
\newblock URL \url{https://arxiv.org/abs/2412.20404}.

\bibitem[Zhu et~al.(2022)Zhu, Wu, Zhu, Jiang, Tang, Zhang, Liu, and
  Loy]{zhu2022celebvhqlargescalevideofacial}
Hao Zhu, Wayne Wu, Wentao Zhu, Liming Jiang, Siwei Tang, Li~Zhang, Ziwei Liu,
  and Chen~Change Loy.
\newblock {CelebV-HQ}: A large-scale video facial attributes dataset.
\newblock In \emph{European conference on computer vision, ECCV'22}, pages
  650--667. Springer, 2022.

\bibitem[Zhu et~al.(2026)Zhu, Zhao, He, Su, Li, and
  Zhu]{zhu2026causalforcingautoregressivediffusion}
Hongzhou Zhu, Min Zhao, Guande He, Hang Su, Chongxuan Li, and Jun Zhu.
\newblock Causal forcing: Autoregressive diffusion distillation done right for
  high-quality real-time interactive video generation, 2026.
\newblock URL \url{https://arxiv.org/abs/2602.02214}.

\end{thebibliography}
